\documentclass[a4paper]{article}

\usepackage{preprint}

\usepackage[utf8]{inputenc} 
\usepackage{enumitem}
\usepackage{url}            
\usepackage{nicefrac}       
\usepackage{microtype}      
\usepackage{graphicx}
\usepackage{subcaption}
\usepackage[square,numbers,sort&compress]{natbib}
\usepackage{etoc}
\usepackage{doi}
\usepackage[dvipsnames,svgnames,hyperref,table]{xcolor}
\usepackage{algorithm}
\usepackage[noend]{algpseudocode}
\usepackage{tensor}

\usepackage{amsmath}
\usepackage{mathtools}
\usepackage{amsfonts}
\usepackage{amssymb}
\usepackage{dsfont}
\usepackage[none]{hyphenat}

\DeclareMathOperator*{\argmin}{argmin}

\makeatletter
\newcommand{\ostar}{\mathbin{\mathpalette\make@circled\star}}
\newcommand{\make@circled}[2]{%
  \ooalign{$\m@th#1\smallbigcirc{#1}$\cr\hidewidth$\m@th#1#2$\hidewidth\cr}%
}
\newcommand{\smallbigcirc}[1]{%
  \vcenter{\hbox{\scalebox{0.77778}{$\m@th#1\bigcirc$}}}%
}
\makeatother

\let\svthefootnote\thefootnote
\newcommand\freefootnote[1]{%
  \let\thefootnote\relax%
  \footnotetext{#1}%
  \let\thefootnote\svthefootnote%
}

\usepackage{booktabs,cellspace,multirow}
\makeatother

\usepackage[capitalize, noabbrev]{cleveref}
\usepackage{hyperref}
\hypersetup{colorlinks, breaklinks, linkcolor=WildStrawberry, urlcolor=LimeGreen, citecolor=VioletRed}

\usepackage{authblk}
\newcommand{\titlefootnotetext}[2]{{\renewcommand{\thefootnote}{#1}\footnotetext[0]{#2}}}

\title{FourCastNet 3: A geometric approach to probabilistic\\ machine-learning weather forecasting at scale}

\author[1,*]{Boris Bonev}
\author[1,*]{Thorsten Kurth}
\author[2,3]{Ankur Mahesh}
\author[1]{Mauro Bisson}
\author[1]{Jean Kossaifi}
\author[1]{Karthik Kashinath}
\author[4]{Anima Anandkumar}
\author[2,3]{William D. Collins}
\author[1]{Michael S. Pritchard}
\author[1]{Alexander Keller}
\affil[1]{NVIDIA Corporation, Santa Clara, CA 95051, United States}
\affil[2]{Lawrence Berkeley National Laboratory, Berkeley, CA 94720, United States}
\affil[3]{University of California, Berkeley, CA 94720, United States}
\affil[4]{California Institute of Technology, Pasadena, CA 91125, United States}

\renewcommand{\shorttitle}{FourCastNet 3}

\hypersetup{
pdftitle={FourCastNet 3: A geometric approach to probabilistic machine-learning weather forecasting at scale},
pdfsubject={cs.LG, physics.ao-ph, stat.ML, eess.SP},
pdfauthor={Boris Bonev, Thorsten Kurth, Ankur Mahesh, Mauro Bisson, Jean Kossaifi, Karthik Kashinath, Anima Anandkumar, William D. Collins, Michael S. Pritchard, Alex Keller},
pdfkeywords={weather prediction, ensemble prediction, geometric machine learning, probabilistic machine learning, signal processing, neural operators},
}

\usepackage{todonotes}

\begin{document}
\maketitle

\titlefootnotetext{*}{Equal contribution. Correspondence to \texttt{\{bbonev, tkurth\}@nvidia.com}.}

\begin{abstract}
FourCastNet 3 advances global weather modeling by implementing a scalable, geometric machine learning (ML) approach to probabilistic ensemble forecasting. The approach is designed to respect spherical geometry and to accurately model the spatially correlated probabilistic nature of the problem, resulting in stable spectra and realistic dynamics across multiple scales. FourCastNet 3 delivers forecasting accuracy that surpasses leading conventional ensemble models and rivals the best diffusion-based methods, while producing forecasts 8 to 60 times faster than these approaches. In contrast to other ML approaches, FourCastNet 3 demonstrates excellent probabilistic calibration and retains realistic spectra, even at extended lead times of up to 60 days.
All of these advances are realized using a purely convolutional neural network architecture tailored for spherical geometry.
Scalable and efficient large-scale training on 1024 GPUs and more is enabled by a novel training paradigm for combined model- and data-parallelism, inspired by domain decomposition methods in classical numerical models. Additionally, FourCastNet 3 enables rapid inference on a single GPU, producing a 60-day global forecast at 0.25\textdegree, 6-hourly resolution in under 4 minutes. 
Its computational efficiency, medium-range probabilistic skill, spectral fidelity, and rollout stability at subseasonal timescales make it a strong candidate for improving meteorological forecasting and early warning systems through large ensemble predictions.
\end{abstract}

\etocdepthtag.toc{mtmanuscript}
\etocsettagdepth{mtmanuscript}{subsection}
\etocsettagdepth{mtappendix}{none}

\begin{figure}[tb]
    \centering
    \includegraphics[width=\textwidth]{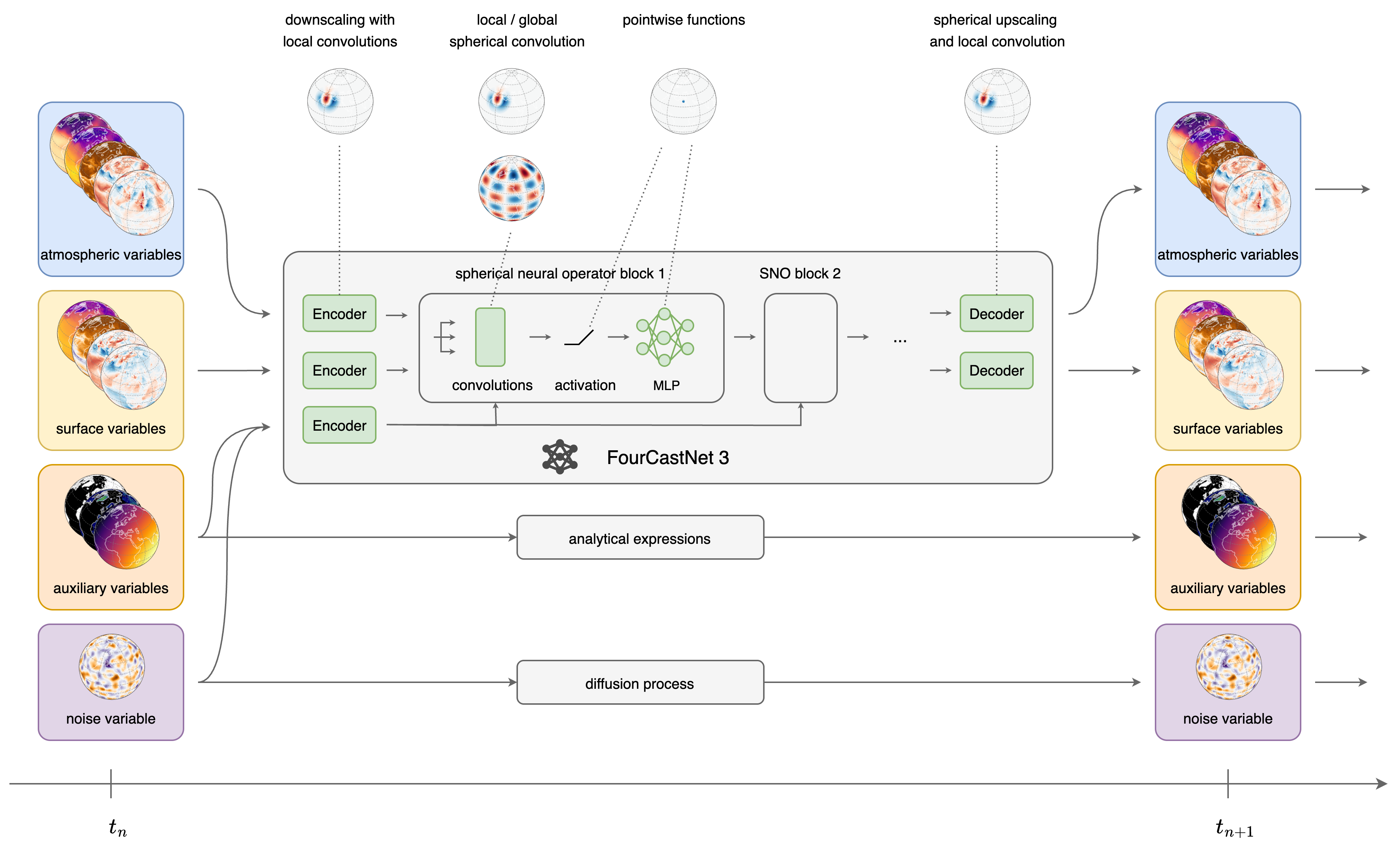}
    \caption{Schematic of the FourCastNet 3 model. The model predicts the state of the atmosphere at the next timestep, given the state at the previous timestep. Auxiliary variables such as the cosine zenith angle are computed from analytical expressions for each timestep and appended to the input. A hidden Markov model is obtained by conditioning FourCastNet 3 on a stochastic latent variable whose temporal dynamics are governed by a diffusion process on the sphere. The model itself is formed by an encoder, a decoder and 8 neural operator blocks. Each of these operations can be grouped into local, global and pointwise operations and therefore be formulated on arbitrary grids and resolutions, making FourCastNet 3 discretization independent. Green boxes illustrate learnable operations.}
    \label{fig:fcn3_schematic}
\end{figure}

\section{Introduction}
\label{sec:introduction}

\freefootnote{Code available at: \url{https://github.com/NVIDIA/makani}}

Numerical weather prediction (NWP) is central to modern meteorology, underpinning our ability to accurately understand and forecast atmospheric phenomena~\cite{Lynch2008}. Advances in mathematical modeling, computational power, and data assimilation have made NWP essential for weather forecasting, hazard mitigation, energy management, and climate studies.

Traditional NWP models, however, are computationally intensive, limiting their ability to deliver rapid, large-scale probabilistic forecasts. Recently, machine learning (ML) approaches have surpassed traditional NWP in forecast skill and speed, enabling rapid generation of large ensembles and opening new possibilities for weather and climate prediction~\cite{Pathak2022,Bi2022,Lam2022,kurth2022fourcastnetacceleratingglobalhighresolution,Price2024}. These advances support improved sampling of rare events and longer-range forecasts~\cite{Mahesh2024a,Mahesh2024b,Weyn2021}.
Despite these benefits, ML models face challenges: they may struggle with out-of-distribution events, physical consistency, and long-term stability~\cite{Bonev2023,Watt-Meyer2023,Karlbauer2024,Guan2025,Cresswell-Clay2025}.  Furthermore, commonly used evaluation metrics fail to fully capture the accuracy with which these models approximate the underlying dynamical systems \cite{Fang2025}. Hybrid models that combine ML and traditional NWP offer partial solutions~\cite{Kochkov2023}, but suffer from the same computational bottlenecks due to the Courant-Friedrichs-Lewy (CFL) condition, making them expensive to evaluate, especially at high resolution. Additionally, deterministic ML models often exhibit excessive smoothing, which are closer to ensemble averages, lacking the fidelity of traditional deterministic forecasts~\cite{Rasp2023,Brenowitz2024,Subich2025}.

Recently, probabilistic approaches have aimed to address the latter problem~\cite{Lessig2023,Price2024,Andrae2024,Lang2024}. GenCast~\cite{Price2024}, the state-of-the-art probabilistic ML model, has proven that a denoising diffusion model approach~\cite{Ho2020} is effective at modeling the probabilistic nature of atmospheric phenomena. However, this comes at a significant cost overhead during inference, due to the iterative nature of denoising. \citet{Lang2024} use a scoring rule based objective function instead \cite{Gneiting2007}, with the implied computational benefits over the diffusion approach. While effectively addressing blurring, both approaches lead to build-up of small-scale noise, requiring an ad-hoc truncation strategy in the latter case to suppress it. This build-up can be a precursor to blow-up in NWP models \cite{Lauritzen2011} and attaining stable spectra remains a key challenge.

Most of today's leading ML weather models repurpose mature architectures such as transformers and graph- neural-networks that were fundamentally developed for other scientific ML tasks~\cite{Vaswani2017,Pfaff2021}. This pragmatic approach enables competitive medium-range skill, as demonstrated by numerous models, with little to disambiguate between them \cite{Rasp2023}. As going beyond medium-range forecasts requires additional properties beyond medium-range skill, bespoke geometric approaches offer a simple and elegant alternative.
These methods are faithful to the underlying geometry, its topology and the symmetries of the underlying physics. However, bespoke methods come with significant engineering challenges, requiring custom implementations and engineering frameworks to achieve the necessary scale of model training that is in turn needed to achieve competitive skill~\cite{Kaplan2020}.

\paragraph{FourCastNet 3}
We introduce FourCastNet 3 (FCN3), a skillful, probabilistic ML weather forecasting system built as a hidden Markov model based on spherical signal processing primitives and a probabilistic loss function in the spectral domain. Our method is purely convolutional, leveraging both local and global spherical convolution kernels to better model the physical processes at various scales involved in weather phenomena, while respecting spherical geometry and its inherent symmetries, enabling realistic ensemble members.

FCN3 is trained end-to-end as an ensemble forecasting model, at scale. This approach retains the large speed-ups offered by ML models, facilitating one-step generation of ensemble members, making it both computationally efficient and accurate. To enable training of a large FCN3 model, on a large hourly dataset with multiple ensemble members, and multiple timesteps, we develop a hybrid machine learning paradigm for simultaneous model- and data-parallelism, inspired by traditional NWP methods. The computational domain is decomposed to simultaneously distribute both the model and the data during training. This is combined with distributed batch- and ensemble-parallelism, resulting in extremely efficient and scalable training, which enabled seamlessly scaling training to over 1000 GPUs.

FCN3 outperforms the integrated forecasting system's ensemble method (IFS-ENS)~\cite{Haiden2024}, the golden standard for traditional NWP methods, and nearly matches the medium-range forecast skill of GenCast~\cite{Price2024}, the leading probabilistic ML weather model, at double the temporal resolution. A single forecast of 15 days is computed in 60 seconds on a single NVIDIA H100 GPU - a speedup of ~8x over GenCast and ~60x over IFS-ENS. Simultaneously, it offers the key benefit of retaining stable predictions and accurate spectra well into the subseasonal range with lead times of up to 60 days. 
This key achievement mitigates the issue of blurring and addresses the issue of build-up of small-scale noise. The probabilistic skill, stability, spectral fidelity and low inference cost make FCN3 an interesting model with the potential of generating large ensembles with potential applications spanning medium-range to subseasonal forecasting.


\section{Probabilistic forecasts with hidden Markov models}
\label{sec:probabilistic_forecast}

FourCastNet 3 (FCN3) is formulated as a probabilistic model to address the chaotic nature of atmospheric phenomena. Given the current atmospheric state $u_n$ on a $0.25^\circ$ grid at a time $t_n$, it predicts the state $u_{n+1} = F_{\theta}(u_n, t_n, z_n)$ at the next time step $t_{n+1}$, 6 hours into the future. Stochasticity is introduced through a hidden Markov model approach, where the model takes an extra conditioning input $z_n$ - a random noise vector drawn from a number of spherical diffusion processes with different length- and timescales \cite{Palmer2009}. \Cref{fig:fcn3_schematic} depicts this setup, and a detailed description is found in \Cref{sec:problem_statement}.

The parameters $\theta$ of the model $ F_{\theta}$ are optimized with the aim of accurately approximating atmospheric processes and matching the observed spatio-temporal distributions of physical variables. FCN3 uses an end-to-end ensemble training approach, minimizing a composite probabilistic loss function \eqref{eq:combined_loss} based on the continuously ranked probability score (CRPS) \eqref{eq:fair_spread_skill_crps}. This objective compares the predictive ensemble of marginals to ground-truth observations.
Although training with the CRPS objective has been shown to produce models with high predictive skill, these models have not generated ensemble members with physically accurate spectra that  correctly capture spatial correlations \cite{Lessig2023,Kochkov2023,Lang2024}.
Although the scalar-valued Continuous Ranked Probability Score (CRPS) \eqref{eq:crps} is a proper scoring rule -- meaning it is uniquely minimized when the predictive distribution matches the target distribution -- this property does not extend to summary scores that aggregate individual CRPS values across marginals, as is commonly done when forecasting spatial or multivariate variables.
This is particularly problematic for multi-variate spatial processes, where the CRPS can be minimized in a point-wise manner by an unphysical ensemble. To address this issue, we combine the spatial, point-wise CRPS loss term with a loss term in the spectral domain. A similar approach using a low-pass filtered spectral loss term has previously been adopted by \citet{Kochkov2023}, but failed to accurately capture the high-frequency behavior of the underlying processes. Our approach weights spectral coefficients according to their multiplicity and enforces a good match of the their distributions across all wavelengths.
A detailed discussion of the objective function and its motivation are provided in \Cref{sec:objective_function}.

\section{Spherical neural operator architecture}
\label{sec:spherical_neural_operator}

Although a combined spectral and spatial probabilistic loss function encourages the learned operator to be accurately represented across scales, the concrete parameterization is equally important in determining the space of learnable operators and therefore their properties. As such, we choose a geometric approach grounded in signal processing principles and symmetry considerations:

FCN3 is a spherical neural operator architecture and relies heavily on local and global spherical group convolutions. More precisely, global convolution filters are parameterized in the spectral domain by leveraging the convolution theorem on the sphere and the associated spherical harmonic transform (SHT) \cite{Bonev2023}. This approach resembles classical pseudo-spectral methods such as IFS, which compute the PDE operator in the spectral domain. Additionally, we employ spherical group convolutions\footnote{Group convolutions are convolutions formulated w.r.t. a symmetry group. For the two-dimensional sphere, this is the rotation group of three-dimensional rotations $SO(3)$.} with learnable, locally supported kernels. This is implemented using the framework for discrete-continuous (DISCO) convolutions on the sphere~\cite{Ocampo2022,Liu2024}, which formulate the convolution in the continuous domain and approximate the integral with a quadrature rule. This formulation enables anisotropic filters that are better suited to approximate atmospheric phenomena such as adiabatic flow confined to vertically tilted isentropes with characteristic morphology, or blocked flow around topographic features. The localized convolutional approach also resembles finite differencing - another building block encountered in most classical NWP models.

Building on these convolutional principles, the overall FCN3 architecture is organized into three main components: an encoder, a processor composed of several spherical neural operator blocks, and a decoder (see \Cref{fig:fcn3_schematic}). These blocks adopt the structure of the popular ConvNeXt architecture~\cite{Liu2022}, which contain a convolution, a GeLU activation function \cite{Hendrycks2016}, a point-wise multi-layer perceptron (MLP) and a skip connection. We deliberately omit layer normalization, motivated by the importance of absolute magnitudes in physical processes. The convolution filters are either parameterized in the spectral domain or as approximately spherically equivariant local convolutions \cite{Bonev2023,Liu2024}. In the latter case, we choose smooth localized filter functions, parameterized by linear combinations of Morlet wavelets on a disk. Through experimentation, we find that a ratio of four local blocks to one global block yields the best forecast skill. The encoder layer is  comprised of a single local spherical convolutions and down-samples the $721\times1440$ input/output signals to a latent representation on a $360\times720$ Gaussian grid with an embedding dimension of $641$. The decoder uses a combination of bilinear spherical interpolation and local spherical convolution to up-sample the latent signals to the native resolution while mitigating aliasing errors. Both encoder and decoder encode do not perform any channel mixing and instead encode input signals separately, to avoid the mixing of signals with vastly different spectral properties. Finally, water channels are passed through a smooth, spline-based output activation function which constrains them to positive values, while reducing the amount of high-frequency noise introduced through the non-linearity.

In contrast to most ML weather models which predict tendencies, i.e. the difference between the prediction and the input, FCN3 predicts the next state directly. Empirically, we find that this approach works better in avoiding the build-up of high-frequency artifacts. Moreover, predicting tendencies may be interpreted as restricting the model to Euler time-stepping, which may adversely affect the space of learnable operators \cite{Lang2024}. A detailed account of signal-processing considerations on the sphere and our filter parameterizations is provided in \Cref{sec:signal_processing}. Furthermore, architectural choices and hyperparameters are discussed in detail in \Cref{sec:model_architecture}.

\section{Scalable training through hybrid parallelism}
\label{sec:scalable_training}

Training models with large internal representations such as FCN3 requires more memory than what is available on a single GPU for their forward and backward passes. This memory requirement is further exacerbated by autoregressive rollouts, where multiple forward and backward passes need to be fit into GPU memory. These considerations limit the size of the model to the memory available per GPU, and thus set the maximum scale for most models. While some models such as GraphCast use gradient checkpoint to enable the memory-intensive training \cite{Lam2022}, this comes with the significant downside of trading memory for compute, increasing already long iteration times further in training.

By distributing models across multiple GPUs, model parallelism offers an alternative path for practitioners to reduce the memory requirements and train much bigger models. This approach greatly improved the fidelity and performance of modern ML models \cite{DBLP:journals/corr/abs-2004-13336, deepspeed2020, zero2020} and is the foundation of the success of current large language models (LLM) such as ChatGPT 4~\cite{openai2024gpt4technicalreport}, Llama 3~\cite{grattafiori2024llama3herdmodels}, and others. Neural networks generally scale well with available data and oftentimes, training larger models comes with an increase in skill, as long as more training data is available \cite{Kaplan2020}. This creates a unique challenge for scientific ML methods, where the training data is often high-dimensional, in comparison to language modeling or computer vision tasks. In the case of FCN3, a typical sample at $0.25^\circ$ resolution consists of $721\times1440$ floating points per variable, and multiple tens of variables are normally used for skillful predictions. This renders ML driven weather prediction considerably more data-intensive than many other ML tasks.

\begin{figure}[htbp]
    \centering
    \includegraphics[width=1.0\textwidth]{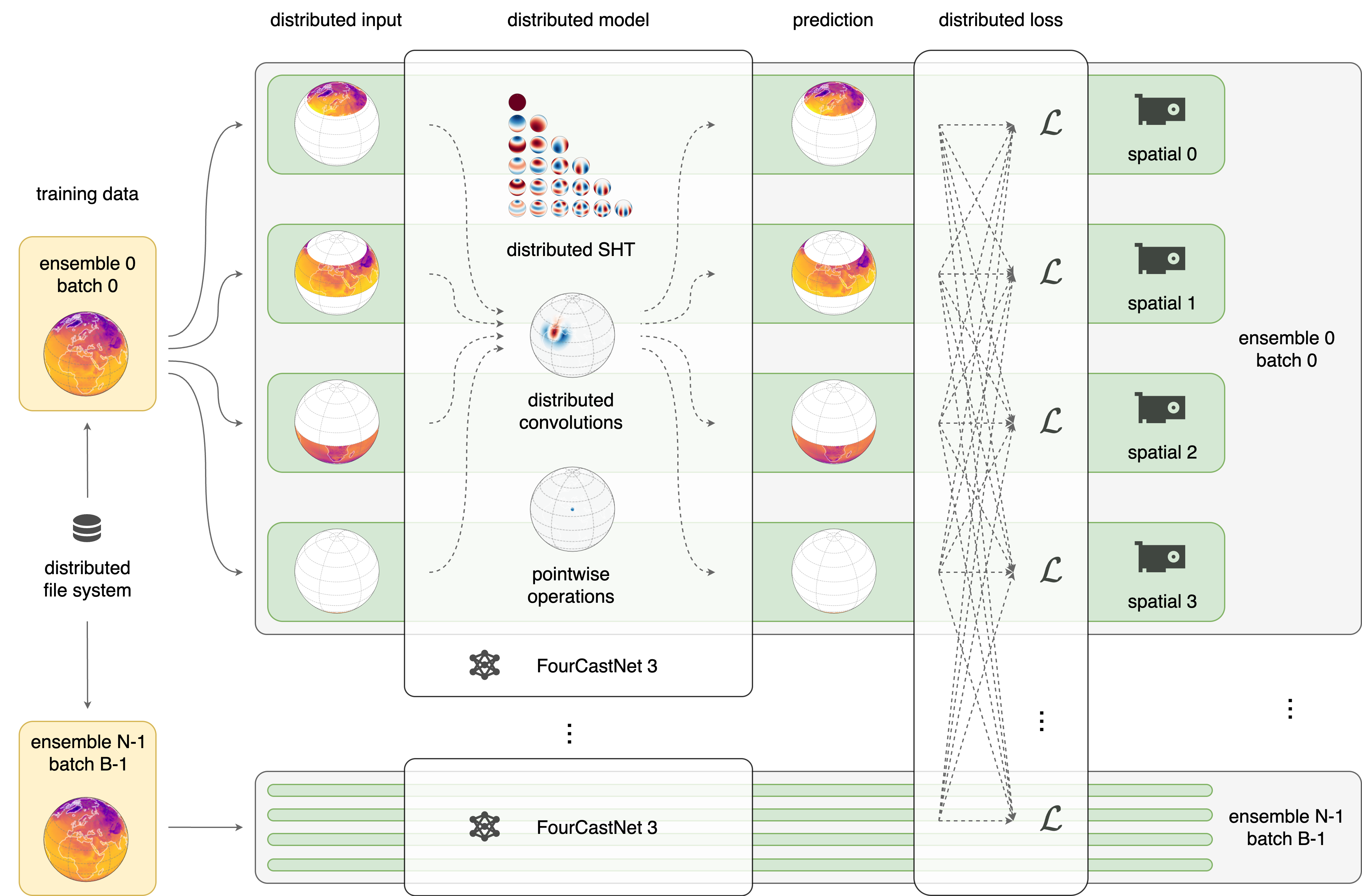}
    \caption{Illustration of model- and data-parallelism for training of FourCastNet 3. In the given example, the input data is spatially distributed across four ranks (green boxes) by splitting it across the latitude. This reduces the memory footprint of the input, prediction and activations within the network. The training data is read in a sharded fashion from the distributed file system, simultaneously lowering the required I/O per rank. This domain-decomposition requires the model and it's weights to be distributed; spherical harmonic transforms and discrete-continuous convolutions are distributed and split across the four ranks. In addition to the spatial model-parallelism, data-parallelism is utilized to distribute individual ensemble members and batch samples (grey boxes). Finally, the ensemble loss for a single sample is computed by taking the entire ensemble information across ensemble parallel ranks and spatial parallel ranks. On top of this, batch parallelism is utilized (not illustrated in this figure).}
    \label{fig:fcn3_model_parallel}
\end{figure}

Model parallelism is inspired by classical numerical methods, where not only the model and weights are split across ranks, but also the data which the model processes. Model parallelism is typically achieved through feature-space parallelism, i.e. by splitting the feature maps across multiple GPU. This approach is heavily used in modern distributed LLMs, alongside other parallelism paradigms such as pipeline- and traditional batch-parallelism. To enable the training of FCN3, we implement spatial model parallelism (also referred to as domain parallelism), where both the model and data are split across ranks by employing a spatial domain decomposition. This approach is inspired by traditional distributed scientific computing applications and requires the implementation of distributed variants of all spatial algorithms (see \Cref{fig:fcn3_model_parallel}).
Besides these two approaches as well as traditional batch parallelism, another approach is to split members of the same forecasting ensemble across multiple GPU. This variant of data parallelism is highly efficient because different ensemble members are computationally independent until the loss computation, which usually requires some communication across the ensemble group of GPUs.

The training of FCN3 requires spatial model parallelism via domain decomposition as well as ensemble and batch parallelism. We will refer to the former as model and the latter two as data parallelism. We have implemented all of these features in Makani, a framework for large-scale distributed training of ML based weather models. For a more detailed description of parallelization features, cf. section \ref{sec:parallelism}.

This paradigm enables us to train large principled models by scaling training to thousands of GPUs and more. FCN3 is trained on historic atmospheric ERA5 reanalysis data ranging from 1980 to 2016. ERA5 is a multi-decadal, self-consistent record and represents our best understanding of Earth's atmospheric system~\cite{Hersbach2020}. Training is split into stages, thereby forming a curriculum training approach.
The initial pre-training phase focuses on the model’s 6-hourly prediction skill, by utilizing all hourly samples from the ERA5 training dataset, constructing 6 hour lead time input-target-pairs that start at each discrete UTC hour.
The model is trained for 208,320 gradient descent steps on this dataset with a batch size of 16 and an ensemble size of 16. This initial training stage was carried out on 1024 NVIDIA H100 on the NVIDIA Eos Supercomputer for a total of 78 hours. In the second pre-training phase, the model is trained on 6-hourly initial conditions using 4 autoregressive rollout steps. This is performed for 5,040 steps while lowering the learning rate every 840 steps. The second pre-training stage took 15 hours on 512 NVIDIA A100 GPUs to complete and was carried out on the NERSC Perlmutter system. The final is fine-tuned on 6-hourly samples ranging from 2012 to 2016 to account for potential drifts in the distribution and improve performance on data that lie in the near- to medium-term future.  This final stage is carried out on 256 NVIDIA H100 GPUs on the Eos system and took 8 hours to complete. As a single model instance does not fit on a 80Gb VRAM GPU, we leverage the previously described spatial parallelism, splitting the data and the model. This ranges from a 4-fold split in pretraining to a 16-fold split during finetuning, due to the increased memory requirements from autoregressive training. Details of the training methodology and setup are outlined in \Cref{sec:training}.
\section{Results}
\label{sec:main_results}

\begin{figure}[tb]
    \centering
    \includegraphics[width=\textwidth]{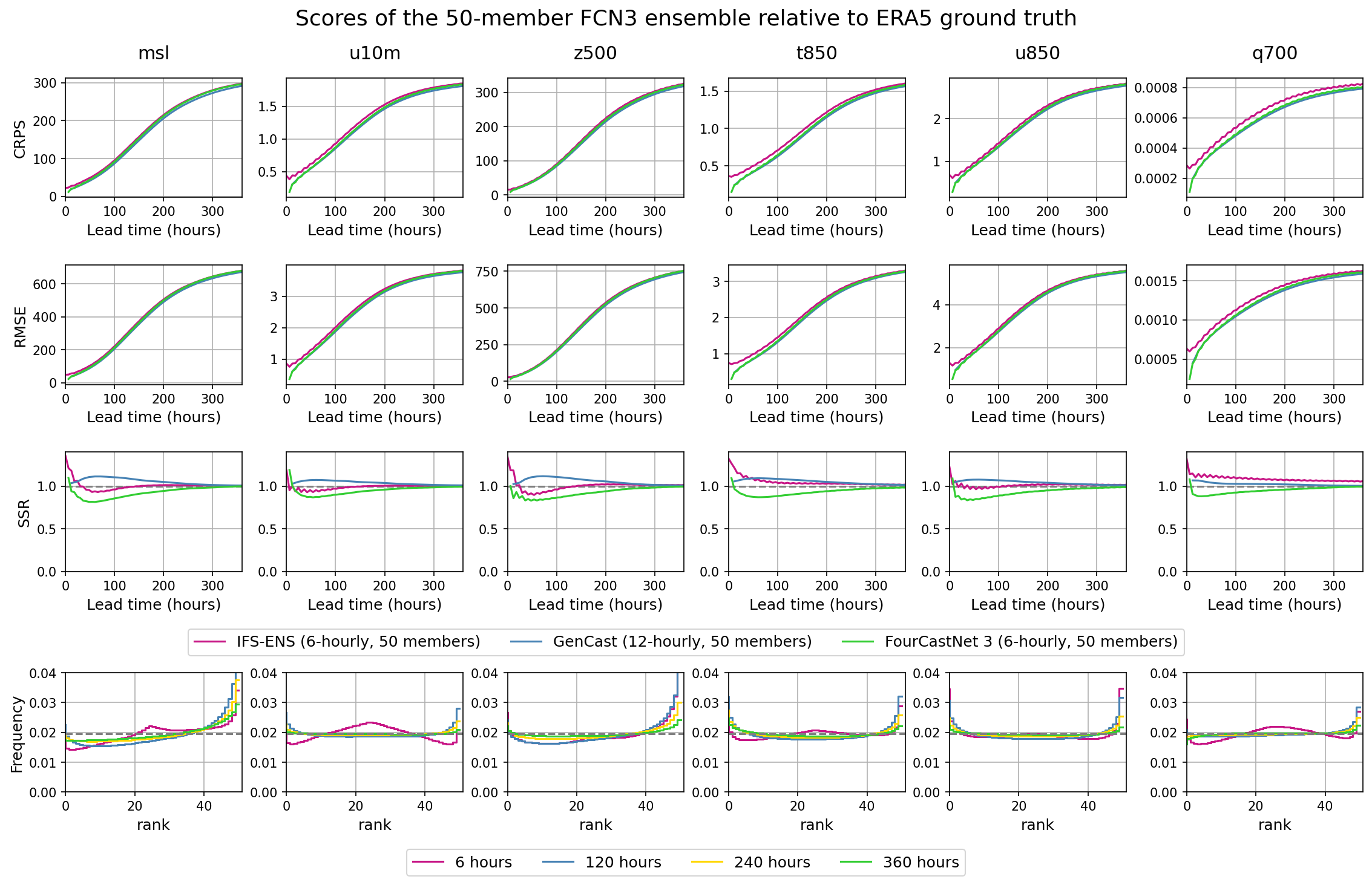}
    \caption{Probabilistic skill of FourCastNet 3 relative to the ERA5 ground truth. Continuously ranked probabilistic scores (lower is better), ensemble mean RMSE (lower is better), spread-skill ratios (closer to one is better) and rank-histograms (more uniform is better) are reported from top to bottom. The scores are computed over 12-hourly initial conditions ranging from 2020-01-01 00:00:00 UTC to 2020-12-31 23:59:00 UTC.}
    \label{fig:main_scores}
\end{figure}

Key performance scores of FCN3 such as continuously ranked probability score (CRPS) and ensemble-mean RMSE are averaged over 12-hourly initial conditions in the out-of-sample year 2020 and reported in \Cref{fig:main_scores}. FCN3 beats the gold-standard physics-based NWP model IFS-ENS by a margin that is virtually indistinguishable from GenCast, the state-of-the-art data-driven weather model. Our approach enables direct, one-step generation of ensemble members and can generate a single 15-day forecast at a temporal resolution of 6 hours and a spatial resolution of $0.25^\circ$ in a matter of 60 seconds on a single NVIDIA H100 GPU. In comparison, a 15-day forecast of GenCast takes 8 minutes on a Cloud TPU v5 instance (at half the temporal resolution) \cite{Price2024}, and an IFS forecast takes about one hour on 96 AMD Epyc Rome CPUs (at 9km operational resolution) \cite{Alexe2024}. Barring the differences in hardware and resolution, this constitutes a speed-up of $\sim$8x over GenCast and a speed-up of $\sim$60x over IFS-ENS.

Crucially, the 50-member FCN3 ensemble forecast is well-calibrated with spread-skill ratios approaching 1, indicating interchangeability between observations and ensemble members in the forecast. This is confirmed via rank-histograms, which report the frequencies of the ordinal ranks of the observation within the predictive ensemble. The temporal evolution of the rank histograms closely mirrors the spread-skill ratios, indicating a slightly over-dispersive ensemble at short lead times of up to 24 hours, which then becomes under-dispersive and then gradually relaxes to a flat rank-histogram. These results are especially encouraging, given that the evaluated 50 member ensemble is larger than the 16 ensemble members used in training, indicating that even larger ensembles are justifiable to test during inference.
\begin{figure}[tb]
    \centering
    \includegraphics[width=\textwidth]{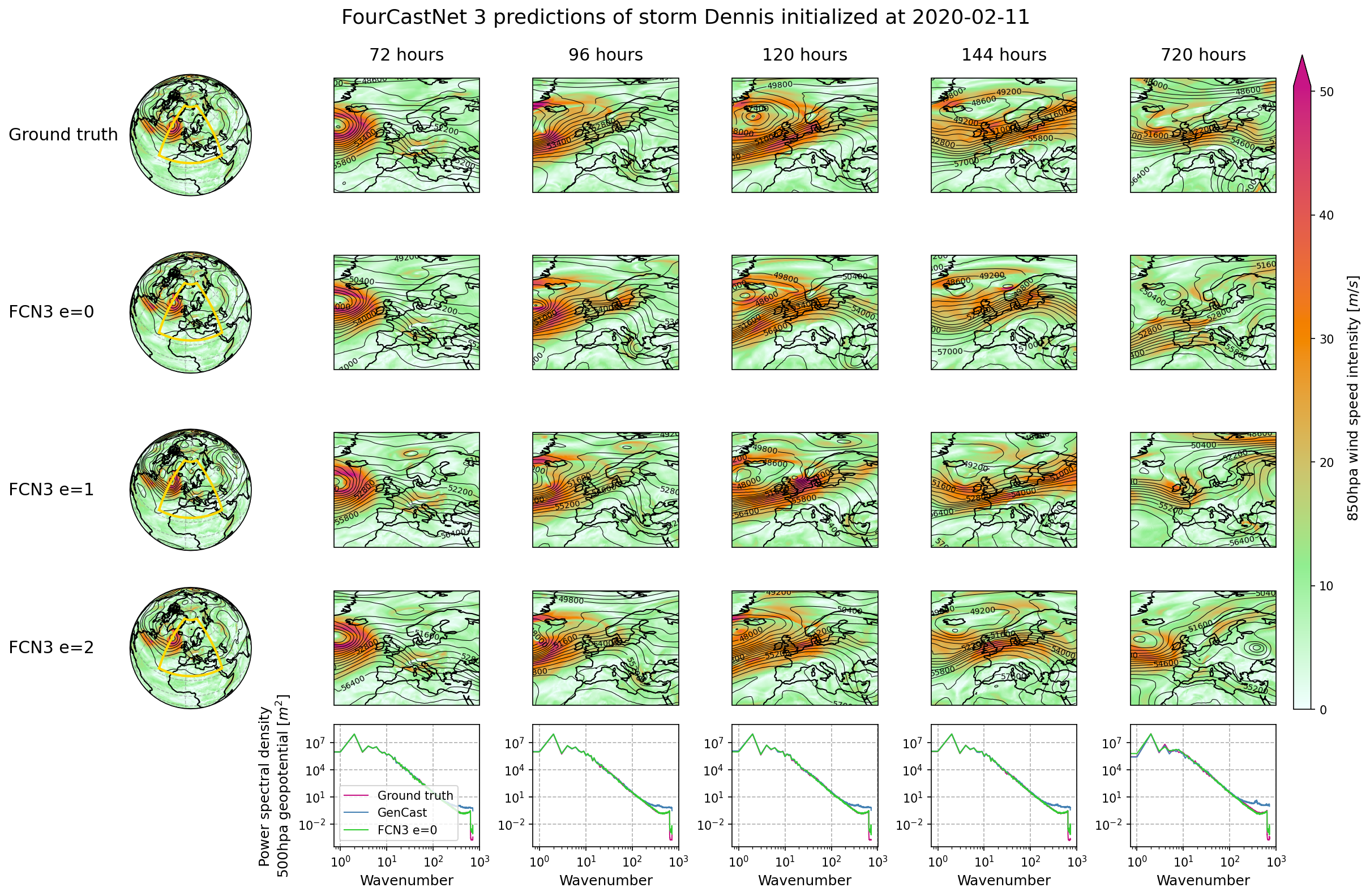}
    \caption{FourCastNet 3 prediction of storm Dennis initialized on 2020-02-11 at 00:00:00 UTC. The plot depicts wind-speeds at a pressure level of 850hPa and isohypses (height contours) of the 500hPa geopotential height. FCN3 accurately predicts the storm and its landfall 5 days in advance, with different ensemble members depicting different scenarios. FCN3 skillfully predicts global weather phenomena at a spatial resolution of $0.25^\circ$ and a temporal resolution of 6 hours. FCN3 exhibits exceptionally accurate and stable spectra even after extended rollouts of 30 days (720 hours) and more.}
    \label{fig:storm_dennis}
\end{figure}

It is important to investigate case studies, since scores such as ensemble-mean RMSE and CRPS are incomplete metrics that alone do not provide a comprehensive view of a probabilistic weather forecast. For example, the CRPS score only evaluates the accuracy of the predictive distribution point-wise and does not take tempo-spatial correlations into account. As such, a perfect forecast from the ground-truth distribution, which is scrambled by shuffling the ensemble members at each point will result in unphysical predictions yet still retain the optimal CRPS score. Similarly, RMSE scores can be easily improved by blurring forecasts, rendering them useless for all practical purposes. A key challenge in data-driven weather models is to reproduce the physical fidelity of traditional NWP models and reduce possible spurious correlations that stem from the data-driven approach.

\Cref{fig:storm_dennis} examines a case study showing wind intensities at 850hPa and geopotential height at 500hPa of a FCN3 forecast initialized on 2020-02-11 at 00:00:00 UTC, 48 hours before the extra-tropical storm Dennis made its landfall over Ireland and the British Isles. The close-up plots in \Cref{fig:storm_dennis} indicate that FCN3 is capable of faithfully simulating this event, reproducing both realistic wind intensities and appropriate co-variation of flow with the pressure field. This is confirmed in the angular power spectral density (PSD) of the 500hPa geopotential height, reported in the bottom row. FCN3 retains perfectly the correct slopes in the power spectra, a desirable property towards better ML weather models with high physical fidelity. Even at long lead times of 30 days, we observe no apparent degradation of the angular power spectra and predictions retain their effective resolution, remaining sharp even at long lead times.

\begin{figure}[tb]
    \centering
    \includegraphics[width=\textwidth]{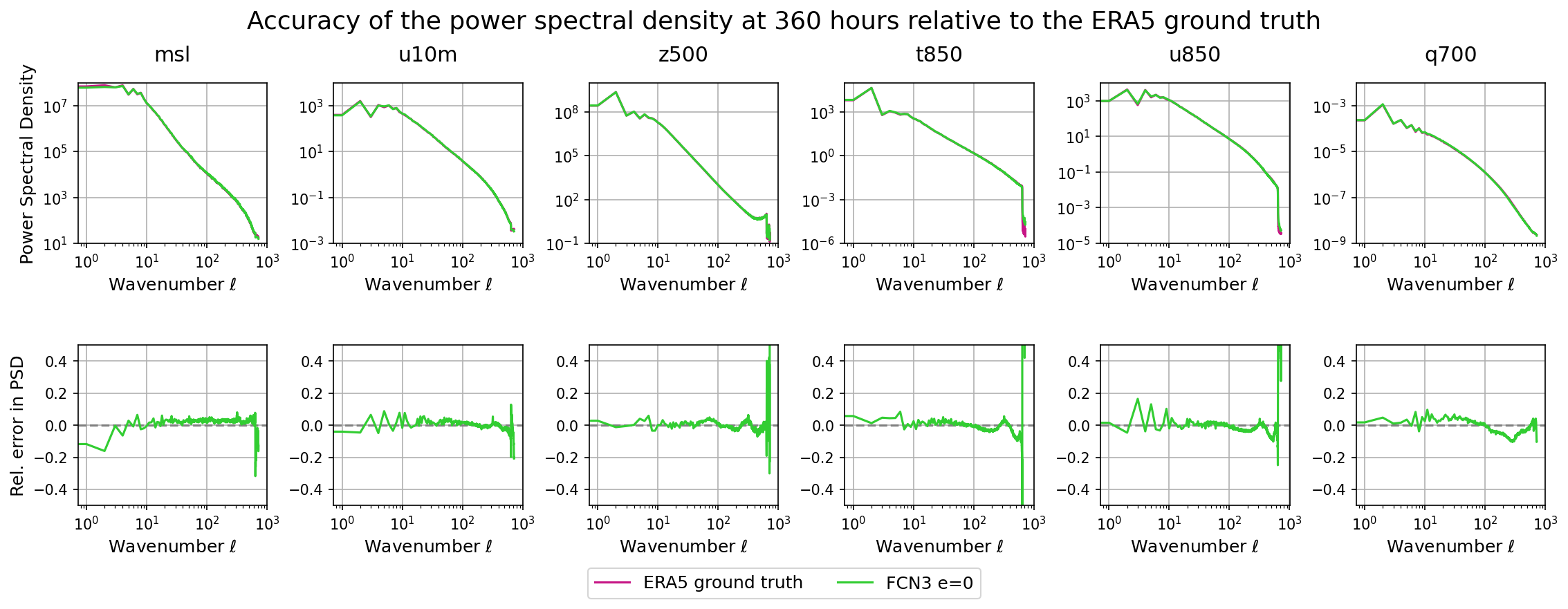}
    \caption{Comparison of angular power spectral densities of a single FourCastNet 3 ensemble member at a lead time of 360 hours to the ERA5 ground truth. Power spectral densities are averaged over 12-hourly initial conditions ranging from 2020-01-01 00:00:00 UTC to 2020-12-31 23:59:00 UTC.}
    \label{fig:psd_accuracy}
\end{figure}

The spectral fidelity of FCN3 is also observed in \Cref{fig:psd_accuracy}, which depicts power spectral densities averaged over the entire evaluation year of 2020 and the respective relative error w.r.t. the angular power spectrum of the ERA5 ground truth. Even at high wavenumbers, we observe that the relative error remains bounded with deviations ranging from $-0.2$ to $0.2$.

We postulate that the spectral properties are a result of our careful architectural design choices, which reflect geometrical and signal-processing principles, and the combined CRPS loss function which enforces the correct local and global distribution, thus encouraging the model to learn the correct spatial correlations.
Competing, deterministic ML weather models typically display a decay of high-frequency information, which appears as blurring. Even the CRPS-trained hybrid weather model NeuralGCM shows significant blurring in high-frequency modes. Moreover, newer, probabilistic ML weather models such as GenCast \cite{Price2024} and AIFS-CRPS \cite{Lang2024} cannot faithfully retain the correct spectral signatures and show a build-up of high-frequency modes, as illustrated in \Cref{fig:storm_dennis}. In traditional NWP models, such build-ups can be a precursor to an imminent blow-up \cite{Lauritzen2011}. As such, this constitutes a major milestone towards physically faithful data-driven, probabilistic weather models, which can be efficiently evaluated even at longer lead times.

Additional evaluation of FCN3, angular and zonal power spectral densities, alongside physical consistency tests, are provided in \Cref{sec:results}. The detailed evaluation confirms that FCN3 is a probabilistically skillful, computationally efficient global weather model, with unprecedented spectral fidelity and a high degree of physical realism. Forecasts remain stable well into the subseasonal range of 60 days, thus paving the way toward subseasonal forecasts and large ensembles at these lead times.

\section{Conclusions}
\label{sec:conclusions}

We present FourCastNet 3 (FCN3), a novel probabilistic weather forecasting model that leverages spherical signal processing and a hidden-Markov ensemble formulation, trained end-to-end with a probabilistic objective in both spectral and spatial domains. FCN3 achieves skillful and computationally efficient forecasts, outperforming traditional numerical weather prediction (NWP) methods and matching the performance of state-of-the-art diffusion models at a fraction of the computational cost. This is accomplished using a purely convolutional architecture based on spherical group convolutions, in contrast to the prevailing transformer-based approaches. Notably, FCN3 generates physically realistic spectra across all wavelengths up to the cutoff in the training data, avoiding the overly smooth or spurious high-frequency artifacts that challenge other machine learning models. This fidelity enables stable, sharp forecasts even at extended lead times of up to 60 days, positioning FCN3 as a promising tool for subseasonal prediction with large ensembles.

FCN3 introduces major computational and practical improvements that make large-scale, high-resolution ensemble forecasting more accessible than ever. Its massively parallel training workflows, model and ensemble parallelism, and low inference cost enable rapid, efficient production of large ensemble forecasts. In-situ diagnostics and scoring can be performed during model execution, eliminating the need to store terabytes of data and removing storage and I/O bottlenecks that have historically limited ensemble analysis. All key components, including training and inference code, are fully open-source, providing the research community with transparent, reproducible tools for both operational and experimental weather prediction.

As an ensemble model, FCN3 enables detailed exploration of multiple plausible future weather scenarios from a single initialization, making it a powerful tool for studying atmospheric dynamics, predictability, and the statistics of
low-probability, high-impact events. Looking ahead, we plan to extend FCN3 to include precipitation as a diagnostic output and to integrate data assimilation uncertainty, further broadening its applicability and impact. Together, these innovations position FCN3 as a robust, efficient, and extensible foundation for next-generation probabilistic weather forecasting and atmospheric science research.

\section*{Data and materials availability}
\label{sec:data_and_materials_availability}
FourCastNet 3's training code is available in \texttt{Makani}, a training framework used for scale training of ML weather models to 1000s of GPUs. It is openly available at \url{https://github.com/NVIDIA/makani} under the Apache License 2.0. The ERA5 training data is openly available at \url{https://cds.climate.copernicus.eu/datasets/reanalysis-era5-single-levels}. Finally, \texttt{torch-harmonics}, our library for machine-learning and differentiable signal processing on the sphere, is openly available at \url{https://github.com/NVIDIA/torch-harmonics} under the BSD-3-Clause license. 

\section*{Acknowledgements}

This research was supported by NVIDIA as well as the Director, Office of Science, Office of Biological and Environmental Research of the
U.S. Department of Energy under Contract No. DE-AC02-05CH11231 and by the Regional and Global Model Analysis Program area within
the Earth and Environmental Systems Modeling Program. The research used resources of the National Energy Research Scientific Computing
Center (NERSC), also supported by the Office of Science of the U.S. Department of Energy, under Contract No. DE-AC02-05CH11231.
The computation for this paper was supported in part by the DOE Advanced Scientific Computing Research (ASCR) Leadership Computing
Challenge (ALCC) 2024-2025 award ‘Huge Ensembles of Weather Extremes using the Fourier Forecasting Neural Network’ to William
Collins (LBNL).

We are grateful to our colleagues Dmitry Alexeev, Noah Brenowitz, Alberto Carpentieri, Dale Durran, Dallas Foster, Peter Harrington, Jan Kautz, Marius Koch, Jussi Leinonen, Morteza Mardani, Steve Marschner, Peter Messmer, Thomas M\"uller, Merlin Nimier-David, Andrea Paris, Jaideep Pathak, Suman Ravuri, Ira Shokar, Shashank Subramanian for helpful and encouraging discussions, as well as feedback on our work.

We thank the European Centre of Medium-Range Weather Forecasting (ECMWF) for publishing the ERA5 dataset and enabling this line of research.

\bibliographystyle{unsrttrunc}
\bibliography{references_bbonev, references_tkurth, references_amahesh}

\begin{thebibliography}{72}
\providecommand{\natexlab}[1]{#1}
\providecommand{\url}[1]{\texttt{#1}}
\expandafter\ifx\csname urlstyle\endcsname\relax
  \providecommand{\doi}[1]{doi: #1}\else
  \providecommand{\doi}{doi: \begingroup \urlstyle{rm}\Url}\fi

\bibitem[Lynch(2008)]{Lynch2008}
Peter Lynch.
\newblock {The origins of computer weather prediction and climate modeling}.
\newblock \emph{Journal of Computational Physics}, 227:\penalty0 3431--3444, 2008.
\newblock ISSN 0021-9991.
\newblock \doi{https://doi.org/10.1016/j.jcp.2007.02.034}.
\newblock URL \url{https://www.sciencedirect.com/science/article/pii/S0021999107000952}.
\newblock Predicting weather, climate and extreme events.

\bibitem[Pathak et~al.(2022)Pathak, Subramanian, Harrington, Raja, Chattopadhyay, Mardani, Kurth, Hall, Li, Azizzadenesheli, Hassanzadeh, Kashinath, and Anandkumar]{Pathak2022}
Jaideep Pathak, Shashank Subramanian, Peter Harrington, Sanjeev Raja, Ashesh Chattopadhyay, Morteza Mardani, Thorsten Kurth, David Hall, Zongyi Li, Kamyar Azizzadenesheli, Pedram Hassanzadeh, Karthik Kashinath, and Animashree Anandkumar.
\newblock {FourCastNet: A Global Data-driven High-resolution Weather Model using Adaptive Fourier Neural Operators}, 2 2022.
\newblock URL \url{http://arxiv.org/abs/2202.11214}.

\bibitem[Bi et~al.(2022)Bi, Xie, Zhang, Chen, Gu, and Tian]{Bi2022}
Kaifeng Bi, Lingxi Xie, Hengheng Zhang, Xin Chen, Xiaotao Gu, and Qi~Tian.
\newblock {Pangu-Weather: A 3D High-Resolution Model for Fast and Accurate Global Weather Forecast}, 11 2022.
\newblock URL \url{http://arxiv.org/abs/2211.02556}.

\bibitem[Lam et~al.(2022)Lam, Sanchez-Gonzalez, Willson, Wirnsberger, Fortunato, Pritzel, Ravuri, Ewalds, Alet, Eaton-Rosen, Hu, Merose, Hoyer, Holland, Stott, Vinyals, Mohamed, and Battaglia]{Lam2022}
Remi Lam, Alvaro Sanchez-Gonzalez, Matthew Willson, Peter Wirnsberger, Meire Fortunato, Alexander Pritzel, Suman Ravuri, Timo Ewalds, Ferran Alet, Zach Eaton-Rosen, Weihua Hu, Alexander Merose, Stephan Hoyer, George Holland, Jacklynn Stott, et~al.
\newblock {GraphCast: Learning skillful medium-range global weather forecasting}, 12 2022.
\newblock URL \url{http://arxiv.org/abs/2212.12794}.

\bibitem[Kurth et~al.(2022)Kurth, Subramanian, Harrington, Pathak, Mardani, Hall, Miele, Kashinath, and Anandkumar]{kurth2022fourcastnetacceleratingglobalhighresolution}
Thorsten Kurth, Shashank Subramanian, Peter Harrington, Jaideep Pathak, Morteza Mardani, David Hall, Andrea Miele, Karthik Kashinath, and Animashree Anandkumar.
\newblock {FourCastNet: Accelerating Global High-Resolution Weather Forecasting using Adaptive Fourier Neural Operators}, 2022.
\newblock URL \url{https://arxiv.org/abs/2208.05419}.

\bibitem[Price et~al.(2024)Price, Sanchez-Gonzalez, Alet, Andersson, El-Kadi, Masters, Ewalds, Stott, Mohamed, Battaglia, Lam, and Willson]{Price2024}
Ilan Price, Alvaro Sanchez-Gonzalez, Ferran Alet, Tom~R. Andersson, Andrew El-Kadi, Dominic Masters, Timo Ewalds, Jacklynn Stott, Shakir Mohamed, Peter Battaglia, Remi Lam, and Matthew Willson.
\newblock {Probabilistic weather forecasting with machine learning}.
\newblock \emph{Nature}, 1 2024.
\newblock ISSN 14764687.
\newblock \doi{10.1038/s41586-024-08252-9}.

\bibitem[Mahesh et~al.(2024{\natexlab{a}})Mahesh, Collins, Bonev, Brenowitz, Cohen, Elms, Harrington, Kashinath, Kurth, North, OBrien, Pritchard, Pruitt, Risser, Subramanian, and Willard]{Mahesh2024a}
Ankur Mahesh, William Collins, Boris Bonev, Noah Brenowitz, Yair Cohen, Joshua Elms, Peter Harrington, Karthik Kashinath, Thorsten Kurth, Joshua North, Travis OBrien, Michael Pritchard, David Pruitt, Mark Risser, Shashank Subramanian, and Jared Willard.
\newblock {Huge Ensembles Part I: Design of Ensemble Weather Forecasts using Spherical Fourier Neural Operators}, 8 2024{\natexlab{a}}.
\newblock URL \url{http://arxiv.org/abs/2408.03100}.

\bibitem[Mahesh et~al.(2024{\natexlab{b}})Mahesh, Collins, Bonev, Brenowitz, Cohen, Harrington, Kashinath, Kurth, North, OBrien, Pritchard, Pruitt, Risser, Subramanian, and Willard]{Mahesh2024b}
Ankur Mahesh, William Collins, Boris Bonev, Noah Brenowitz, Yair Cohen, Peter Harrington, Karthik Kashinath, Thorsten Kurth, Joshua North, Travis OBrien, Michael Pritchard, David Pruitt, Mark Risser, Shashank Subramanian, and Jared Willard.
\newblock {Huge Ensembles Part II: Properties of a Huge Ensemble of Hindcasts Generated with Spherical Fourier Neural Operators}, 8 2024{\natexlab{b}}.
\newblock URL \url{http://arxiv.org/abs/2408.01581}.

\bibitem[Weyn et~al.(2021)Weyn, Durran, Caruana, and Cresswell-Clay]{Weyn2021}
Jonathan~A. Weyn, Dale~R. Durran, Rich Caruana, and Nathaniel Cresswell-Clay.
\newblock {Sub-seasonal forecasting with a large ensemble of deep-learning weather prediction models}, 2 2021.
\newblock URL \url{http://arxiv.org/abs/2102.05107 http://dx.doi.org/10.1029/2021MS002502}.

\bibitem[Bonev et~al.(2023)Bonev, Kurth, Hundt, Pathak, Baust, Kashinath, and Anandkumar]{Bonev2023}
Boris Bonev, Thorsten Kurth, Christian Hundt, Jaideep Pathak, Maximilian Baust, Karthik Kashinath, and Anima Anandkumar.
\newblock {Spherical Fourier Neural Operators: Learning Stable Dynamics on the Sphere}.
\newblock \emph{Proceedings of the 40th International Conference on Machine Learning}, 202:\penalty0 2806--2823, 6 2023.
\newblock URL \url{http://arxiv.org/abs/2306.03838}.

\bibitem[Watt-Meyer et~al.(2023)Watt-Meyer, Dresdner, McGibbon, Clark, Henn, Duncan, Brenowitz, Kashinath, Pritchard, Bonev, Peters, and Bretherton]{Watt-Meyer2023}
Oliver Watt-Meyer, Gideon Dresdner, Jeremy McGibbon, Spencer~K. Clark, Brian Henn, James Duncan, Noah~D. Brenowitz, Karthik Kashinath, Michael~S. Pritchard, Boris Bonev, Matthew~E. Peters, and Christopher~S. Bretherton.
\newblock {ACE: A fast, skillful learned global atmospheric model for climate prediction}.
\newblock 10 2023.
\newblock URL \url{http://arxiv.org/abs/2310.02074}.

\bibitem[Karlbauer et~al.(2024)Karlbauer, Cresswell-Clay, Durran, Moreno, Kurth, Bonev, Brenowitz, and Butz]{Karlbauer2024}
Matthias Karlbauer, Nathaniel Cresswell-Clay, Dale~R. Durran, Raul~A. Moreno, Thorsten Kurth, Boris Bonev, Noah Brenowitz, and Martin~V. Butz.
\newblock {Advancing Parsimonious Deep Learning Weather Prediction Using the HEALPix Mesh}.
\newblock \emph{Journal of Advances in Modeling Earth Systems}, 16, 8 2024.
\newblock ISSN 19422466.
\newblock \doi{10.1029/2023MS004021}.

\bibitem[Guan et~al.(2025)Guan, Arcomano, Chattopadhyay, and Maulik]{Guan2025}
Haiwen Guan, Troy Arcomano, Ashesh Chattopadhyay, and Romit Maulik.
\newblock Lucie: A lightweight uncoupled climate emulator with long-term stability and physical consistency for o(1000)-member ensembles, 4 2025.
\newblock URL \url{http://arxiv.org/abs/2405.16297}.

\bibitem[Cresswell-Clay et~al.(2025)Cresswell-Clay, Liu, Durran, Liu, Espinosa, Moreno, and Karlbauer]{Cresswell-Clay2025}
Nathaniel Cresswell-Clay, Bowen Liu, Dale Durran, Zihui Liu, Zachary~I. Espinosa, Raul Moreno, and Matthias Karlbauer.
\newblock A deep learning earth system model for efficient simulation of the observed climate, 2 2025.
\newblock URL \url{http://arxiv.org/abs/2409.16247}.

\bibitem[Fang and Mengaldo(2025)]{Fang2025}
Zhou Fang and Gianmarco Mengaldo.
\newblock Dynamical errors in machine learning forecasts, 4 2025.
\newblock URL \url{http://arxiv.org/abs/2504.11074}.

\bibitem[Kochkov et~al.(2023)Kochkov, Yuval, Langmore, Norgaard, Smith, Mooers, Klöwer, Lottes, Rasp, Düben, Hatfield, Battaglia, Sanchez-Gonzalez, Willson, Brenner, and Hoyer]{Kochkov2023}
Dmitrii Kochkov, Janni Yuval, Ian Langmore, Peter Norgaard, Jamie Smith, Griffin Mooers, Milan Klöwer, James Lottes, Stephan Rasp, Peter Düben, Sam Hatfield, Peter Battaglia, Alvaro Sanchez-Gonzalez, Matthew Willson, Michael~P. Brenner, and Stephan Hoyer.
\newblock {Neural General Circulation Models for Weather and Climate}, 11 2023.
\newblock URL \url{http://arxiv.org/abs/2311.07222 http://dx.doi.org/10.1038/s41586-024-07744-y}.

\bibitem[Rasp et~al.(2023)Rasp, Hoyer, Merose, Langmore, Battaglia, Russel, Sanchez-Gonzalez, Yang, Carver, Agrawal, Chantry, Bouallegue, Dueben, Bromberg, Sisk, Barrington, Bell, and Sha]{Rasp2023}
Stephan Rasp, Stephan Hoyer, Alexander Merose, Ian Langmore, Peter Battaglia, Tyler Russel, Alvaro Sanchez-Gonzalez, Vivian Yang, Rob Carver, Shreya Agrawal, Matthew Chantry, Zied~Ben Bouallegue, Peter Dueben, Carla Bromberg, Jared Sisk, et~al.
\newblock {WeatherBench 2: A benchmark for the next generation of data-driven global weather models}, 8 2023.
\newblock URL \url{http://arxiv.org/abs/2308.15560}.

\bibitem[Brenowitz et~al.(2024)Brenowitz, Cohen, Pathak, Mahesh, Bonev, Kurth, Durran, Harrington, and Pritchard]{Brenowitz2024}
Noah~D. Brenowitz, Yair Cohen, Jaideep Pathak, Ankur Mahesh, Boris Bonev, Thorsten Kurth, Dale~R. Durran, Peter Harrington, and Michael~S. Pritchard.
\newblock {A Practical Probabilistic Benchmark for AI Weather Models}.
\newblock 1 2024.
\newblock URL \url{http://arxiv.org/abs/2401.15305}.

\bibitem[Subich et~al.(2025)Subich, Husain, Separovic, and Yang]{Subich2025}
Christopher Subich, Syed~Zahid Husain, Leo Separovic, and Jing Yang.
\newblock {Fixing the Double Penalty in Data-Driven Weather Forecasting Through a Modified Spherical Harmonic Loss Function}, 1 2025.
\newblock URL \url{http://arxiv.org/abs/2501.19374}.

\bibitem[Lessig et~al.(2023)Lessig, Luise, Gong, Langguth, Stadtler, and Schultz]{Lessig2023}
Christian Lessig, Ilaria Luise, Bing Gong, Michael Langguth, Scarlet Stadtler, and Martin Schultz.
\newblock {AtmoRep: A stochastic model of atmosphere dynamics using large scale representation learning}, 8 2023.
\newblock URL \url{http://arxiv.org/abs/2308.13280}.

\bibitem[Andrae et~al.(2024)Andrae, Landelius, Oskarsson, and Lindsten]{Andrae2024}
Martin Andrae, Tomas Landelius, Joel Oskarsson, and Fredrik Lindsten.
\newblock {Continuous Ensemble Weather Forecasting with Diffusion models}, 10 2024.
\newblock URL \url{http://arxiv.org/abs/2410.05431}.

\bibitem[Lang et~al.(2024)Lang, Alexe, Clare, Roberts, Adewoyin, Bouallègue, Chantry, Dramsch, Dueben, Hahner, Maciel, Prieto-Nemesio, O'Brien, Pinault, Polster, Raoult, Tietsche, and Leutbecher]{Lang2024}
Simon Lang, Mihai Alexe, Mariana C.~A. Clare, Christopher Roberts, Rilwan Adewoyin, Zied~Ben Bouallègue, Matthew Chantry, Jesper Dramsch, Peter~D. Dueben, Sara Hahner, Pedro Maciel, Ana Prieto-Nemesio, Cathal O'Brien, Florian Pinault, Jan Polster, et~al.
\newblock {AIFS-CRPS: Ensemble forecasting using a model trained with a loss function based on the Continuous Ranked Probability Score}, 12 2024.
\newblock URL \url{http://arxiv.org/abs/2412.15832}.

\bibitem[Ho et~al.(2020)Ho, Jain, and Abbeel]{Ho2020}
Jonathan Ho, Ajay Jain, and Pieter Abbeel.
\newblock {Denoising Diffusion Probabilistic Models}, 6 2020.
\newblock URL \url{http://arxiv.org/abs/2006.11239}.

\bibitem[Gneiting and Raftery(2007)]{Gneiting2007}
Tilmann Gneiting and Adrian~E Raftery.
\newblock {Strictly Proper Scoring Rules, Prediction, and Estimation}.
\newblock \emph{Journal of the American Statistical Association}, 102:\penalty0 359--378, 3 2007.
\newblock ISSN 0162-1459.
\newblock \doi{10.1198/016214506000001437}.
\newblock URL \url{http://www.tandfonline.com/doi/abs/10.1198/016214506000001437}.

\bibitem[Lauritzen et~al.(2011)Lauritzen, Jablonowski, Taylor, and Nair]{Lauritzen2011}
Peter Lauritzen, Christiane Jablonowski, Mark Taylor, and Ramachandran Nair.
\newblock \emph{Numerical techniques for global atmospheric models}.
\newblock Lecture notes in computational science and engineering ; 80. Springer, Berlin ;, 2011.
\newblock ISBN 9783642116391.

\bibitem[Vaswani et~al.(2017)Vaswani, Shazeer, Parmar, Uszkoreit, Jones, Gomez, Kaiser, and Polosukhin]{Vaswani2017}
Ashish Vaswani, Noam Shazeer, Niki Parmar, Jakob Uszkoreit, Llion Jones, Aidan~N. Gomez, Lukasz Kaiser, and Illia Polosukhin.
\newblock {Attention Is All You Need}.
\newblock \emph{Advances in Neural Information Processing Systems}, 6 2017.
\newblock URL \url{http://arxiv.org/abs/1706.03762}.

\bibitem[Pfaff et~al.(2021)Pfaff, Fortunato, Sanchez-Gonzalez, and Battaglia]{Pfaff2021}
Tobias Pfaff, Meire Fortunato, Alvaro Sanchez-Gonzalez, and Peter~W. Battaglia.
\newblock Learning mesh-based simulation with graph networks.
\newblock 6 2021.
\newblock URL \url{http://arxiv.org/abs/2010.03409}.

\bibitem[Kaplan et~al.(2020)Kaplan, McCandlish, Henighan, Brown, Chess, Child, Gray, Radford, Wu, and Amodei]{Kaplan2020}
Jared Kaplan, Sam McCandlish, Tom Henighan, Tom~B. Brown, Benjamin Chess, Rewon Child, Scott Gray, Alec Radford, Jeffrey Wu, and Dario Amodei.
\newblock Scaling laws for neural language models.
\newblock 1 2020.
\newblock URL \url{http://arxiv.org/abs/2001.08361}.

\bibitem[Haiden et~al.(2024)Haiden, Janousek, Vitart, Tanguy, Prates, and Chevalier]{Haiden2024}
Thomas Haiden, Martin Janousek, Frédéric Vitart, Maliko Tanguy, Fernando Prates, and Matthieu Chevalier.
\newblock Evaluation of ecmwf forecasts.
\newblock \emph{ECMWF Newsletter}, 2024.
\newblock \doi{10.21957/52F2F31351}.
\newblock URL \url{https://www.ecmwf.int/en/elibrary/81582-evaluation-ecmwf-forecasts}.

\bibitem[Palmer et~al.(2009)Palmer, Buizza, Doblas-Reyes, Jung, Leutbecher, Shutts, Steinheimer, and Weisheimer]{Palmer2009}
T~N Palmer, R~Buizza, F~Doblas-Reyes, T~Jung, M~Leutbecher, G~J Shutts, M~Steinheimer, and A~Weisheimer.
\newblock {Stochastic Parametrization and Model Uncertainty}.
\newblock Technical report, ECMWF, 2009.
\newblock URL \url{http://www.ecmwf.int/publications/}.

\bibitem[Ocampo et~al.(2022)Ocampo, Price, and McEwen]{Ocampo2022}
Jeremy Ocampo, Matthew~A. Price, and Jason~D. McEwen.
\newblock {Scalable and Equivariant Spherical CNNs by Discrete-Continuous (DISCO) Convolutions}, 9 2022.
\newblock URL \url{http://arxiv.org/abs/2209.13603}.

\bibitem[Liu-Schiaffini et~al.(2024)Liu-Schiaffini, Berner, Bonev, Kurth, Azizzadenesheli, and Anandkumar]{Liu2024}
Miguel Liu-Schiaffini, Julius Berner, Boris Bonev, Thorsten Kurth, Kamyar Azizzadenesheli, and Anima Anandkumar.
\newblock {Neural Operators with Localized Integral and Differential Kernels}, 2 2024.
\newblock URL \url{http://arxiv.org/abs/2402.16845}.

\bibitem[Liu et~al.(2022)Liu, Mao, Wu, Feichtenhofer, Darrell, and Xie]{Liu2022}
Zhuang Liu, Hanzi Mao, Chao-Yuan Wu, Christoph Feichtenhofer, Trevor Darrell, and Saining Xie.
\newblock {A ConvNet for the 2020s}, 1 2022.
\newblock URL \url{http://arxiv.org/abs/2201.03545}.

\bibitem[Hendrycks and Gimpel(2016)]{Hendrycks2016}
Dan Hendrycks and Kevin Gimpel.
\newblock {Gaussian Error Linear Units (GELUs)}.
\newblock 6 2016.
\newblock URL \url{http://arxiv.org/abs/1606.08415}.

\bibitem[Xu et~al.(2020)Xu, Lee, Chen, Choi, Hechtman, and Wang]{DBLP:journals/corr/abs-2004-13336}
Yuanzhong Xu, HyoukJoong Lee, Dehao Chen, Hongjun Choi, Blake~A. Hechtman, and Shibo Wang.
\newblock {Automatic Cross-Replica Sharding of Weight Update in Data-Parallel Training}.
\newblock \emph{CoRR}, abs/2004.13336, 2020.
\newblock URL \url{https://arxiv.org/abs/2004.13336}.

\bibitem[Rasley et~al.(2020)Rasley, Rajbhandari, Ruwase, and He]{deepspeed2020}
Jeff Rasley, Samyam Rajbhandari, Olatunji Ruwase, and Yuxiong He.
\newblock {DeepSpeed: System Optimizations Enable Training Deep Learning Models with Over 100 Billion Parameters}.
\newblock In \emph{{Proceedings of the 26th ACM SIGKDD International Conference on Knowledge Discovery \& Data Mining}}, KDD '20, page 3505–3506, New York, NY, USA, 2020. Association for Computing Machinery.
\newblock ISBN 9781450379984.
\newblock \doi{10.1145/3394486.3406703}.
\newblock URL \url{https://doi.org/10.1145/3394486.3406703}.

\bibitem[Rajbhandari et~al.(2020)Rajbhandari, Rasley, Ruwase, and He]{zero2020}
Samyam Rajbhandari, Jeff Rasley, Olatunji Ruwase, and Yuxiong He.
\newblock {ZeRO: Memory Optimizations Toward Training Trillion Parameter Models}.
\newblock ArXiv, May 2020.
\newblock URL \url{https://www.microsoft.com/en-us/research/publication/zero-memory-optimizations-toward-training-trillion-parameter-models/}.

\bibitem[OpenAI et~al.(2024)OpenAI, Achiam, Adler, Agarwal, Ahmad, Akkaya, Aleman, Almeida, Altenschmidt, Altman, Anadkat, Avila, Babuschkin, Balaji, Balcom, Baltescu, Bao, Bavarian, Belgum, Bello, Berdine, Bernadett-Shapiro, Berner, Bogdonoff, Boiko, Boyd, Brakman, Brockman, Brooks, Brundage, Button, Cai, Campbell, Cann, Carey, Carlson, Carmichael, Chan, Chang, Chantzis, Chen, Chen, Chen, Chen, Chen, Chess, Cho, Chu, Chung, Cummings, Currier, Dai, Decareaux, Degry, Deutsch, Deville, Dhar, Dohan, Dowling, Dunning, Ecoffet, Eleti, Eloundou, Farhi, Fedus, Felix, Fishman, Forte, Fulford, Gao, Georges, Gibson, Goel, Gogineni, Goh, Gontijo-Lopes, Gordon, Grafstein, Gray, Greene, Gross, Gu, Guo, Hallacy, Han, Harris, He, Heaton, Heidecke, Hesse, Hickey, Hickey, Hoeschele, Houghton, Hsu, Hu, Hu, Huizinga, Jain, Jain, Jang, Jiang, Jiang, Jin, Jin, Jomoto, Jonn, Jun, Kaftan, Łukasz Kaiser, Kamali, Kanitscheider, Keskar, Khan, Kilpatrick, Kim, Kim, Kim, Kirchner, Kiros, Knight, Kokotajlo, Łukasz Kondraciuk, Kondrich,
  Konstantinidis, Kosic, Krueger, Kuo, Lampe, Lan, Lee, Leike, Leung, Levy, Li, Lim, Lin, Lin, Litwin, Lopez, Lowe, Lue, Makanju, Malfacini, Manning, Markov, Markovski, Martin, Mayer, Mayne, McGrew, McKinney, McLeavey, McMillan, McNeil, Medina, Mehta, Menick, Metz, Mishchenko, Mishkin, Monaco, Morikawa, Mossing, Mu, Murati, Murk, Mély, Nair, Nakano, Nayak, Neelakantan, Ngo, Noh, Ouyang, O'Keefe, Pachocki, Paino, Palermo, Pantuliano, Parascandolo, Parish, Parparita, Passos, Pavlov, Peng, Perelman, de~Avila Belbute~Peres, Petrov, de~Oliveira~Pinto, Michael, Pokorny, Pokrass, Pong, Powell, Power, Power, Proehl, Puri, Radford, Rae, Ramesh, Raymond, Real, Rimbach, Ross, Rotsted, Roussez, Ryder, Saltarelli, Sanders, Santurkar, Sastry, Schmidt, Schnurr, Schulman, Selsam, Sheppard, Sherbakov, Shieh, Shoker, Shyam, Sidor, Sigler, Simens, Sitkin, Slama, Sohl, Sokolowsky, Song, Staudacher, Such, Summers, Sutskever, Tang, Tezak, Thompson, Tillet, Tootoonchian, Tseng, Tuggle, Turley, Tworek, Uribe, Vallone, Vijayvergiya,
  Voss, Wainwright, Wang, Wang, Wang, Ward, Wei, Weinmann, Welihinda, Welinder, Weng, Weng, Wiethoff, Willner, Winter, Wolrich, Wong, Workman, Wu, Wu, Wu, Xiao, Xu, Yoo, Yu, Yuan, Zaremba, Zellers, Zhang, Zhang, Zhao, Zheng, Zhuang, Zhuk, and Zoph]{openai2024gpt4technicalreport}
OpenAI, Josh Achiam, Steven Adler, Sandhini Agarwal, Lama Ahmad, Ilge Akkaya, Florencia~Leoni Aleman, Diogo Almeida, Janko Altenschmidt, Sam Altman, Shyamal Anadkat, Red Avila, Igor Babuschkin, Suchir Balaji, Valerie Balcom, et~al.
\newblock {GPT-4 Technical Report}, 2024.
\newblock URL \url{https://arxiv.org/abs/2303.08774}.

\bibitem[Grattafiori et~al.(2024)Grattafiori, Dubey, Jauhri, Pandey, Kadian, Al-Dahle, Letman, Mathur, Schelten, Vaughan, Yang, Fan, Goyal, Hartshorn, Yang, Mitra, Sravankumar, Korenev, Hinsvark, Rao, Zhang, Rodriguez, Gregerson, Spataru, Roziere, Biron, Tang, Chern, Caucheteux, Nayak, Bi, Marra, McConnell, Keller, Touret, Wu, Wong, Ferrer, Nikolaidis, Allonsius, Song, Pintz, Livshits, Wyatt, Esiobu, Choudhary, Mahajan, Garcia-Olano, Perino, Hupkes, Lakomkin, AlBadawy, Lobanova, Dinan, Smith, Radenovic, Guzmán, Zhang, Synnaeve, Lee, Anderson, Thattai, Nail, Mialon, Pang, Cucurell, Nguyen, Korevaar, Xu, Touvron, Zarov, Ibarra, Kloumann, Misra, Evtimov, Zhang, Copet, Lee, Geffert, Vranes, Park, Mahadeokar, Shah, van~der Linde, Billock, Hong, Lee, Fu, Chi, Huang, Liu, Wang, Yu, Bitton, Spisak, Park, Rocca, Johnstun, Saxe, Jia, Alwala, Prasad, Upasani, Plawiak, Li, Heafield, Stone, El-Arini, Iyer, Malik, Chiu, Bhalla, Lakhotia, Rantala-Yeary, van~der Maaten, Chen, Tan, Jenkins, Martin, Madaan, Malo, Blecher,
  Landzaat, de~Oliveira, Muzzi, Pasupuleti, Singh, Paluri, Kardas, Tsimpoukelli, Oldham, Rita, Pavlova, Kambadur, Lewis, Si, Singh, Hassan, Goyal, Torabi, Bashlykov, Bogoychev, Chatterji, Zhang, Duchenne, Çelebi, Alrassy, Zhang, Li, Vasic, Weng, Bhargava, Dubal, Krishnan, Koura, Xu, He, Dong, Srinivasan, Ganapathy, Calderer, Cabral, Stojnic, Raileanu, Maheswari, Girdhar, Patel, Sauvestre, Polidoro, Sumbaly, Taylor, Silva, Hou, Wang, Hosseini, Chennabasappa, Singh, Bell, Kim, Edunov, Nie, Narang, Raparthy, Shen, Wan, Bhosale, Zhang, Vandenhende, Batra, Whitman, Sootla, Collot, Gururangan, Borodinsky, Herman, Fowler, Sheasha, Georgiou, Scialom, Speckbacher, Mihaylov, Xiao, Karn, Goswami, Gupta, Ramanathan, Kerkez, Gonguet, Do, Vogeti, Albiero, Petrovic, Chu, Xiong, Fu, Meers, Martinet, Wang, Wang, Tan, Xia, Xie, Jia, Wang, Goldschlag, Gaur, Babaei, Wen, Song, Zhang, Li, Mao, Coudert, Yan, Chen, Papakipos, Singh, Srivastava, Jain, Kelsey, Shajnfeld, Gangidi, Victoria, Goldstand, Menon, Sharma, Boesenberg,
  Baevski, Feinstein, Kallet, Sangani, Teo, Yunus, Lupu, Alvarado, Caples, Gu, Ho, Poulton, Ryan, Ramchandani, Dong, Franco, Goyal, Saraf, Chowdhury, Gabriel, Bharambe, Eisenman, Yazdan, James, Maurer, Leonhardi, Huang, Loyd, Paola, Paranjape, Liu, Wu, Ni, Hancock, Wasti, Spence, Stojkovic, Gamido, Montalvo, Parker, Burton, Mejia, Liu, Wang, Kim, Zhou, Hu, Chu, Cai, Tindal, Feichtenhofer, Gao, Civin, Beaty, Kreymer, Li, Adkins, Xu, Testuggine, David, Parikh, Liskovich, Foss, Wang, Le, Holland, Dowling, Jamil, Montgomery, Presani, Hahn, Wood, Le, Brinkman, Arcaute, Dunbar, Smothers, Sun, Kreuk, Tian, Kokkinos, Ozgenel, Caggioni, Kanayet, Seide, Florez, Schwarz, Badeer, Swee, Halpern, Herman, Sizov, Guangyi, Zhang, Lakshminarayanan, Inan, Shojanazeri, Zou, Wang, Zha, Habeeb, Rudolph, Suk, Aspegren, Goldman, Zhan, Damlaj, Molybog, Tufanov, Leontiadis, Veliche, Gat, Weissman, Geboski, Kohli, Lam, Asher, Gaya, Marcus, Tang, Chan, Zhen, Reizenstein, Teboul, Zhong, Jin, Yang, Cummings, Carvill, Shepard, McPhie,
  Torres, Ginsburg, Wang, Wu, U, Saxena, Khandelwal, Zand, Matosich, Veeraraghavan, Michelena, Li, Jagadeesh, Huang, Chawla, Huang, Chen, Garg, A, Silva, Bell, Zhang, Guo, Yu, Moshkovich, Wehrstedt, Khabsa, Avalani, Bhatt, Mankus, Hasson, Lennie, Reso, Groshev, Naumov, Lathi, Keneally, Liu, Seltzer, Valko, Restrepo, Patel, Vyatskov, Samvelyan, Clark, Macey, Wang, Hermoso, Metanat, Rastegari, Bansal, Santhanam, Parks, White, Bawa, Singhal, Egebo, Usunier, Mehta, Laptev, Dong, Cheng, Chernoguz, Hart, Salpekar, Kalinli, Kent, Parekh, Saab, Balaji, Rittner, Bontrager, Roux, Dollar, Zvyagina, Ratanchandani, Yuvraj, Liang, Alao, Rodriguez, Ayub, Murthy, Nayani, Mitra, Parthasarathy, Li, Hogan, Battey, Wang, Howes, Rinott, Mehta, Siby, Bondu, Datta, Chugh, Hunt, Dhillon, Sidorov, Pan, Mahajan, Verma, Yamamoto, Ramaswamy, Lindsay, Lindsay, Feng, Lin, Zha, Patil, Shankar, Zhang, Zhang, Wang, Agarwal, Sajuyigbe, Chintala, Max, Chen, Kehoe, Satterfield, Govindaprasad, Gupta, Deng, Cho, Virk, Subramanian, Choudhury,
  Goldman, Remez, Glaser, Best, Koehler, Robinson, Li, Zhang, Matthews, Chou, Shaked, Vontimitta, Ajayi, Montanez, Mohan, Kumar, Mangla, Ionescu, Poenaru, Mihailescu, Ivanov, Li, Wang, Jiang, Bouaziz, Constable, Tang, Wu, Wang, Wu, Gao, Kleinman, Chen, Hu, Jia, Qi, Li, Zhang, Zhang, Adi, Nam, Yu, Wang, Zhao, Hao, Qian, Li, He, Rait, DeVito, Rosnbrick, Wen, Yang, Zhao, and Ma]{grattafiori2024llama3herdmodels}
Aaron Grattafiori, Abhimanyu Dubey, Abhinav Jauhri, Abhinav Pandey, Abhishek Kadian, Ahmad Al-Dahle, Aiesha Letman, Akhil Mathur, Alan Schelten, Alex Vaughan, Amy Yang, Angela Fan, Anirudh Goyal, Anthony Hartshorn, Aobo Yang, et~al.
\newblock {The Llama 3 Herd of Models}, 2024.
\newblock URL \url{https://arxiv.org/abs/2407.21783}.

\bibitem[Hersbach et~al.(2020)Hersbach, Bell, Berrisford, Hirahara, Horányi, Muñoz-Sabater, Nicolas, Peubey, Radu, Schepers, Simmons, Soci, Abdalla, Abellan, Balsamo, Bechtold, Biavati, Bidlot, Bonavita, Chiara, Dahlgren, Dee, Diamantakis, Dragani, Flemming, Forbes, Fuentes, Geer, Haimberger, Healy, Hogan, Hólm, Janisková, Keeley, Laloyaux, Lopez, Lupu, Radnoti, de~Rosnay, Rozum, Vamborg, Villaume, and Thépaut]{Hersbach2020}
Hans Hersbach, Bill Bell, Paul Berrisford, Shoji Hirahara, András Horányi, Joaquín Muñoz-Sabater, Julien Nicolas, Carole Peubey, Raluca Radu, Dinand Schepers, Adrian Simmons, Cornel Soci, Saleh Abdalla, Xavier Abellan, Gianpaolo Balsamo, et~al.
\newblock {The ERA5 global reanalysis}.
\newblock \emph{Quarterly Journal of the Royal Meteorological Society}, 146:\penalty0 1999--2049, 7 2020.
\newblock ISSN 1477870X.
\newblock \doi{10.1002/qj.3803}.

\bibitem[Alexe et~al.(2024)Alexe, Lang, Clare, Martin, Leutbecher, Roberts, Magnusson, Chantry, Adewoyin, Prieto-Nemesio, Dramsch, Pinault, and Raoult]{Alexe2024}
Mihai Alexe, Simon Lang, Mariana Clare, Martin, Leutbecher, Christopher Roberts, Linus Magnusson, Matthew Chantry, Rilwan Adewoyin, Ana Prieto-Nemesio, Jesper Dramsch, Florian Pinault, and Baudouin Raoult.
\newblock Data-driven ensemble forecasting with the aifs.
\newblock \url{https://www.ecmwf.int/en/newsletter/181/earth-system-science/data-driven-ensemble-forecasting-aifs}, October 2024.
\newblock Accessed: 2025-07-06.

\bibitem[Song et~al.(2023)Song, Dhariwal, Chen, and Sutskever]{Song2023}
Yang Song, Prafulla Dhariwal, Mark Chen, and Ilya Sutskever.
\newblock {Consistency Models}, 3 2023.
\newblock URL \url{http://arxiv.org/abs/2303.01469}.

\bibitem[Goodfellow et~al.(2014)Goodfellow, Pouget-Abadie, Mirza, Xu, Warde-Farley, Ozair, Courville, and Bengio]{Goodfellow2014}
Ian~J. Goodfellow, Jean Pouget-Abadie, Mehdi Mirza, Bing Xu, David Warde-Farley, Sherjil Ozair, Aaron Courville, and Yoshua Bengio.
\newblock {Generative Adversarial Networks}, 6 2014.
\newblock URL \url{http://arxiv.org/abs/1406.2661}.

\bibitem[Driscoll and Healy(1994)]{Driscoll1994}
J.R. Driscoll and D.M. Healy.
\newblock {Computing Fourier Transforms and Convolutions on the 2-Sphere}.
\newblock \emph{Advances in Applied Mathematics}, 15:\penalty0 202--250, 6 1994.
\newblock ISSN 01968858.
\newblock \doi{10.1006/aama.1994.1008}.
\newblock URL \url{https://linkinghub.elsevier.com/retrieve/pii/S0196885884710086}.

\bibitem[McEwen and Wiaux(2011)]{McEwen2011}
Jason.~D. McEwen and Yves Wiaux.
\newblock {A novel sampling theorem on the sphere}, 10 2011.
\newblock URL \url{http://arxiv.org/abs/1110.6298 http://dx.doi.org/10.1109/TSP.2011.2166394}.

\bibitem[Clenshaw and Curtis(1960)]{Clenshaw1960}
C~W Clenshaw and A~R Curtis.
\newblock A method for numerical integration on an automatic computer.
\newblock \emph{Numerische Mathematik}, 2:\penalty0 197--205, 1960.
\newblock ISSN 0945-3245.
\newblock \doi{10.1007/BF01386223}.
\newblock URL \url{https://doi.org/10.1007/BF01386223}.

\bibitem[Golub and Welsch(1969)]{Golub1969}
Gene~H. Golub and John~H. Welsch.
\newblock {Calculation of Gauss quadrature rules}.
\newblock \emph{Mathematics of Computation}, 23:\penalty0 221--230, 1969.
\newblock ISSN 0025-5718.
\newblock \doi{10.1090/S0025-5718-69-99647-1}.
\newblock URL \url{https://www.ams.org/mcom/1969-23-106/S0025-5718-69-99647-1/}.

\bibitem[Cohen and Welling(2016)]{Cohen2016}
Taco~S. Cohen and Max Welling.
\newblock {Group Equivariant Convolutional Networks}, 2 2016.
\newblock URL \url{http://arxiv.org/abs/1602.07576}.

\bibitem[Schaeffer(2013)]{Schaeffer2013}
Nathanaël Schaeffer.
\newblock {Efficient spherical harmonic transforms aimed at pseudospectral numerical simulations}.
\newblock \emph{Geochemistry, Geophysics, Geosystems}, 14:\penalty0 751--758, 3 2013.
\newblock ISSN 15252027.
\newblock \doi{10.1002/ggge.20071}.

\bibitem[Cobb et~al.(2020)Cobb, Wallis, Mavor-Parker, Marignier, Price, d'Avezac, and McEwen]{Cobb2020}
Oliver~J. Cobb, Christopher G.~R. Wallis, Augustine~N. Mavor-Parker, Augustin Marignier, Matthew~A. Price, Mayeul d'Avezac, and Jason~D. McEwen.
\newblock {Efficient Generalized Spherical CNNs}, 10 2020.
\newblock URL \url{http://arxiv.org/abs/2010.11661}.

\bibitem[Giraldo et~al.(2002)Giraldo, Hesthaven, and Warburton]{Giraldo2002}
F.X. Giraldo, J.S. Hesthaven, and T.~Warburton.
\newblock {Nodal High-Order Discontinuous Galerkin Methods for the Spherical Shallow Water Equations}.
\newblock \emph{Journal of Computational Physics}, 181:\penalty0 499--525, 9 2002.
\newblock ISSN 00219991.
\newblock \doi{10.1006/jcph.2002.7139}.
\newblock URL \url{https://linkinghub.elsevier.com/retrieve/pii/S0021999102971391}.

\bibitem[LeVeque(1992)]{LeVeque1992}
Randall~J. LeVeque.
\newblock \emph{{Numerical Methods for Conservation Laws}}.
\newblock Birkhäuser Basel, 1992.
\newblock ISBN 978-3-7643-2723-1.
\newblock \doi{10.1007/978-3-0348-8629-1}.
\newblock URL \url{http://link.springer.com/10.1007/978-3-0348-8629-1}.

\bibitem[Ronneberger et~al.(2015)Ronneberger, Fischer, and Brox]{Ronneberger2015}
Olaf Ronneberger, Philipp Fischer, and Thomas Brox.
\newblock {U-Net: Convolutional Networks for Biomedical Image Segmentation}, 5 2015.
\newblock URL \url{http://arxiv.org/abs/1505.04597}.

\bibitem[Harris(1978)]{Harris1978}
F.J. Harris.
\newblock {On the use of windows for harmonic analysis with the discrete Fourier transform}.
\newblock \emph{Proceedings of the IEEE}, 66:\penalty0 51--83, 1978.
\newblock ISSN 0018-9219.
\newblock \doi{10.1109/PROC.1978.10837}.
\newblock URL \url{http://ieeexplore.ieee.org/document/1455106/}.

\bibitem[Paszke et~al.(2019)Paszke, Gross, Massa, Lerer, Bradbury, Chanan, Killeen, Lin, Gimelshein, Antiga, Desmaison, Köpf, Yang, DeVito, Raison, Tejani, Chilamkurthy, Steiner, Fang, Bai, and Chintala]{Paszke2019}
Adam Paszke, Sam Gross, Francisco Massa, Adam Lerer, James Bradbury, Gregory Chanan, Trevor Killeen, Zeming Lin, Natalia Gimelshein, Luca Antiga, Alban Desmaison, Andreas Köpf, Edward Yang, Zach DeVito, Martin Raison, et~al.
\newblock {PyTorch: An Imperative Style, High-Performance Deep Learning Library}, 12 2019.
\newblock URL \url{http://arxiv.org/abs/1912.01703}.

\bibitem[Li et~al.(2015)Li, Yang, Martin, Ho, and Ying]{Li2015}
Yingzhou Li, Haizhao Yang, Eileen~R. Martin, Kenneth~L. Ho, and Lexing Ying.
\newblock {Butterfly Factorization}.
\newblock \emph{Multiscale Modeling \& Simulation}, 13:\penalty0 714--732, 1 2015.
\newblock ISSN 1540-3459.
\newblock \doi{10.1137/15M1007173}.
\newblock URL \url{http://epubs.siam.org/doi/10.1137/15M1007173}.

\bibitem[Bi et~al.(2023)Bi, Xie, Zhang, Chen, Gu, and Tian]{Bi2023}
Kaifeng Bi, Lingxi Xie, Hengheng Zhang, Xin Chen, Xiaotao Gu, and Qi~Tian.
\newblock {Accurate medium-range global weather forecasting with 3D neural networks}.
\newblock \emph{Nature}, 619:\penalty0 533--538, 7 2023.
\newblock ISSN 14764687.
\newblock \doi{10.1038/s41586-023-06185-3}.

\bibitem[Chen et~al.(2023)Chen, Zhong, Zhang, Xu, Chen, Zhu, Li, Qian, and Chen]{Fuxi2023}
L.~Chen, X.~Zhong, F.~Zhang, Y.~Xu, Y.~Chen, F.~Zhu, H.~Li, Y.~Qian, and L.~Chen.
\newblock {FuXi: a cascade machine learning forecasting system for 15-day global weather forecast}.
\newblock \emph{npj Climate and Atmospheric Science}, 6:\penalty0 190, 2023.
\newblock \doi{10.1038/s41612-023-00512-1}.

\bibitem[Touvron et~al.(2021)Touvron, Cord, Sablayrolles, Synnaeve, and J{\'{e}}gou]{layerscale2021}
Hugo Touvron, Matthieu Cord, Alexandre Sablayrolles, Gabriel Synnaeve, and Herv{\'{e}} J{\'{e}}gou.
\newblock {Going deeper with Image Transformers}.
\newblock \emph{CoRR}, abs/2103.17239, 2021.
\newblock URL \url{https://arxiv.org/abs/2103.17239}.

\bibitem[He et~al.(2015)He, Zhang, Ren, and Sun]{He2015}
Kaiming He, Xiangyu Zhang, Shaoqing Ren, and Jian Sun.
\newblock {Delving Deep into Rectifiers: Surpassing Human-Level Performance on ImageNet Classification}, 2 2015.
\newblock URL \url{http://arxiv.org/abs/1502.01852}.

\bibitem[Siddiqui et~al.(2024)Siddiqui, Kossaifi, Bonev, Choy, Kautz, Krueger, and Azizzadenesheli]{Siddiqui2024}
Shoaib~Ahmed Siddiqui, Jean Kossaifi, Boris Bonev, Christopher Choy, Jan Kautz, David Krueger, and Kamyar Azizzadenesheli.
\newblock {Exploring the design space of deep-learning-based weather forecasting systems}, 10 2024.
\newblock URL \url{http://arxiv.org/abs/2410.07472}.

\bibitem[Karras et~al.(2021)Karras, Aittala, Laine, Härkönen, Hellsten, Lehtinen, and Aila]{Karras2021}
Tero Karras, Miika Aittala, Samuli Laine, Erik Härkönen, Janne Hellsten, Jaakko Lehtinen, and Timo Aila.
\newblock {Alias-Free Generative Adversarial Networks}, 6 2021.
\newblock URL \url{http://arxiv.org/abs/2106.12423}.

\bibitem[Fortin et~al.(2014)Fortin, Abaza, Anctil, and Turcotte]{Fortin2014}
Vincent Fortin, M.~Abaza, F.~Anctil, and R.~Turcotte.
\newblock {Why should ensemble spread match the RMSE of the ensemble mean?}
\newblock \emph{Journal of Hydrometeorology}, 15:\penalty0 1708--1713, 8 2014.
\newblock ISSN 15257541.
\newblock \doi{10.1175/JHM-D-14-0008.1}.

\bibitem[Gneiting and Raftery(2004)]{Gneiting2004}
Tilmann Gneiting and Adrian~E Raftery.
\newblock {Strictly Proper Scoring Rules, Prediction, and Estimation}, 2004.
\newblock ISSN 0704-0188.

\bibitem[Zamo and Naveau(2018)]{Zamo2018}
Michaël Zamo and Philippe Naveau.
\newblock {Estimation of the Continuous Ranked Probability Score with Limited Information and Applications to Ensemble Weather Forecasts}.
\newblock \emph{Mathematical Geosciences}, 50:\penalty0 209--234, 2 2018.
\newblock ISSN 18748953.
\newblock \doi{10.1007/s11004-017-9709-7}.

\bibitem[Kingma and Ba(2014)]{Kingma2014}
Diederik~P. Kingma and Jimmy Ba.
\newblock {Adam: A Method for Stochastic Optimization}, 12 2014.
\newblock URL \url{http://arxiv.org/abs/1412.6980}.

\bibitem[Carrassi et~al.(2017)Carrassi, Bocquet, Bertino, and Evensen]{Carrassi2017}
Alberto Carrassi, Marc Bocquet, Laurent Bertino, and Geir Evensen.
\newblock {Data Assimilation in the Geosciences - An overview on methods, issues and perspectives}, 9 2017.
\newblock URL \url{http://arxiv.org/abs/1709.02798}.

\bibitem[Rasp and Lerch(2018)]{Rasp2018}
Stephan Rasp and Sebastian Lerch.
\newblock {Neural Networks for Postprocessing Ensemble Weather Forecasts}.
\newblock \emph{Monthly Weather Review}, pages 3885--3900, 2018.
\newblock \doi{10.1175/MWR-D-18}.
\newblock URL \url{https://doi.org/10.1175/MWR-D-18-}.

\bibitem[Arcomano et~al.(2020)Arcomano, Szunyogh, Pathak, Wikner, Hunt, and Ott]{Arcomano2020}
Troy Arcomano, Istvan Szunyogh, Jaideep Pathak, Alexander Wikner, Brian~R. Hunt, and Edward Ott.
\newblock {A Machine Learning-Based Global Atmospheric Forecast Model}.
\newblock \emph{Geophysical Research Letters}, 47, 5 2020.
\newblock ISSN 0094-8276.
\newblock \doi{10.1029/2020GL087776}.
\newblock URL \url{https://agupubs.onlinelibrary.wiley.com/doi/10.1029/2020GL087776}.

\bibitem[Hamill(2001)]{Hamill2001}
Thomas~M. Hamill.
\newblock {Interpretation of Rank Histograms for Verifying Ensemble Forecasts}.
\newblock \emph{Monthly Weather Review}, 129:\penalty0 550--560, 3 2001.
\newblock ISSN 0027-0644.
\newblock \doi{10.1175/1520-0493(2001)129<0550:IORHFV>2.0.CO;2}.
\newblock URL \url{http://journals.ametsoc.org/doi/10.1175/1520-0493(2001)129<0550:IORHFV>2.0.CO;2}.

\bibitem[Tulloch and Smith(2006)]{Tulloch2006}
R.~Tulloch and K.~S. Smith.
\newblock {A theory for the atmospheric energy spectrum: Depth-limited temperature anomalies at the tropopause}.
\newblock \emph{Proceedings of the National Academy of Sciences}, 103\penalty0 (40):\penalty0 14690--14694, 2006.
\newblock \doi{10.1073/pnas.0605494103}.
\newblock URL \url{https://www.pnas.org/doi/abs/10.1073/pnas.0605494103}.

\bibitem[Zhao et~al.(2023)Zhao, Gu, Varma, Luo, Huang, Xu, Wright, Shojanazeri, Ott, Shleifer, Desmaison, Balioglu, Damania, Nguyen, Chauhan, Hao, Mathews, and Li]{Zhao2023}
Yanli Zhao, Andrew Gu, Rohan Varma, Liang Luo, Chien~Chin Huang, Min Xu, Less Wright, Hamid Shojanazeri, Myle Ott, Sam Shleifer, Alban Desmaison, Can Balioglu, Pritam Damania, Bernard Nguyen, Geeta Chauhan, et~al.
\newblock Pytorch fsdp: Experiences on scaling fully sharded data parallel.
\newblock In \emph{Proceedings of the VLDB Endowment}, volume~16, pages 3848--3860. VLDB Endowment, 2023.
\newblock \doi{10.14778/3611540.3611569}.

\end{thebibliography}

\newpage

\appendix

\etocdepthtag.toc{mtappendix}
\etocsettagdepth{mtmanuscript}{none}
\etocsettagdepth{mtappendix}{subsubsection}
\tableofcontents

\newpage

\section{Problem statement}
\label{sec:problem_statement}

\subsection{Deterministic forecasting}
We aim to predict the dynamics of Earth's atmosphere using a machine learning (ML) model trained on available data. The state of the atmosphere is represented as a vector-valued function $u(x, t)$ over the sphere, where $x \in S^2$ and time $t \in [0, \infty)$. The dynamics are modeled as a discrete dynamical system:
\begin{equation}
    \label{eq:dynamic_model}
    u_{n+1} = F_{\theta}(u_n, t_n),
\end{equation}
where the current state $u_n \coloneqq u(x, t_n)$ at the time $t_n$ is mapped to the next state $u_{n+1}$ at the time $t_{n+1}$ by the learned operator $F_{\theta}$ which approximates the unknown, true operator $F^*$.\footnote{To simplify the discussion, we do not disambiguate between spatially discretized $u_n$ and its semi-continuous counterpart $u_n(x)$.}

The unknown parameter vector $\theta$ is obtained by minimizing a suitable objective function $\mathcal{L}_\text{det}(\cdot,\cdot)$, which measures the discrepancy between the ground-truth solution $u^*_{n+1} = F^*(u_n, t_n)$ and the prediction of the learned mapping $F_\theta(u_n, t_n)$:
\begin{equation}
    \label{eq:deterministic_objective}
    \theta^* = \argmin_\theta \sum_n \mathcal{L}_\text{det}\left( F_\theta(u_n, t_n),u^*_{n+1}\right).
\end{equation}
The loss is summed over the dataset $\{u^*_n\}_{n=0}^N$, which consists of ground-truth observations of the state vector of the dynamical system at various times.

\subsection{Probabilistic forecasting}
The limitation of deterministic modeling lies in the objective \eqref{eq:deterministic_objective}, where the model is trained to minimize the error in a single deterministic prediction. While this approach may work well for purely deterministic processes with a sufficiently powerful model $F_\theta$, it falls short when the underlying map $F^*$ is inherently uncertain. In such cases, the model tends to learn a blurry estimator which minimizes the expected value of \eqref{eq:deterministic_objective} \cite{Subich2025, Mahesh2024a, Mahesh2024b}.

To address this issue, we aim to estimate the conditional probability distribution $p(u_{n+1} | u_n, t_n)$ rather than predicting a single sample $u_{n+1}$ given $u_n$. While this distribution could be directly modeled using a network $p_\theta(u_{n+1} | u_n, t_n)$, this approach introduces the additional complexity of sampling from a complicated predictive distribution. 

Generative models offer a more practical alternative by approximating the sampling process directly, which is often easier to optimize. Our goal is to implement a generative model $F_{\theta}(u_n, z_n, t_n)$, which takes the current state $u_n$, time $t_n$, and a suitable latent noise variable $z_n$ to generate a single sample
\begin{equation}
    \label{eq:hidden_markov_model}
    u_{n+1} = F_{\theta}(u_n, z_n, t_n).
\end{equation}
The optimal mapping
\begin{equation}
    \label{eq:generative_model}
    F^*(u_n, z_n, t_n) \sim p(u^*_{n+1} | u_n, t_n),
\end{equation}
produces a random variable distributed according to the ground-truth conditional probability density function $p(u^*_{n+1} | u_n, t_n)$ for an appropriate choice of the distribution of the latent variable $z_n$. By sampling this probabilistic function $N_\text{ens}$ times with different noise realizations $z_{n,e}$, we generate an ensemble of predictions $\{u_{ n+1,e} | e=1,2,\dots, N_\text{ens}\}$, which approximately models the conditional distribution $p(u_{n+1} | u_n)$.

There exist many approaches to train generative models of the form \eqref{eq:generative_model}, such as diffusion models, consistency models or generative adversarial networks \citep{Ho2020, Song2023, Goodfellow2014}. We take a direct approach and minimize a scoring function $\mathcal{L}_\text{ens}$, which assesses the quality of the ensemble forecast directly. As such, the score is computed from the predictive ensemble $u_{n+1,e} = F_\theta(u_n, z_{n,e}, t_n)$ and the single ground truth solution $u^*_{n+1}$ that occurred, and then summed over the training data:
\begin{equation}
    \label{eq:ensemble_objective}
    \theta^* = \argmin_\theta \sum_n \mathcal{L}_\text{ens}\left( \{F_\theta(u_n, z_{n,e}, t_n)\}^{N_\text{ens}}_{e=1},u^*_{n+1} \right).
\end{equation}
In the ideal case, the ground truth trajectory is interchangeable with the individual ensemble members $u_e$, which should be reflected in a minimization of $\mathcal{L}_\text{ens}$. The specific choice of $\mathcal{L}_\text{ens}$ is discussed further in \Cref{sec:evaluation_metrics} and \Cref{sec:objective_function}.

\section{Signal processing on the sphere}
\label{sec:signal_processing}

On a high level, $F_\theta$ acts as a map between functions defined over the sphere. To this end, we employ techniques from signal processing on the sphere to transform spherical signals to other spherical signals.  This chapter outlines the main building blocks employed in the FourCastNet 3 architecture, as well as the evaluation and analysis of its outputs.

\subsection{Grids and quadrature rules on the sphere}
\label{sec:grids_and_quadrature}

Our networks process real-valued input signals defined on the unit sphere, $u: S^2 \rightarrow \mathbb{R}^n$ with coordinates  
\begin{equation}
    x(\vartheta, \varphi) = 
    \begin{bmatrix}
        \sin \vartheta \cos \varphi\\
        \sin \vartheta \sin \varphi\\
        \cos \vartheta
    \end{bmatrix},
\end{equation}
defined by the colatitude $\vartheta\in [0,\pi]$ and longitude $\varphi\in[0,2\pi)$.
Processing these signals on a computer requires discretizing them through a sampling scheme. While there are various sampling theorems on the sphere \cite{Driscoll1994, McEwen2011}, we will restrict our discussion to sampling schemes motivated by quadrature rules. Many of the operations discussed in this chapter require evaluating integrals of the form
\begin{equation}
    \label{eq:integral_sphere}
    \int_{S^2} u(x) \; \mathrm{d}\mu(x) = \int^{2\pi}_{0} \int^\pi_{0} u(x(\vartheta, \varphi)) \; \sin \vartheta \, \mathrm{d} \vartheta \, \mathrm{d} \varphi,
\end{equation}
where $u \in L^1(S^2)$ is an integrable function on the sphere and $\mu(x)$ denotes the Haar measure on the sphere. We approximate this integral using a quadrature rule
\begin{equation}
    \int_{S^2} u(x) \; \mathrm{d}\mu(x) \approx \sum_{i=1}^{n_\text{grid}} u(x_i) \; \omega_i,
\end{equation}
which is characterized by a choice of grid points $\{x_i\}^{n_\text{grid}}_{i=1}$ and corresponding quadrature weights $\{\omega_i\}^{n_\text{grid}}_{i=1}$.

In the context of our model, we are concerned with two types of spherical grids: equiangular (lat/lon-) grids and the Gaussian grids. Both can be formed as a tensor product
\begin{equation}
    \left\{x(\vartheta, \varphi) | \; \forall \vartheta \in \{\vartheta_i\},\, \varphi \in \{\varphi_j\} \right\}
\end{equation}
of one-dimensional grids for the latitude $\{\vartheta_j\}$ and longitude $\{\varphi_k\}$, respectively.
In particular, the equiangular grid uses equally spaced grid points in both longitude and latitude, such that
\begin{subequations}
\label{eq:latlon_grid}
\begin{alignat}{2}
\vartheta_i &= \pi i / n_\text{lat} \quad && \text{for } i=0,1,\dots, n_\text{lat}-1, \\
\varphi_j &= 2 \pi j / n_\text{lon} \quad && \text{for } j=0,1,\dots, n_\text{lon}-1,
\end{alignat}
\end{subequations}
where $n_\text{lat}$ and $n_\text{lon}$ denote the number of grid points in latitude and longitude. The quadrature weights for the combined spherical grid are
\begin{equation}
    \label{eq:latlon_quad_weights}
    \omega_{ij} = \sin \vartheta_i \, \Delta\vartheta_i \, \Delta\varphi_j =  \frac{2 \pi^2}{n_\text{lat} n_\text{lon}} \sin \vartheta_i,
\end{equation}
which approximately sums to $4 \pi$, the area of the unit sphere. These weights correspond to trapezoidal quadrature weights in spherical coordinates \footnote{It is possible to obtain higher order accurate quadrature weights for the same lat/lon-grid, in the form of Clenshaw-Curtis quadrature weights \cite{Clenshaw1960}. To keep results compatible with other models, we use the simpler quadrature formula \eqref{eq:latlon_quad_weights}}. 

The Gaussian grids on the sphere replace the latitude grid in \eqref{eq:latlon_grid} with a set of Gauss-Legendre nodes $\{\vartheta_i\}_i^{n_\text{lat}}$ such that $\cos{\vartheta_i}$ is the $i$-th root of the Legendre polynomial $P_{n_\text{lat}}$, i.e.
\begin{equation}
    P_{n_\text{lat}}(\cos{\vartheta_i}) = 0.
\end{equation}
The usage of Gauss-Legendre nodes in the $\cos{\vartheta}$ domains enables the exact integration of \eqref{eq:integral_sphere}, for polynomial integrands in $\cos{\vartheta}$ and a polynomial degree up to degree $2 n_\text{lat} - 1$  \cite{Golub1969}. This is particularly useful for the exact integration of spherical harmonics, where the latitude-dependent part is a polynomial in $\cos{\vartheta}$.

\begin{figure}[htb]
    \centering
    \includegraphics[width=0.6\linewidth, trim={100pt 80pt 100pt 80pt}, clip]{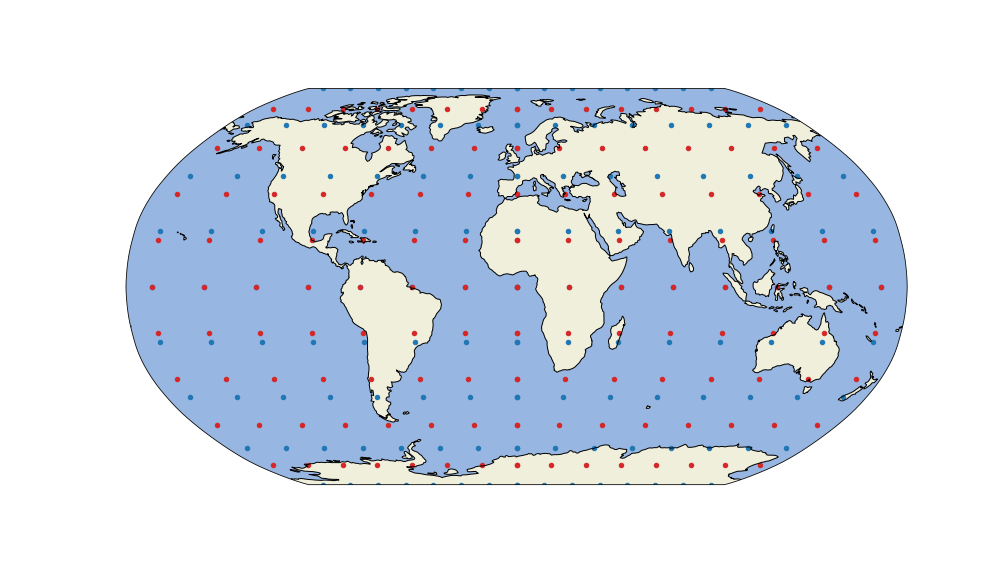}
    \caption{Illustration of the equiangular (lat/lon-) grid in blue and a Gaussian grid in red. Both grids $n_\text{lat}=9$ and $n_\text{lon}=16$.}
    \label{fig:s2_grids}
\end{figure}

An example of both grids is illustrated in \Cref{fig:s2_grids} for $n_\text{lat}=9$ and $n_\text{lon}=16$.

\subsection{Convolutions on the sphere}
\label{sec:s2_convolutions}
Convolutions are attractive in machine learning, as they can encode linear operators that are invariant under translations. To generalize this to the sphere, we formulate the convolution in terms of linear operations invariant with respect to rotations $R$ in the three-dimensional rotation group $SO(3)$. That is, we rely on the (continuous) convolution extended to Lie groups and quotient spaces of Lie Groups, known as group convolution (see e.g. \citet{Cohen2016, Ocampo2022, Driscoll1994}). On the sphere, for two functions $u, k \in L^2(S^2)$, and a rotation $R \in SO(3)$, we define the convolution
\begin{equation}
    \label{eq:group_conv}
    (u \star k)(R) = \int_{S^2} u(x)\,\overline{k(R^{-1}x)}\;\mathrm{d}\mu(x),
\end{equation}
where $\mathrm{d}\mu(x)$ is the invariant Haar measure on $S^2$. In other words, the output is obtained by (passively left-)rotating the filter function $k$ using a rotation $R = Z(\alpha)Y(\beta)Z(\gamma)$, which may be parametrized by the Euler angles $\alpha, \beta, \gamma$ and rotations $Z$ and $Y$ around the z- and y- axes and then taking the $S^2$ inner product to obtain the value for this rotation. As such, the output is a function over $SO(3)$, despite the inputs being functions defined on $S^2$

To obtain a convolution that yields signals on the sphere, we may use the fact that $S^2$ is isomorphic to the quotient $SO(3)/SO(2) \simeq S^2$. As such, we can derive the $S^2$ convolution by restricting the rotation $R$ to $R \in SO(3)/SO(2)$, or in other words, assuming that $\gamma=0$. For a given point $x \in S^2$ and its associated rotation $R_{x} \in SO(3)/SO(2)$, we obtain the spherical convolution
\begin{align}
    \label{eq:s2_conv}
    (u \ostar k)(x)
    &= \int_{S^2} u(x^\prime)\,\overline{k(R_{x}^{-1}x^\prime)}\;\mathrm{d}\mu(x^\prime),
\end{align}
which yields an output in $L^2(S^2)$.

With the continuous convolution defined, we can derive discretized approximations of \eqref{eq:s2_conv} which can be evaluated on a computer. In particular, we have two distinct variations; one using the spectral theorem on the sphere and using spherical harmonic transforms and one using a direct discretization of \eqref{eq:s2_conv}.

\subsection{Spherical harmonic transforms}
\label{spherical_harmonics_transforms}

On the sphere, the Fourier transform is generalized by the spherical harmonic transform (SHT)
\begin{align}
\label{eq:sht}
\widehat{u}_\ell^m =\int_{S^2} u(x) \; \overline{Y_\ell^m(x)} \; \mathrm{d} \mu(x) \, = \int^{2\pi}_{0} \int^\pi_{0} u(\vartheta, \varphi) \; \overline{Y_\ell^m(\vartheta, \varphi)} \; \sin \vartheta \, \mathrm{d} \vartheta \, \mathrm{d} \varphi
\end{align}
and its inverse 
\begin{equation}
\label{eq:isht}
u(\vartheta, \varphi) = \sum_{\ell \in \mathbb{N}} \sum_{|m| \leq l} \hat{u}_\ell^m Y_\ell^m(\vartheta, \varphi),
\end{equation}
which decompose the Hilbert-space $L^2(S^2)$ into subspaces that are invariant under certain rotations $R \in SO(3)/SO(2)$. These subspaces are spanned by the spherical harmonics
\begin{subequations}
\begin{align}
    \label{eq:spherical_harmonics}
    Y_\ell^m(\vartheta, \varphi) &= (-1)^m c_\ell^m P_\ell^m(\cos \vartheta) e^{\mathrm{i} m \varphi}=\widehat{P}_\ell^m(\cos \vartheta) e^{\mathrm{i} m \varphi}, \\
    c_\ell^m &:=\sqrt{\frac{2 \ell+1}{4 \pi} \frac{(\ell-m) !}{(\ell+m) !}},
\end{align}
\end{subequations}
where $P_\ell^m(\cos \vartheta)$ are the associated Legendre polynomials. The normalization factor $c^m_\ell$ normalizes the spherical harmonics w.r.t. the $L^2(S^2)$ inner product, s.t.
\begin{equation}
    \int_{S^2} Y_\ell^m(x) \; \overline{Y_{\ell'}^{m'}(x)} \;\mathrm{d}\mu(x)= \delta_{\ell\ell'} \delta_{mm'}.
\end{equation}
An efficient algorithm to compute the SHT is presented by \citet{Schaeffer2013}, which exploits the tensor-product structure of the spherical harmonics to decompose the integral in \Cref{eq:sht} into a projection into the Fourier basis and a projection onto the associated Legendre polynomials. This can be achieved using a Fast Fourier Transform (FFT) along longitudes and a matrix-vector-multiplication along latitudes, resulting in $\mathcal{O}(n_\text{lat}^2 n_\text{lon} \log n_\text{lon})$ floating-point operations to evaluate the SHT.

\subsection{Convolution theorem and spectral convolutions}
\label{sec:convolution_theorem}

The convolution theorem connects the spherical harmonic transform \eqref{eq:sht} to the spherical convolution \eqref{eq:s2_conv} for axisymmetric filters $k(\vartheta) \in L^2(S^2)$. For a given $l \geq 0$, $|m| \leq l$, the spherical harmonic transform of the convolution $\widehat{f \ostar k}$ is then given by
\begin{align}
    \label{eq:s2_spectral_conv}
    (\widehat{u \ostar k})_\ell^m &= \int_{S^2} (u \ostar k)(x) \; \overline{Y_\ell^m(x)} \; \mathrm{d} \mu(x) \nonumber\\
    &= \widehat{u}_\ell^m \, \widehat{k}_\ell^0,
\end{align}
i.e. as a product of the spherical harmonic transforms $\widehat{u}_\ell^m$ and $\widehat{k}_\ell^m$ of $u$ and $k$, respectively. The restriction to axisymmetric filters $k$ is rooted in the fact that the 2-sphere is not a group; thus, orientable filters have ambiguous rotations when they are rotated on the sphere \cite{Ocampo2022}.

\begin{figure}[htb]
    \centering
    \begin{subfigure}{.3\linewidth}
        \centering
        \includegraphics[width=0.95\linewidth, trim={20pt 20pt 20pt 20pt}, clip]{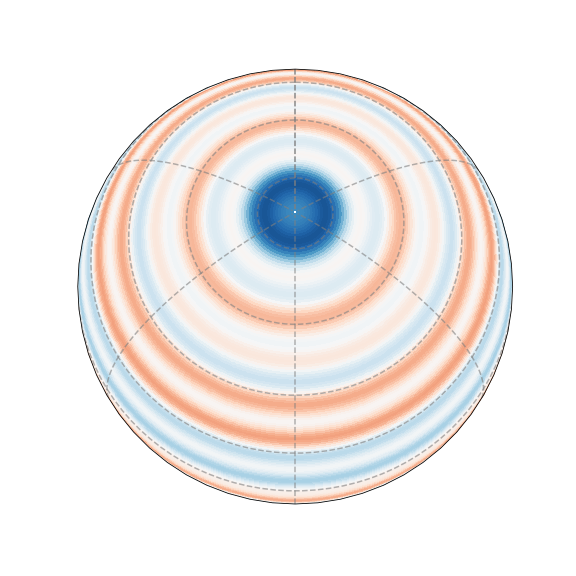}
        \caption{spectral convolution filter}
        \label{fig:random_spectral_filter}
    \end{subfigure}
    \hspace{2cm}
    \begin{subfigure}{.3\linewidth}
        \centering
        \includegraphics[width=0.95\linewidth, trim={20pt 20pt 20pt 20pt}, clip]{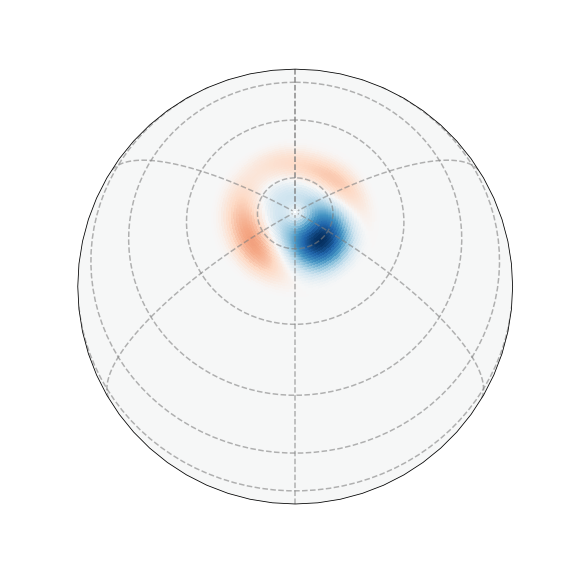}
        \caption{local convolution filter}
        \label{fig:random_disco_filter}
    \end{subfigure}
    \caption{Depiction of random filters $k(x)$ resulting from the filter parametrization. The spectral filter on the left is parameterized as a linear combination of globally supported spherical harmonics, whereas the local filter on the right is parameterized as a linear combination of locally supported, potentially anisotropic basis functions.}
    \label{fig:random_filters}
\end{figure}

The spherical convolution theorem forms the core building block of spherical Fourier neural operators (SFNO) \cite{Bonev2023}, as filters are parametrized in the spectral domain according to \eqref{eq:s2_spectral_conv}.
A randomly initialized spectral filter is illustrated in \Cref{fig:random_spectral_filter}. For a detailed treatment of the convolution theorem on the sphere, we refer the reader to \citet{Driscoll1994, Cobb2020}.

\subsection{Discrete-continuous convolutions}
The spectral parametrization of a convolution \eqref{eq:s2_spectral_conv} conveniently encodes spherical convolutions with smooth, globally supported convolution kernels. Moreover, the nature of the spectral convolution requires the kernel to be isotropic, i.e. to be radially symmetric, which results in the filters depending on the spherical harmonic degree $l$ but not on $m$.

In many applications, however, we would like to encode a locally supported convolution kernel, which results in a non-local filter in spectral space. This is particularly relevant for hyperbolic PDEs\footnote{Atmospheric dynamics are commonly modeled and studied by the shallow water equations on a rotating sphere, which is a hyperbolic PDE system \cite{Giraldo2002}.}, where the solution operator is known to be locally supported \cite{LeVeque1992}. Moreover, anisotropic filters are particularly desirable for the weather system, where the dynamics play out with multiple preferential asymmetries, such as a greater tendency for zonal vs. meridional flows, adiabatic flow confined to vertically tilted isentropes with characteristic morphology, and blocked flow around topographic features. 

To address this, \citet{Ocampo2022} propose a direct discretization of the convolution \eqref{eq:s2_conv}, where the rotation of filters is carried out analytically in the continuous domain and the integral is approximated with numerical integration. This yields the discrete-continuous (DISCO) convolution
\begin{align}
    \label{eq:disco_convolution}
    (u \ostar k)(x) &= \int_{S^2} u(x)\,\overline{k(R_{x}^{-1} x^\prime)}\;\mathrm{d}\mu(x^\prime) \nonumber\\
    &\approx \sum^{n_\text{lat} n_\text{lon}}_{j=1} \overline{k(R_{x}^{-1} x_j)} \, u(x_j) \, \omega_j,
\end{align}
for a suitable quadrature rule with grid points $x_j \in S^2$ and associated quadrature weights $\omega_j$. By analogy, we can also define the transposed discrete-continuous convolution
\begin{align}
    \label{eq:transposed_disco_convolution}
    (u \ostar^\dagger k)(x) &= \int_{S^2}\overline{k(R^{-1}_{x^\prime}x)}\,u(x^\prime)\;\mathrm{d}\mu(x^\prime) \nonumber\\
    &\approx \sum^{n_\text{lat} n_\text{lon}}_{j=1} \overline{k(R_{x_j}^{-1} x)} \, u(x_j) \, \omega_j,
\end{align}
where the integral is taken over all possible rotations $R_{x^\prime}$. This is analogous to transposed convolutions, typically found in computer-vision architectures such as U-Nets \cite{Ronneberger2015}.

For a discrete set of output locations $R_{x_i} \in S^2$, \cref{eq:disco_convolution} becomes a straight-forward matrix-vector multiplication
\begin{align}
    \label{eq:disco_convolution_matrix}
    \sum_{j=1}^{n_\text{lat} n_\text{lon}} \overline{k(R_i^{-1} x_j)} \, u(x_j) \, \omega_j = \sum_{j=1}^{n_\text{lat} n_\text{lon}} K_{ij} \, u(x_j) \, \omega_j
\end{align}
with $K_{ij} \coloneqq \overline{k(R_i^{-1} x_j)}$. In the case where $k(x)$ is compactly supported, $K_{ij}$ is a sparse matrix with the number of non-zero entries per row depending on the resolution of the grid $x_j$ and the support of $k$. The resulting matrix-vector multiplication has linear asymptotic time complexity, as it can be performed using $\mathcal{O}(n_\text{lat}n_\text{lon})$ floating-point operations.

To obtain a learnable filter, $k$ is parametrized as a linear combination
\begin{equation}
    k(x) = \sum_{b=1}^{n_\text{basis}} w_b\; \tilde{k}_b(x)
\end{equation}
of basis functions $\tilde{k}_b(x)$ and learnable weights $w_b$. In practice, we use learnable tensors to merge the features of $u$ and basis functions in the same step, which is similar to a regular convolution in $\mathbb{R}^2$ that mixes features and neighborhood information for each grid point. 

\subsubsection{Filter basis parameterization}
\begin{figure}[htb]
    \centering
    \begin{subfigure}{.19\linewidth}
        \centering
        \includegraphics[width=0.95\linewidth, trim={20pt 20pt 20pt 20pt}, clip]{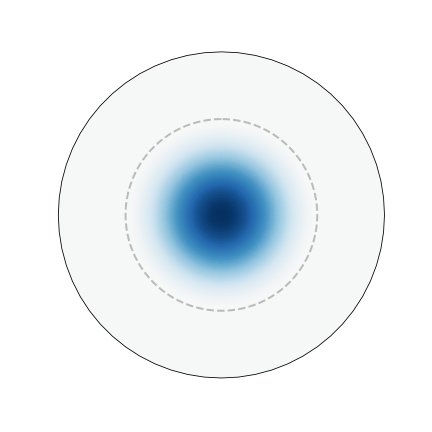}
        \caption{$\ell=0, m=0$}
    \end{subfigure}
    \begin{subfigure}{.19\linewidth}
        \centering
        \includegraphics[width=0.95\linewidth, trim={20pt 20pt 20pt 20pt}, clip]{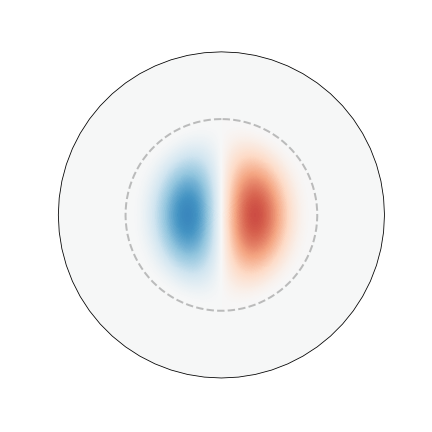}
        \caption{$\ell=0, m=1$}
    \end{subfigure}
    \begin{subfigure}{.19\linewidth}
        \centering
        \includegraphics[width=0.95\linewidth, trim={20pt 20pt 20pt 20pt}, clip]{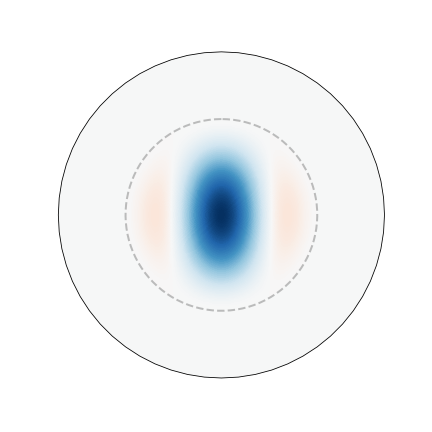}
        \caption{$\ell=0, m=2$}
    \end{subfigure}
    \begin{subfigure}{.19\linewidth}
        \centering
        \includegraphics[width=0.95\linewidth, trim={20pt 20pt 20pt 20pt}, clip]{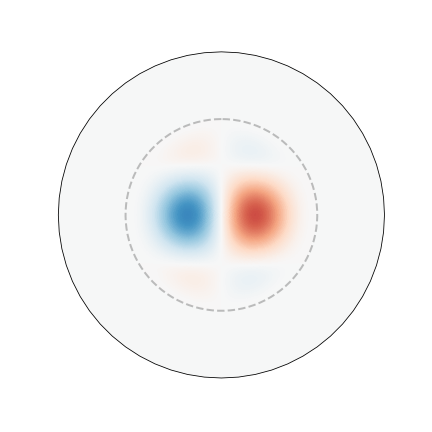}
        \caption{$\ell=2, m=1$}
    \end{subfigure}
    \begin{subfigure}{.19\linewidth}
        \centering
        \includegraphics[width=0.95\linewidth, trim={20pt 20pt 20pt 20pt}, clip]{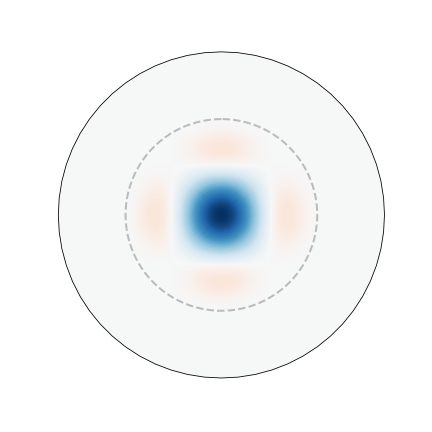}
        \caption{$\ell=2, m=2$}
    \end{subfigure}
    \caption{Illustration of the Morlet-type wavelet filter basis, used to represent localized filters on the sphere. The basis functions are centered at the North Pole, and the dotted line represents the cutoff radius $\vartheta_\text{cutoff}$.}
    \label{fig:morlet_filter_basis}
\end{figure}

While many basis functions can be chosen, we implement a filter-basis inspired by Morlet-like wavelet
\begin{subequations}
\begin{align}
    \tilde{k}_{\ell m}(\vartheta', \varphi) &= h(\vartheta) e^{(i \pi \ell \,\vartheta' \sin \varphi)} e^{(i \pi m \,\vartheta' \cos \varphi)} \\
    h(\vartheta') &= \cos^2\left(\frac{\pi}{2}\vartheta'\right)
\end{align}
\end{subequations}
defined on the compact disk $\vartheta' = \vartheta/\vartheta_\text{cutoff} \in [0, 1], \varphi \in [0, 2\pi)$, centered at the North Pole. The basis functions are constructed by multiplying the Hann windowing function $h$ to Fourier basis functions of degrees $\ell$ and $m$, along the axes spanned by $\vartheta' \sin \varphi$ and $\vartheta' \cos \varphi$. The choice of the Hann windowing function $h(\vartheta')$ \cite{Harris1978} ensures that smooth, compactly supported filters are learned while keeping the convolution tensor $K_{ij}$ sparse. \Cref{fig:morlet_filter_basis} depicts various basis functions for varying choices of $\ell$ and $m$.

\subsection{Interpolation of spherical signals}
\label{sec:interpolation}

A simple way of resampling functions on the sphere is given by the SHT, i.e. by converting a signal to its spectral representation, then evaluating on any grid using the inverse SHT. This process is alias-free, as the power spectrum is appropriately truncated to only contain modes lower than the maximum mode representable on either input or output grid. The disadvantage of this approach is that it can lead to ringing effects (Gibbs phenomenon) due to the sharp cutoff in the frequency domain.

An alternative is bilinear interpolation in longitude and latitude, which can be used to switch between grids and resolutions. Given a function $u$ defined on the input grid $\vartheta_i, \varphi_i$, we find the interpolated output $u'$ at position $\vartheta_i \leq \vartheta \le \vartheta_{i+1}$, $\varphi_i \leq \varphi \le \varphi_{i+1}$ by computing
\begin{align}
    \label{eq:bilinear_interpolation}
    u'(\vartheta, \varphi) =& (1-w_\vartheta)\;(1-w_\varphi)\;u(\vartheta_{i}, \varphi_j) + w_\vartheta\;(1-w_\varphi)\;u(\vartheta_{i+1}, \varphi_j) \nonumber \\
    &+ (1-w_\vartheta)\;w_\varphi\;u(\vartheta_i, \varphi_{j+1}) + w_\vartheta\; w_\varphi\;u(\vartheta_{i+1}, \varphi_{j+1}),
\end{align}
with the interpolation weights $w_\vartheta = \frac{\vartheta - \vartheta_i}{\vartheta_{i+1} - \vartheta_i}$ and $w_\varphi = \frac{\varphi - \varphi_j}{\varphi_{j+1} - \varphi_j}$.

To account for the topology of the sphere, some extra care has to be applied at the boundaries of the grid. For the longitudinal interpolation, we pick $\varphi_{j+1} = \varphi_0$ if $\varphi > \varphi_{n_\text{lon}-1}$, thus accounting for the periodicity in the longitudinal direction. For grids that do not include the poles, we first extend the grid to the poles, defining $\vartheta_{-1} = 0$, $\vartheta_{n_\text{lat}} = \pi$. The function at the poles is set to the weighted mean of the closest latitude-ring
\begin{subequations}
    \label{eq:polar_extension}
    \begin{align}
        u(\vartheta_{-1}, \varphi) &= \frac{1}{2 \pi}\sum_{j=0}^{n_\text{lon}} \Delta \varphi_j \, u(\vartheta_0, \varphi_j) \\ 
        u(\vartheta_{n_\text{lat}}, \varphi) &= \frac{1}{2 \pi}\sum_{j=0}^{n_\text{lon}} \Delta \varphi_j \, u(\vartheta_{n_\text{lat}-1}, \varphi_j),
    \end{align}
\end{subequations}
where the weight $\Delta \varphi_j = 2 \pi / n_\text{lon}$ denotes the longitudinal quadrature weight of each grid point.
By extending the solution to the poles, we can use the bilinear interpolation \eqref{eq:bilinear_interpolation} as well for points that lie between the last latitude ring near the pole and the pole itself.

\subsection{Spherical diffusion process}
\label{sec:spherical_diffusion_process}
As a source of randomness in our models, we require a stochastic process on the sphere, which accurately captures spatio-temporal correlations. To achieve this, we make use of the diffusion process described in \citet{Palmer2009}. This Gaussian process is parameterized in the spectral domain according to
\begin{equation}
    \label{eq:spherical_diffusion}
    z_n = z(x, t_{n}) = \phi \, z(x, t_{n-1}) + \sum_{\ell\in\mathbb{N}}\sum_{|m| \leq \ell} \sigma_\ell \, \eta_\ell \, Y^m_\ell(x).
\end{equation}
The source of randomness is the random variable $\eta_\ell \sim \mathcal{N}(0,1)$ and follows a standard Gaussian distribution while $\sigma_\ell$ and $\phi$ determine the variance of individual spatial and temporal length-scales, parameterized according to
\begin{subequations}
\begin{align}
    \phi &= e^{-\lambda}, \\
    \sigma_\ell &= F_0 \, e^{-\frac{k_T}{2} \, \ell (\ell+1)}, \\
    F_{0} &= \sigma \, \sqrt{\frac{2 \pi \, (1-\phi^2)}{ \sum_{\ell > 0} (2 \ell + 1) \, e^{-k_T \, \ell (\ell+1)}}}.
\end{align}
\end{subequations}
The variance of the process is determined by $\sigma$, whereas $\lambda$ and $k_T$ determine temporal and spatial length-scales respectively.

\subsection{Implementation in PyTorch}
\label{sec:implementation_torch_harmonics}
The integration of the aforementioned algorithms and methodologies into modern deep-learning pipelines requires efficient and differentiable GPU implementations with interfaces to deep-learning frameworks such as PyTorch \cite{Paszke2019}. 
We develop \texttt{torch-harmonics}, a library for differentiable signal processing on the sphere, publicly available under the BSD-3 clause license at \url{https://github.com/NVIDIA/torch-harmonics}. \texttt{torch-harmonics} offers differentiable implementations of spherical harmonic transforms, vector spherical harmonic transforms, convolutions and other algorithms discussed in this chapter. To enable efficient execution and training on GPUs, these operations are implemented as efficient CUDA kernels supporting distributed computation. For details regarding their distributed implementations, we refer the reader to \Cref{sec:parallelism}.
\section{FourCastNet 3 architecture}
\label{sec:model_architecture}

FourCastNet 3 is designed with the following guiding principles: spherical geometry is respected throughout the entire network to obtain a model that treats signals as functions on the sphere, mirroring physical operators and respecting symmetry principles. Moreover, all operations are discretization-agnostic which makes it a neural operator that can be evaluated and re-trained at arbitrary resolutions/discretizations, similar to traditional numerical models. Finally, all the chosen components can be distributed with reasonable communication overheads to facilitate efficient distributed training and inference of large models.

The resulting architecture is a spherical neural operator (SNO), and based on previous works on spherical Fourier neural operators (SFNO)~\cite{Bonev2023} and localized neural operators presented in~\cite{Liu2024}. As such, FourCastNet 3 contains three main building blocks: \emph{pointwise operations} parametrized by multi-layer perceptrons (MLPs), \emph{global spherical convolutions} parametrized by the spectral convolution theorem \eqref{eq:s2_spectral_conv} and \emph{local spherical convolutions} parametrized by the discrete-continuous convolution~\eqref{eq:disco_convolution}. Together, we expect these elements to facilitate processing of a full spectrum of multiscale atmospheric dynamical signals ranging from planetary waves to local orographic flows and fronts. The following sections outline the architecture in detail.

\subsection{Inputs and outputs}
\label{sec:inputs_outputs}

\begin{table}[tb]
\rowcolors{2}{white}{gray!15}
\caption{Input and output variables to FourCastNet 3. A detailed overview of the dataset is provided in \Cref{sec:training_data}, and variable designations are specified in \Cref{tab:era5_variables}.}
\label{tab:input_output_variables}
\centering
\begin{tabular}{lp{8.8cm}rr}
\toprule
Variable Type & Channels & Input & Output \\
\midrule
\multirow{1}{*}{Atmospheric variables} & \texttt{z---}, \texttt{t---}, \texttt{u---}, \texttt{v---} and \texttt{q---} at 13 atmospheric levels & \multirow{1}{*}{$\checkmark$} & \multirow{1}{*}{$\checkmark$}\\
\multirow{1}{*}{Surface variables} & \texttt{u10m}, \texttt{v10m}, \texttt{u100m}, \texttt{v100m}, \texttt{t2m}, \texttt{msl}, \texttt{tcwv} & \multirow{1}{*}{$\checkmark$} & \multirow{1}{*}{$\checkmark$}\\
\multirow{2}{*}{Auxiliary variables} & land-sea mask land, land-sea mask sea, orography, solar cosine zenith angle & \multirow{2}{*}{$\checkmark$} & \\
\multirow{3}{*}{Noise variables} & 8 noise variables with different length scales: $k_T \in \{3.08 \cdot 10^{-5}, 1.23\cdot 10^{-4}, 4.93\cdot 10^{-4}, 1.97\cdot 10^{-3}, 7.89\cdot 10^{-3}, 3.16\cdot 10^{-2}, 1.26\cdot 10^{-1}, 5.05\cdot 10^{-1}\}$, $\lambda=1$ and $\sigma=1$ & \multirow{3}{*}{$\checkmark$} & \\
\bottomrule
\end{tabular}
\end{table}

FCN3 maps the current weather state $u_n$, containing both atmospheric and surface variables to the next output state $u_{n+1}$. In addition to these prognostic variables, the model is conditioned with auxiliary variables $a_n$, which are either static variables or easily computed from a prior state. These auxiliary variables include land-sea-masks, orography and the cosine of the zenith angle of the sun. Finally, to obtain a probabilistic map, a source of randomness is passed to the model in the form of noise variables $z_n$. These noise variables are sampled from 8 spherical diffusion processes according to \Cref{sec:spherical_diffusion_process} with various length- and time-scales.

The model then predicts the next weather state $u_{n+1}$ based on the inputs, whereas the auxiliary and random variables can be easily updated by the processes which define them. All the input and output data are defined on the equiangular grid at a resolution of $721 \times 1440$ which corresponds to an angular resolution of $0.25^\circ$. \Cref{tab:input_output_variables} provides an overview of inputs and outputs and \Cref{sec:training_data} with variable names specified in \Cref{tab:era5_variables}.

\subsection{Macro architecture}
\Cref{fig:fcn3_schematic} depicts an overview of the FCN3 architecture and its main building blocks. FCN3 is loosely based on the SFNO architecture \cite{Bonev2023} and the Fourier Neural Operator (FNO) \cite{Li2015}, which employ an encoder/decoder pair and a number of operator blocks in the latent state.
This is a common approach also encountered in vision architectures and other ML models for atmospheric sciences \cite{Liu2022, Pathak2022, Lam2022, Bi2023}.

In FCN3, the encoder resamples the input fields from a $721 \times 1440$ latitude-longitude grid to an internal, $360 \times 720$ Gaussian grid. Unlike popular transformer approaches \cite{Bi2023, Fuxi2023}, we maintain a spherical representation of the state throughout the architecture. The spherical representation is passed through a series of spherical neural operator blocks, which can be either global or local spherical neural operator blocks, based on either spectral convolutions \cite{Bonev2023} or discrete-continuous convolutions \cite{Ocampo2022, Liu2024}. Auxiliary channels and noise channels serve as conditioning input to all processor blocks and are passed in separately. The resulting final internal state is then passed through a decoder, which resamples it back to the input/output grid. The final step is a point-wise output transformation, which assures that output values of water channels lie in a physically plausible range.

\subsection{Encoder}
The atmospheric variables in the dataset are highly varied in the sense that different variables often have different spatial statistics. It is therefore not desirable to mix channel information in the encoder/decoder, as this would unnecessarily complicate the reconstruction task, by entangling spatial and channel correlations. To avoid mixing unrelated channel information, we follow an approach similar to \cite{Bi2022}, using separate encoder/decoder pairs for different variables.

More concretely, FCN3 uses three separate encoders: one for atmospheric variables, one for surface variables and one for auxiliary and noise variables. Each of these encoders uses a single discrete-continuous convolution to resample the $721 \times 1440$ input fields from their equiangular representation to an internal, down-sampled Gaussian grid, determined by a scale factor.

\begin{figure}[htb]
    \centering
    \begin{subfigure}[t]{.45\linewidth}
        \centering
        \vskip 0pt
        \includegraphics[scale=0.2]{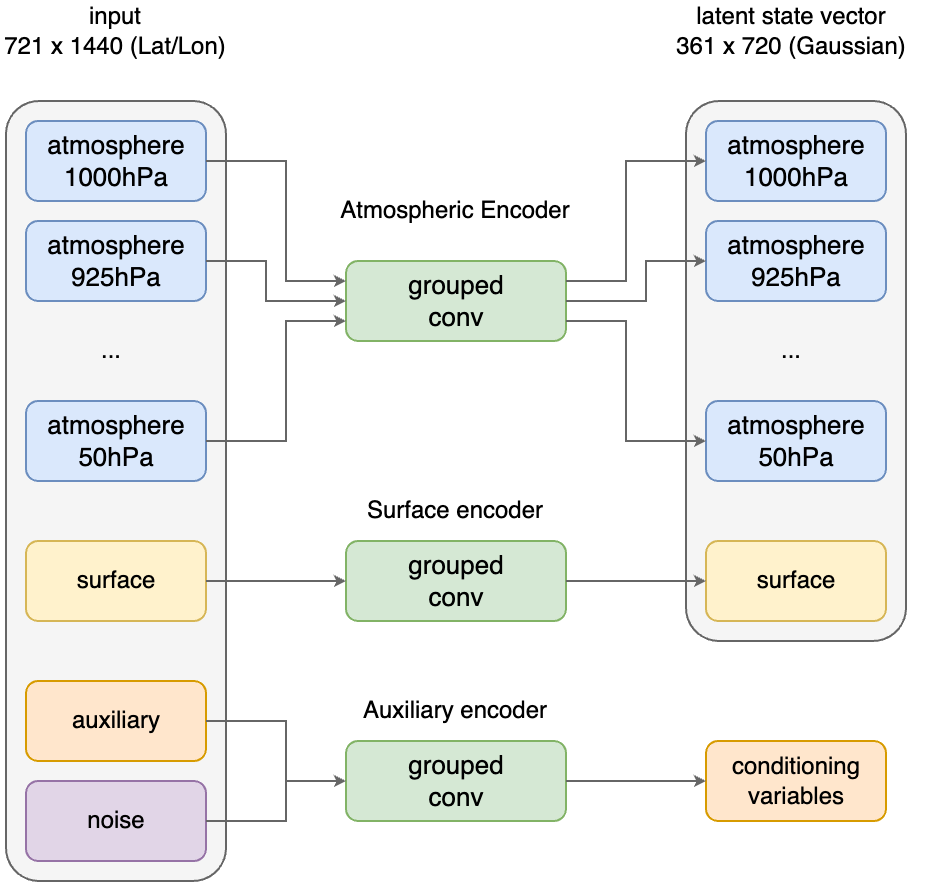}
        \caption{FCN3 encoder}
        \label{fig:fcn3_encoder}
    \end{subfigure}
    \hspace{.5cm}
    \begin{subfigure}[t]{.45\linewidth}
        \centering
        \vskip 0pt
        \includegraphics[scale=0.2]{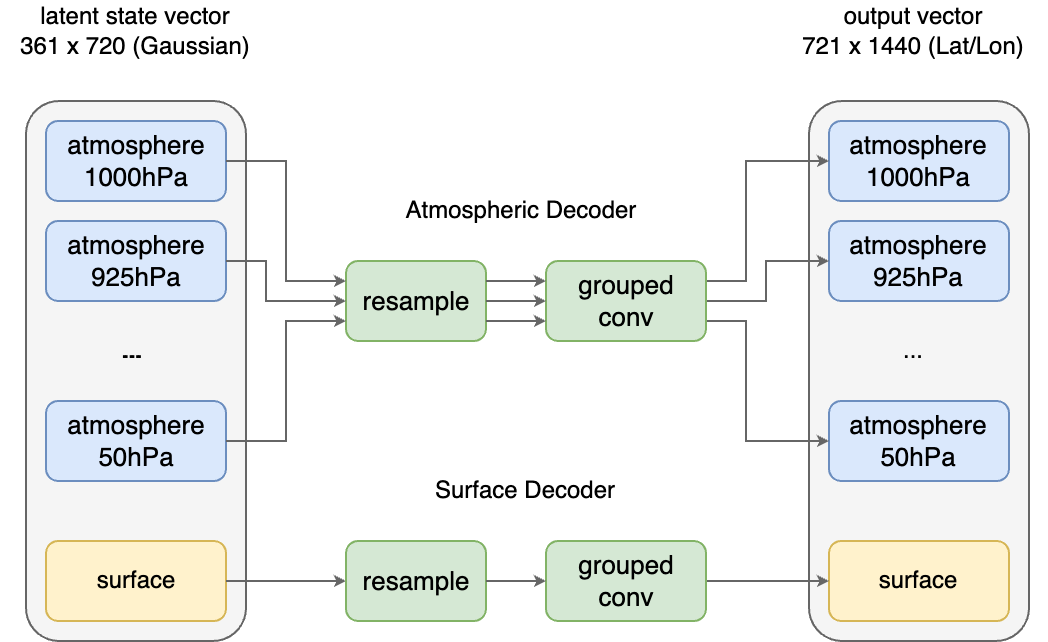}
        \caption{FCN3 decoder}
        \label{fig:fcn3_decoder}
    \end{subfigure}
    \caption{FourCastNet 3 encoder and decoder. Atmospheric, surface and auxiliary channels are all encoded individually, and the encoder is shared across pressure levels. This ensures that uncorrelated channels with distinct spectral signatures are not prematurely mixed. The resulting latent representation is grouped in terms of atmospheric levels and surface channels and consists of 641 channels at a resolution of $360 \times 720$ on the Gaussian grid. The decoder mirrors the design of the encoder, with an added bilinear spherical resampling step to avoid aliasing effects.}
    \label{fig:fcn3_encoder_decoder}
\end{figure}

To further avoid mixing channel information, grouped convolutions are used to encode each channel separately.
For the atmospheric levels, the same encoder is re-used for each pressure level, and the convolution is furthermore separated into 5 groups, such that each of the 5 atmospheric channels (\texttt{z---}, \texttt{t---}, \texttt{u---}, \texttt{v---} and \texttt{q---}) is encoded individually. This is also repeated for the surface and auxiliary channel encoders, where each channel is encoded using a distinct set of filters. This type of encoding is advantageous as the physical signals encountered in atmospheric applications are highly uncorrelated. \Cref{fig:fcn3_encoder} depicts the structure of the encoder and the resulting internal state vector.

\begin{figure}[htb]
    \centering
    \includegraphics[scale=0.2]{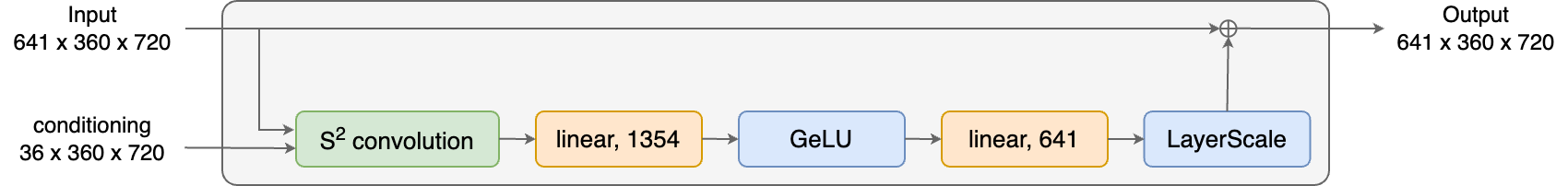}
    \caption{Illustration of a single processor block. Green blocks indicate convolutions which take spatial information into account, whereas orange and blue blocks are pointwise functions in space. Orange blocks are linear layers that mix channel information, whereas blue channels act on individual entries. The structure is an adaptation of the ConvNeXt block for spherical signals.}
    \label{fig:processor_block}
\end{figure}

\subsection{Decoder}
The decoder mirrors the encoder in its design and uses two distinct decoders for atmospheric variables and surface variables, respectively. Like the encoder, the decoder avoids channel-mixing by using grouped convolutions such that each channel is decoded separately from the corresponding channel group in the latent representation. As in the encoder, the atmospheric encoder is reused for each pressure-level. Unlike the encoder, however, the decoder first up-samples the signal to the Gaussian output grid with a $721 \times 1440$ resolution using bilinear interpolation. This is to avoid a transposed convolution for upsampling and the associated uneven overlaps, which lead to checker boarding artifacts, as previously described in \Cref{sec:interpolation}. Although upsampling can also be achieved using SHT upsampling - without filling in extra spectral information - our approach allows adding in high-frequency content, facilitating processing of the entire spectrum.
\Cref{fig:fcn3_decoder} provides an overview of the structure of the decoder. The hyperparameters for both encoder and decoder, as well as resulting embedding dimensions, are listed in \Cref{tab:hyperparameters_architecture}.

\subsection{Neural operator blocks}
The processor is made up of a sequence of neural operator blocks, which are essentially continuous-domain convolutional layers. We adapt the popular ConvNeXt architecture \cite{Liu2022} to spherical signals, which  yields the structure illustrated in \Cref{fig:processor_block}. The processor blocks receive the latent representation of the atmospheric state, and conditioning variables as input. The former is either the output of the encoder or the output of the previous layer, whereas the conditioning variables remain the same for every block, as they encode conditioning information (see \Cref{fig:fcn3_schematic}).

Both inputs are concatenated and passed through a spherical convolution. Depending on whether the processor block is global or local, the convolution can either be parametrized as a global convolution using the spectral parametrization \eqref{eq:s2_spectral_conv} or as a local convolution \eqref{eq:disco_convolution} using discrete-continuous convolutions. As such, the processor block resembles either a SFNO block \cite{Bonev2023} or a local integral neural operator block \cite{Liu2024}, depending on the chosen convolution. We alternate between local and global blocks, where for each global block we use 4 local blocks, as we find this to yield the best performance.

The convolution is followed by a dense two-layer MLP, which is applied to each grid-point on the data. Each of the output channels is then scaled by a learnable parameter \cite{layerscale2021} and added to the residual path within the network.

In contrast to ConvNext we deliberately omit layer normalization since we expect the absolute scale of inputs and outputs to be of importance for the task of predicting signals of physical systems. In practice, we observe that this leads to improved stability of the training process if the network is initialized carefully.
\begin{table}[tbp]
\rowcolors{2}{white}{gray!15}
\caption{Overview of FourCastNet 3 model hyperparameters. The encoder and decoder parameters determine the embedding space and its dimensions.}
\label{tab:hyperparameters_architecture}
\centering
\begin{tabular}{llr}
\toprule
\rowcolor{white}
\multicolumn{3}{c}{Encoder / decoder} \\
\midrule
Input/Output grid                       & & $721 \times 1440$ Lat/Lon grid \\
Internal grid                           & & $360 \times 720$ Gaussian grid \\
                                        & Atmospheric & 13 levels $\times$  5 variables\\
\rowcolor{gray!15}                      & Surface variables & 7 variables \\
\multirow{-3}{*}{Input/output channels} & Auxiliary + noise & 4 + 8 variables \\
                                        & Atmospheric & 13 levels $\times$ 45 channels\\
\rowcolor{white}                        & Surface & 56 channels\\
\rowcolor{white}                        & Auxiliary & 36 channels\\
\cmidrule{2-3}
\multirow{-4}{*}{Embedding dimensions}  & Total: & 677 channels\\
\midrule
\rowcolor{white}
\multicolumn{3}{c}{Processor blocks} \\
\midrule
\rowcolor{gray!15}                                  & Spectral convolution & 2 blocks\\
\multirow{-2}{*}{Spherical neural operator blocks}  & Local convolution & 8 blocks\\
MLPs & Hidden dimension & 1282 channels \\
\midrule
Total parameter count: & & 710'867'670 parameters\\
\bottomrule
\end{tabular}
\end{table}

\subsection{Initialization}
\label{sec:fcn3_initialization}

Due to the omission of normalization layers, extra care needs to be taken with the initialization of the model. While normalization layers can somewhat counteract a bad initialization, a network without normalization layers is prone to fall into bad local minima during training.

\begin{figure}[htb]
    \centering
    \begin{subfigure}{.45\linewidth}
        \centering
        \includegraphics[width=0.95\linewidth]{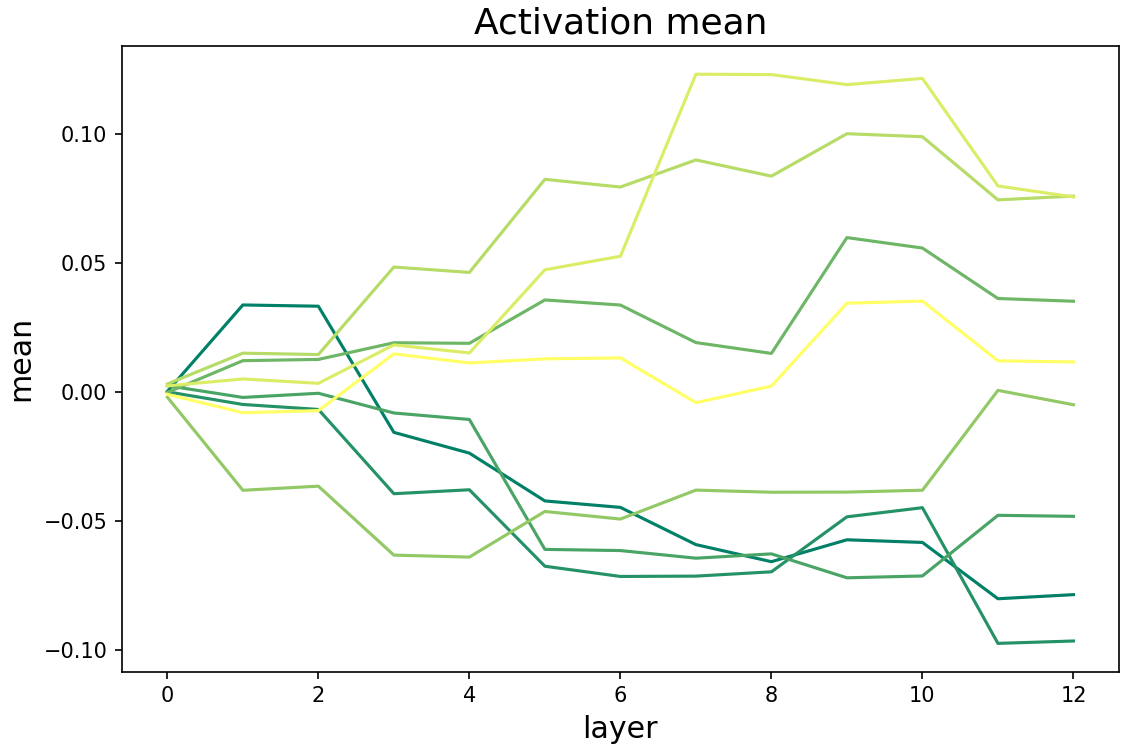}
    \end{subfigure}
    \hspace{0.5cm}
    \begin{subfigure}{.45\linewidth}
        \centering
        \includegraphics[width=0.95\linewidth]{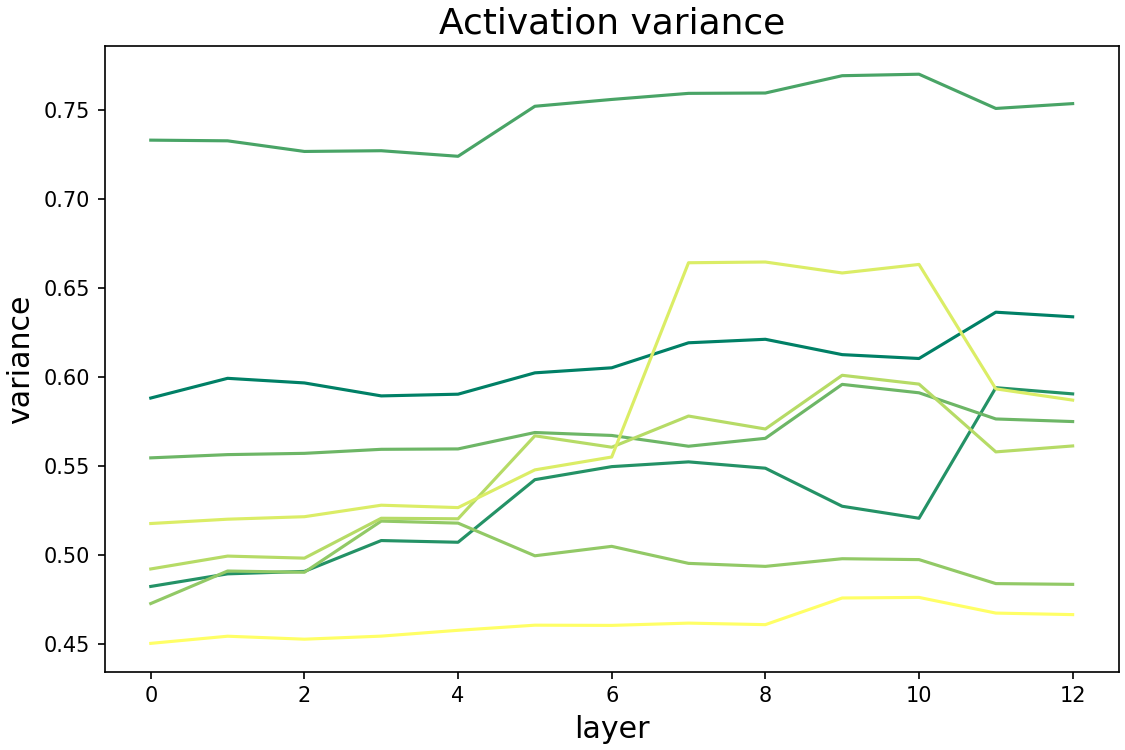}
    \end{subfigure}
    \caption{Mean and variance of activations for 8 randomly initialized instances of FourCastNet 3 and random inputs. Due to our initialization scheme and the usage of LayerSkip, the activations remain bounded despite the lack of normalization layers.}
    \label{fig:activation_magnitudes}
\end{figure}

To initialize FourCastNet 3 properly,  we follow the idea proposed by \citet{He2015} to keep the uncentered variance constant for each layer. To this end, we initialize MLPs using He initialization and adapt this initialization strategy for the spherical convolution layers.
This is sufficient to keep the magnitudes constant unless the architecture contains skip connections. \Cref{fig:activation_magnitudes} depicts the mean and variance of intermediate activations for 8 different instantiations of FCN3. Both mean and variance remain controlled throughout the layers due to our initialization scheme.

\subsection{Residual prediction and high-frequency artifacts}
Residual prediction, where the model predicts $u_{n+1} - u_{n}$ rather than $u_{n+1}$ directly, has proven to be advantageous in ML-based weather prediction \cite{Pathak2022,Lam2022,Bi2023,Siddiqui2024}. This type of prediction is equivalent to a large skip connection that bypasses the model, and often leads to better training dynamics early on \cite{Siddiqui2024}. Through experimentation however, we find this design choice to be less stable in long autoregressive rollouts. This can be attributed to two factors: Firstly, the structure of residual prediction fundamentally limits the model to update rules which resemble explicit Euler time steps. A second observation is that the residual path enables artifacts in the prediction to be passed on and amplified in autoregressive rollouts.

A possible solution for the latter is to apply a low-pass filter on the skip connection. However, this choice requires an ad-hoc cutoff frequency to be chosen and only suppresses high-frequency artifacts. Instead, we choose to omit the large skip connection altogether and rely on the internal skip connections within the operator blocks to modify the solution gradually. As such, the network is modifying the current state gradually, allowing it to filter high-frequency artifacts in autoregressive rollouts.

\subsection{Output transformations}
\label{sec:output_transformation}
Some of the predicted channels represent a quantity of water (specific humidity \texttt{q---} and total column water vapor \texttt{tcwv}). As negative values are non-physical, we pass these channels through an output transformation that has a positive codomain. A simple transformation maps all negative values to zero, while keeping positive values untouched, which corresponds to applying a ReLU activation. However, non-linear transformations are known to introduce unwanted high-frequency content \cite{Karras2021}, and the discontinuity of the ReLU activation in its derivative further amplifies this effect. To alleviate this effect, we use the once continuously differentiable spline function
\begin{equation}
    \label{eq:soft_clamp}
    \mathrm{softclamp}(u) = 
    \begin{cases}
    0, & \text{for } u \leq 0\\
    u^2, & \text{for } 0 < u \leq \frac{1}{2}\\
    u - \frac{1}{4} & \text{for } \frac{1}{2} < u
	\end{cases}
\end{equation}
to smoothly clamp the water channels to the physical range $[0,\infty)$.
\section{Evaluation metrics}
\label{sec:evaluation_metrics}

We define evaluation metrics to measure the performance of trained models. For deterministic models, the prediction $u$ is compared to the ground-truth observation $u^*$. In the probabilistic setting, an ensemble of predictions $\{u_e\}_{e=1}^{N_\text{ens}}$ is compared to a single, ground-truth observation.

To simplify the discussion, we define metrics for scalar valued variables at a single grid point on the sphere and at a single point in time. The extension to vector-valued functions $u_c(x), u_c^*(x)$ on the sphere is achieved by taking the average over the channels $c$, and spatially, over the sphere :
\begin{equation}
    \label{eq:computing_averages}
    \frac{1}{N_\text{channels}}\sum^{N_\text{channels}}_{c=1}  \frac{1}{4 \pi} \int_{S^2} \mathrm{Metric}(u_c(x), u^*_c(x)) \; \mathrm{d}\mu(x).
\end{equation}
Evaluating the metrics therefore requires computation of integrals over the sphere, which are of the form \eqref{eq:integral_sphere}. To do so, quadrature formulae as outlined in \Cref{sec:grids_and_quadrature} are used.

\subsection{Deterministic scores}
\label{sec:metrics_deterministic}
For a single prediction $u$, a straight-forward deterministic metric can be defined by using the $L_p$ distance of functions defined on the sphere. For $p=2$, we obtain the root-mean-square error (RMSE), defined by
\begin{equation}
    \mathrm{RMSE}(u, u^*) \coloneqq \frac{1}{4 \pi} \left(\int_{S^2} |u - u^*|^2 \; \mathrm{d}\mu(x)\right)^{\frac{1}{2}}.
\end{equation}
Similarly, we can set $p=1$ to obtain the $L_1$ error, or mean absolute error (MAE):
\begin{equation}
    \mathrm{MAE}(u, u^*) \coloneqq \frac{1}{4 \pi} \int_{S^2} |u - u^*| \; \mathrm{d}\mu(x).
\end{equation}
Another popular metric for deterministic forecasts is the anomaly correlation coefficient (ACC) defined as
\begin{equation}
    \label{eq:acc}
    \mathrm{ACC}(u, u^*) \coloneqq \frac{\int_{S^2} (u - \mathbb{E}_\text{clim}[u^*]) (u^* - \mathbb{E}_\text{clim}[u^*]) \; \mathrm{d}\mu(x)}{\sqrt{\int_{S^2} (u - \mathbb{E}_\text{clim}[u^*])^2 \; \mathrm{d}\mu(x) \, \int_{S^2} (u^* - \mathbb{E}_\text{clim}[u^*])^2 \, \mathrm{d}\mu(x)}},
\end{equation}
where $\mathbb{E}_\text{clim}[\cdot]$ denotes a climatological average computed from ground-truth historical data.

\subsection{Ensemble skill}
\label{sec:metrics_ensemble}
In conjunction with our deterministic scores, we wish to evaluate the skill of our probabilistic forecasts. For any deterministic metric, we can derive a simple ensemble metric, by replacing the prediction $u$ with the ensemble mean 
\begin{equation}
    \mathbb{E}_\text{ens}[u_e] \coloneqq \frac{1}{N_\text{ens}}\sum_{e=1}^{N_\text{ens}} u_e.
\end{equation}

A simple such metric is the ensemble skill or ensemble mean RMSE, which is simply the $L_2$-error/RMSE of the ensemble mean:
\begin{equation}
    \label{eq:ensemble_skill}
    \mathrm{Skill}[u_e] \coloneqq \mathrm{RMSE}[\mathbb{E}_\text{ens}[u_e], u^*] = \sqrt{\frac{1}{4 \pi}\int_{S^2} (\mathbb{E}_\text{ens}[u_e] - u^*)^2 \,\mathrm{d} \mu}.
\end{equation}

\subsection{Spread/skill ratio}
\label{sec:metrics_spread_skill}
We remark that the integrand in \Cref{eq:ensemble_skill} takes on a similar form to the ensemble variance
\begin{equation}
    \label{eq:ensemble_variance}
    \mathrm{Var}_\text{ens}[u_e] \coloneqq \frac{1}{N_\text{ens} - 1}\sum_{e=1}^{N_\text{ens}} (u_e - \mathbb{E}_\text{ens}[u_e])^2.
\end{equation}
Indeed, given an ideal ensemble, we cannot distinguish the ground truth trajectory $u^*$ from any one of the $N_\text{ens}$ ensemble members $\{u_e\}$ with $ e=1,2,\dots, N_\text{ens}$. Given this assumption of interchangeability, we expect the ensemble variance \eqref{eq:ensemble_variance} and the squared ensemble skill to be equal on average, where one ensemble member is left out to compute the ensemble mean \cite{Fortin2014}.

As such, we expect the ensemble RMSE to grow at the same rate as the ensemble spread, implying
\begin{equation}
    \label{eq:ssr_condition}
    \mathrm{Skill} \approx \sqrt{\frac{N_\text{ens}+1}{N_\text{ens}}} \, \mathrm{Spread},
\end{equation}
where the spread is defined as 
\begin{equation}
    \mathrm{Spread} \coloneqq \sqrt{\frac{1}{4 \pi}\int_{S^2} \frac{1}{N_\text{ens} - 1} \sum^{N_\text{ens}}_e (u_e - \mathbb{E}_\text{ens}[u_e])^2 \, \mathrm{d} \mu}.
\end{equation}
\Cref{eq:ssr_condition} then gives rise to the spread-skill ratio (SSR), defined as
\begin{equation}
    \label{eq:ssr}
    \mathrm{SSR} \coloneqq \sqrt{\frac{N_\text{ens}+1}{N_\text{ens}}} \, \frac{\mathrm{Spread}}{\mathrm{Skill}},
\end{equation}
which measures how well this assumption of interchangeability is satisfied.


\subsection{Continuously ranked probability score}
\label{sec:metrics_crps}
The ensemble skill, while valuable for assessing ensemble forecast performance, exclusively focuses on the ensemble mean $\mathbb{E}_\text{ens}[u_e]$ and does not indicate how accurately the ensemble estimates the ground-truth probability density. To address this limitation, we employ the continuously ranked probability score CRPS, which is a proper scoring function. Its expectation is minimized only when the ensemble members $u_e$ are sampled from the same distribution as the ground-truth observation $u^*$, providing a more comprehensive evaluation of the ensemble's probabilistic accuracy \cite{Gneiting2004}.


For a real-valued observable $u$, the continuously ranked probability score
\begin{equation}
    \label{eq:crps}
    \mathrm{CRPS}(F,u^*) \coloneqq \int_{\mathbb{R}}(F(u)-\mathds{1}(u^* \le u))^{2} \mathrm{d}u,
\end{equation}
compares the predictive cumulative density function (CDF) $F(u)$ to a single, ground-truth observation $u^*$, and its CDF, given by the indicator function $\mathds{1}(u^* \le u)$. \footnote{Some authors define CRPS with the opposite sign so that it is maximized rather than minimized with a more accurate prediction.} Intuitively, CRPS measures how well the observation is centered within the predictive CDF. For an ensemble forecast $\{u_e\}_{e=1}^{N_\text{ens}}$, the integrand in \eqref{eq:crps} can be computed numerically, by inserting the approximate CDF
\begin{equation}
    F_\text{ens}(u) = \frac{1}{N_\text{ens}}\sum_{e=1}^{N_\text{ens}} \mathds{1}(u_e \le u),
\end{equation}
which yields
\begin{align}
    \label{eq:ensemble_crps}
    \mathrm{CRPS}(F_\text{ens},u^*) &= \int_{\mathbb{R}}(F_\text{ens}(u)-\mathds{1}(u^* \le u))^{2} \mathrm{d}u \nonumber \\
    &= \int_{-\infty}^{u^*}F_\text{ens}(u)^{2} \mathrm{d}u + \int_{u^*}^\infty(F_\text{ens}(u)-1)^{2} \mathrm{d}u.
\end{align}
For a single member ensemble $u_1$, this reduces to the point-wise absolute error
\begin{equation}
    \int_{\mathbb{R}} [\mathds{1}(u_1 \le u)-\mathds{1}(u^* \le u)]^{2} \mathrm{d}u = |u_1 - u^*|.
\end{equation}
Assuming that the ensemble members $u_e$ are sorted in ascending order, the ensemble CRPS \eqref{eq:ensemble_crps} can be rewritten as
\begin{align}
    \label{eq:ensemble_crps_cdf}
    \mathrm{CRPS}(F_\text{ens},u^*)
    &= \sum_{e=1}^{e^*}\frac{2 e - 1}{N^2_\text{ens}} (u^* - u_e) + \sum_{e=e^*+1}^{N_\text{ens}}\frac{2 N_\text{ens}+1-2e}{N^2_\text{ens}} (u_e - u^*),
\end{align}
where the observation $u^*$ falls between the members $e^*$ and $e^*+1$. This can be evaluated in $\mathcal{O}(N_\text{ens} \log N_\text{ens})$ time by sorting the ensemble and finding the rank of $u^*$ within the sorted ensemble. The score can then be averaged according to \eqref{eq:computing_averages} to compute a summary score.

\citet{Gneiting2007} show that the CRPS \eqref{eq:crps} can be alternatively formulated as
\begin{align}
    \mathrm{CRPS}(F_\text{ens},u^*)
    &= \mathbb{E}_F[|U - u^*|] - \frac{1}{2} \mathbb{E}_F[|U - U'|],
\end{align}
where $U, U'$ are both independently drawn forecasts distributed according to $F$. This formulation reveals the contribution of both ensemble mean and ensemble spread to the CRPS. Moreover, it lends itself to the alternative numerical formulation
\begin{align}
    \label{eq:spread_skill_crps}
    \mathrm{CRPS}(F_\text{ens},u^*)
    &= \frac{1}{N_\text{ens}} \sum_{e=1}^{N_\text{ens}} |u_e - u^*| - \frac{1}{2N_\text{ens}^2} \sum_{e=1}^{N_\text{ens}} \sum_{i=1}^{N_\text{ens}} |u_e - u_i|
\end{align}
which may be evaluated equally efficiently as \eqref{eq:ensemble_crps_cdf} by pre-sorting the ensemble members. While the formulations \eqref{eq:ensemble_crps} and \eqref{eq:spread_skill_crps} are equivalent, they are both biased in the estimation of the ensemble spread. This can be avoided by using the fair CRPS
\begin{align}
    \label{eq:fair_spread_skill_crps}
    \mathrm{fCRPS}(F_\text{ens},u^*)
    &= \frac{1}{N_\text{ens}} \sum_{e=1}^{N_\text{ens}} |u_e - u^*| - \frac{1}{2N_\text{ens}(N_\text{ens}-1)} \sum_{e=1}^{N_\text{ens}} \sum_{i=1}^{N_\text{ens}} |u_e - u_i|,
\end{align}
which is unbiased in its estimation of the ensemble spread with respect to the number of ensemble members \cite{Zamo2018}. A detailed discussion of how CRPS is computed in a distributed, ensemble-parallel setup, can be found in \Cref{sec:distributed_crps}.
\section{Training Methodology}
\label{sec:training}

FourCastNet 3 is trained on historical best estimates of the observed state of Earth's atmosphere from the ECMWF's Earth Reanalysis 5 (ERA5) dataset. In the following, we outline the training methodology. The choice of objective function for training the architecture is outlined in \Cref{sec:objective_function}. Training consists of two stages, an initial pre-training stage and a fine-tuning stage, outlined in  \Cref{sec:pre-training} and \Cref{sec:fine-tuning}, respectively. Finally, \Cref{sec:training_data} outlines details regarding the training data and its usage.

\subsection{Objective function}
\label{sec:objective_function}
To train FourCastNet 3, we use the objective function
\begin{equation}
    \label{eq:combined_loss}
    \mathcal{L}_\text{ens}[\{u_e\}, u^*] = \sum_{n=1}^{N_\text{times}}\sum^{N_\text{channel}}_{c=1} w_c w_{\Delta t, c} w_n \left(\mathcal{L}_\text{spatial} + \lambda_\text{spectral}\mathcal{L}_\text{spectral}\right),
\end{equation}
which composes a spatially averaged, point-wise CRPS loss term $\mathcal{L}_\text{spatial}$ and a spectral CRPS loss term $\mathcal{L}_\text{spectral}$. This combined loss is averaged over the channels $c=1,2,\dots, N_\text{channel}$ with a channel weight $w_c$, that is defined according to \Cref{tab:era5_variables}. Furthermore, to account for the varying time scales of different variables, an additional temporal weight factor $w_{\Delta t, c}$ is used. $w_{\Delta t, c}$ is defined as the inverse of the spatial and climatological standard deviation of 1-hourly differences of each channel:
\begin{equation}
    w_{\Delta t, c} = \frac{1}{\mathbb{E}_\text{clim}\left[\mathbb{E}_\text{spatial}[u_c(x, t_{n+1}) - u_c(x, t_n)]\right]}.
\end{equation}
This approach to channel-weighting was originally reported in \cite{Lam2022} and carefully balances how much individual predicted variables contribute to the overall loss function. In the case of autoregressive training, the loss is further averaged over lead times $t_1, t_2, \dots$ with the corresponding weight $w_n$.

The individual loss terms are based on the ensemble CRPS \eqref{eq:ensemble_crps} to achieve a probabilistic model. The spatial loss term is implemented as
\begin{equation}
    \label{eq:spatial_loss}
    \mathcal{L}_\text{spatial}[\{u_e\}, u^*] = \frac{1}{4\pi}\int_{S^2} \text{CRPS}[\{u_e(x, t_{n})\}, u^*(x, t_n)]\; \mathrm{d}\mu(x),
\end{equation}
and averages the point-wise CRPS over space. The spatial average is implemented using a quadrature rule as outlined in \Cref{sec:grids_and_quadrature}. 

The analytically evaluated CRPS \eqref{eq:ensemble_crps} is uniquely minimized if the forecast ensemble members $u_e$ are drawn from the same distribution, as the ground truth observation $u^*$. However, this is not the case for summary scores such as \Cref{eq:spatial_loss}. This can be demonstrated by considering an ensemble drawn from the ground-truth distribution and then randomly shuffling its members at each spatial point. As a result, individual members will become implausible, while the summary score remains unchanged with the CRPS remaining the same at each point. This issue is mainly due to the CRPS not observing the full, high-dimensional CDFs but rather a marginal CDF at the given location, variable and lead-time.

We address this issue by adding a spectral loss term
\begin{equation}
    \label{eq:spectral_loss}
     \mathcal{L}_\mathrm{spectral}[\{u_e\}, u^*] = \sum^{n_\text{lat}/2}_{\ell=1} \sum_{|m|\leq \ell} \mathrm{CRPS}[\{\hat{u}^m_\ell(t_{n})\}, u^*(x, t_n)],
\end{equation}
which accounts for spatial correlations by computing the CRPS of each spectral coefficient. A similar approach was also suggested by \citet{Kochkov2023}, who additionally apply a low-pass filter to account for the lower predictability of higher frequencies.
As the probabilistic nature is already accounted for by the CRPS loss term, we opt to directly optimize the spectral CRPS across all frequencies and variables, to match the distribution of the ground truth solution equally across all angular modes.

We remark that the fair version of CRPS \eqref{eq:fair_spread_skill_crps} is unbiased in its estimation of the spread term, and can therefore be problematic as a training objective. For a two-member ensemble, we observe that if $u_1$ is equal to the ground-truth $u^*$, the fair CRPS \eqref{eq:fair_spread_skill_crps} becomes exactly 0, irrespective of the value of $u_2$. As a consequence, $u_2$ can become unbounded. This has also been observed by \citet{Lang2024}, who proposed to use a mixture of biased CRPS \eqref{eq:spread_skill_crps} and the fair CRPS \eqref{eq:fair_spread_skill_crps} for training. In practice, we find that the fair CRPS occasionally leads to instabilities only during the early stages of training. As such, we make use of the CRPS objective in the initial stages with a large ensemble size to minimize the underestimation of the spread, and later switch to the fair version with a smaller ensemble.

\subsection{Pre-training}
\label{sec:pre-training}

\begin{table}[t]
\rowcolors{2}{white}{gray!15}
\caption{Overview of the training regime. Initial pre-training is conducted using a large batch size and a high learning rate (LR), emphasizing good performance on short lead times. In fine-tuning, the focus is shifted towards achieving a better approximation for autoregressive rollouts. To this end, batch size is traded for higher model-parallelism to enable more autoregressive steps and suppress buildup of autoregressive errors.}
\label{tab:training_configuration}
\centering
\begin{tabular}{lccc}
\toprule
& \multicolumn{2}{c}{pre-training} & fine-tuning\\
\cmidrule{2-3}
& \multicolumn{1}{c}{stage 1} & \multicolumn{1}{c}{stage 2} &  \\
\midrule
\multirow{2}{*}{Training dataset} & \multicolumn{1}{p{3.5cm}}{\centering 1-hourly, 1980-2016, 332,800 samples} & \multicolumn{1}{p{3.5cm}}{\centering 6-hourly, 1980-2016, 55,460 samples} & \multicolumn{1}{p{3.5cm}}{\centering 6-hourly, 2012-2016, 5840 samples}\\
Objective function              & \multicolumn{2}{c}{nodal + spectral CRPS} & nodal + spectral fCRPS\\
Gradient descent steps          & 208,320 & 5,040 & 4,380\\
Optimizer                       & \multicolumn{3}{c}{ADAM \cite{Kingma2014}} \\
Learning rate scheduler         & constant & halve every 840 steps & halve every 1,095 steps\\
Initial learning rate           & $5 \cdot 10^{-4}$ & $4 \cdot 10^{-4}$ & $4 \cdot 10^{-6}$ \\
Rollout steps                   & 1 & 4 & 8 \\
Batch size                      & 16 & 32 & 4 \\
Ensemble size                   & 16 & 2 & 4 \\
Model parallelism               & \texttt{lat}=2, \texttt{lon}=2 & \texttt{lat}=2, \texttt{lon}=4 & \texttt{lat}=4, \texttt{lon}=4\\
Number of GPUs                  & 1024 & 512 & 256 \\
\bottomrule
\end{tabular}
\end{table}

The initial training phase emphasizes general model performance and is carried out in two distinct stages:

In the first stage, the model is trained to achieve the best single step loss on the hourly version of the ERA5 dataset ranging from 1979 to 2016, which amounts to a total of 332,800 samples. At this stage, we use the biased CRPS loss function \eqref{eq:combined_loss}, as we find the fair CRPS results in occasional instabilities during training at this stage. This can  be explained by its ambiguity property, which can lead to an unbounded spread \cite{Lang2024}. To compensate for the bias in the biased CRPS training objective, we use 16 ensemble members for training. In conjunction with a batch size of 16, and a spatial model-parallelism of 4 GPUs, this amounts to a total of 1024 NVIDIA H100 GPUs for the initial training. This pre-training stage took 78 hours with an average step time of 1.34 seconds on the NVIDIA Eos Supercomputer.

The second pre-training stage emphasizes autoregressive performance on the 6-hourly initial conditions at UTC times 00:00, 06:00, 12:00, 18:00. To this end, the model is optimized using four autoregressive steps on the 6-hourly subset of the same dataset, which amounts to 55,400 samples. To accommodate for the increased memory requirement, the model and the data are split spatially across 8 ranks.\footnote{Further detail on the simultaneous model- and data-parallelism are provided in \Cref{sec:parallelism}.} For this second stage, we switch to the fair version of our CRPS objective and 2 ensemble members at a batch size of 32. This training stage was carried out on 512 NVIDIA A100 GPUs on the National Energy Research Scientific Computing Center's Perlmutter Supercomputer and took 15 hours to complete.

\subsection{Fine-tuning}
\label{sec:fine-tuning}
Once pre-training is concluded, the model is fine-tuned using 8 autoregressive steps on the last 5 years 2012-2016 of the 6-hourly dataset. The degree of model-parallelism is once more increased to 16 ranks to accommodate the increased memory footprint of the autoregression. Finally, we find it beneficial to enable noise-centering for this last stage of training. To this end, odd ensemble members use the same noise vector as the even ensemble members but are multiplied by $-1$. We find that this choice increases performance both in fine-tuning and inference later on. Fine-tuning took 8 hours on 256 NVIDIA H100 GPUs on NVIDIA's Eos Supercomputer.

All model training was carried out using automatic mixed precision in the \texttt{bf16} format and using the ADAM optimizer \cite{Kingma2014}. Key training parameters for all training stages are summarized in ~\Cref{tab:training_configuration}. 

\subsection{Training data}
\label{sec:training_data}

FourCastNet 3 is trained on the ERA5 dataset, which is a multi-decade, hourly estimate of the Earth's atmospheric state \cite{Hersbach2020}. It is the result of an elaborate data-assimilation and reanalysis process which combines modern numerical weather forecasting methods with historical records of observational data to produce an accurate estimate of the ocean-atmosphere system \cite{Carrassi2017}. As such, ERA5 spans multiple decades while retaining consistent and unchanging dynamics from a single, modern numerical model. This contrasts with operational analysis datasets, which undergo periodic updates due to advancements in numerical methods, computational techniques, and our understanding of geophysical processes. Moreover, raw observations of the Earth's ocean-atmosphere system are sparsely distributed in time and space and vary in quality. Reanalysis datasets like ERA5 assimilate various data sources, considering the uncertainty estimates of respective sources. This approach provides a consistent representation of the Earth's atmospheric history, making it an ideal target for approximating planetary atmospheric dynamics using machine learning models \cite{Rasp2018, Arcomano2020, Weyn2021, Pathak2022, Lam2022, Bonev2023, Bi2023}.

For the purpose of training FCN3, we use a subset of 72 variables from the original dataset, listed in \Cref{tab:era5_variables}. This includes seven surface-level variables and five atmospheric variables at 13 pressure levels. Despite training a 6-hourly model, we use data sampled at an hourly rate between 1980 and 2018 to maximize the size of the dataset, boosting the generalization capabilities of our model. The data is represented on the default $721 \times 1440$ latitude-longitude grid, which amounts to a spatial resolution of roughly $0.25^\circ$. This amounts to a total of 39.5TB of data.

\begin{table}[t]
\rowcolors{2}{gray!15}{white}
\caption{Atmospheric and surface variables predicted by our model. Detailed specifications of each variable can be accessed at \url{https://apps.ecmwf.int/codes/grib/param-db}. The weighting factor of each variable is specified by $w_c$.}
\label{tab:era5_variables}
\centering
\begin{tabular}{llrrr}
\toprule
\rowcolor{white}
Channel        & Description & ECMWF ID & Normalization & $w_c$\\

\midrule
\rowcolor{white}
\multicolumn{5}{c}{Surface variables} \\
\midrule

\texttt{10u}    & 10 meter $u$-wind component                                       & 165       & z-score & $0.1$ \\
\texttt{10v}    & 10 meter $v$-wind component                                       & 166       & z-score & $0.1$ \\
\texttt{100u}   & 100 meter $u$-wind component                                      & 228246    & z-score & $0.1$ \\
\texttt{100v}   & 100 meter $v$-wind component                                      & 228247    & z-score & $0.1$ \\
\texttt{t2m}    & 2 meter temperature                                               & 167       & z-score & $1.0$ \\
\texttt{msl}    & Mean sea level pressure                                           & 151       & z-score & $0.1$ \\
\texttt{tcwv}   & Total column vertically-integrated water vapor                    & 137       & min/max & $0.1$ \\

\midrule
\rowcolor{white}
\multicolumn{5}{c}{Atmospheric variables at pressure level $p$ indicated by \texttt{---} in hPa} \\
\midrule

\texttt{z---}   & Geopotential                                                      & 129       & z-score & $p \cdot 10^{-3}$\\
\texttt{t---}   & Temperature                                                       & 130       & z-score & $p \cdot 10^{-3}$\\
\texttt{u---}   & $u$ component of the wind                                         & 131       & z-score & $p \cdot 10^{-3}$\\
\texttt{v---}   & $v$ component of the wind                                         & 132       & z-score & $p \cdot 10^{-3}$\\
\texttt{q---}   & Specific humidity                                                 & 133       & min/max & $p \cdot 10^{-3}$\\
\bottomrule
\end{tabular}
\end{table}

To train the model, the dataset is split into three parts: the training dataset, spanning 1980 to 2016, the test data spanning 2017 and the out-of-sample validation dataset 2018-2021. The latter is used for validation purposes only, and all the reported metrics are computed on the validation dataset 2020. For the purpose of training, the data is either z-score or min-max normalized. In particular, min-max normalization is chosen for the water-channels (\texttt{q---} and \texttt{tcwv}), which is required for the output normalization (see \Cref{sec:output_transformation}) to work. The normalization constants, i.e. standard deviation, bias, minimum and maximum are spatially averaged over the sphere and then averaged over the entire training dataset. To preserve the direction of wind channels (\texttt{u---} and \texttt{v---}), we assume them to have zero mean and normalize with the standard deviation of the total wind velocity magnitude.
\section{Results}
\label{sec:results}

We discuss the performance of the results FourCastNet 3 model. \Cref{sec:scoring_methodology} describes the scoring methodology. The main probabilistic characteristics of FCN3 are presented in \Cref{sec:probabilistic_scores,sec:ensemble_calibration}. \Cref{sec:deterministic_scores} discusses the deterministic performance of a single ensemble member. Finally, \Cref{sec:individual_predictions,sec:bias,sec:spectral_properties} are concerned with the physical realism and stability of individual ensemble members of FCN3.

\subsection{Scoring methodology}
\label{sec:scoring_methodology}

We score FourCastNet 3 using the procedure outlined in WeatherBench 2 (WB2) \cite{Rasp2023}. However, several limitations in its implementation hinder its scalability for evaluating large ensemble predictions. To address these issues, we develop an alternative implementation in an online fashion, which minimizes I/O operations. Our scoring scripts and inference codes are publicly available at \url{https://github.com/NVIDIA/makani}. This approach is more compatible with the low-cost inference characteristic of ML models and the architecture of tightly coupled high-performance computing (HPC) systems. For details regarding the distributed online inference implementation, we refer the reader to \Cref{sec:distributed_inference}.

Following the evaluation protocol outlined in WB2 \cite{Rasp2023}, we evaluate the skill of FCN3 on 12-hourly initial conditions on the out-of-sample year 2020, starting from 2020-01-01 00:00:00 UTC and ending at 2020-12-31 12:00:00 UTC. Some predictions at the end of the year spill into 2021. For these, data from 2021 is used to compute the metrics.

Unless specified otherwise, the evaluation is performed on a 50 member ensemble and compared to the IFS ensemble of equal size. We remark that the IFS-ENS and GenCast baselines are taken from WB2, which performs scoring against the ARCO-ERA5 dataset (available at \url{https://github.com/google-research/arco-era5}). This ARCO-ERA5 dataset differs slightly from the official ERA5 dataset, which may therefore result in slight discrepancies when scoring against it. The computation of metrics is adapted to match the evaluation of WB2. This includes the use of the unbiased formulations of the CRPS \eqref{eq:fair_spread_skill_crps} and SSR \eqref{eq:ssr} for all reported results. Moreover, spatial integration is performed using the trapezoidal quadrature rule as outlined in WB2.

\subsection{Probabilistic scores}
\label{sec:probabilistic_scores}

\begin{figure}[tbp]
    \centering
    \includegraphics[width=1.0\linewidth]{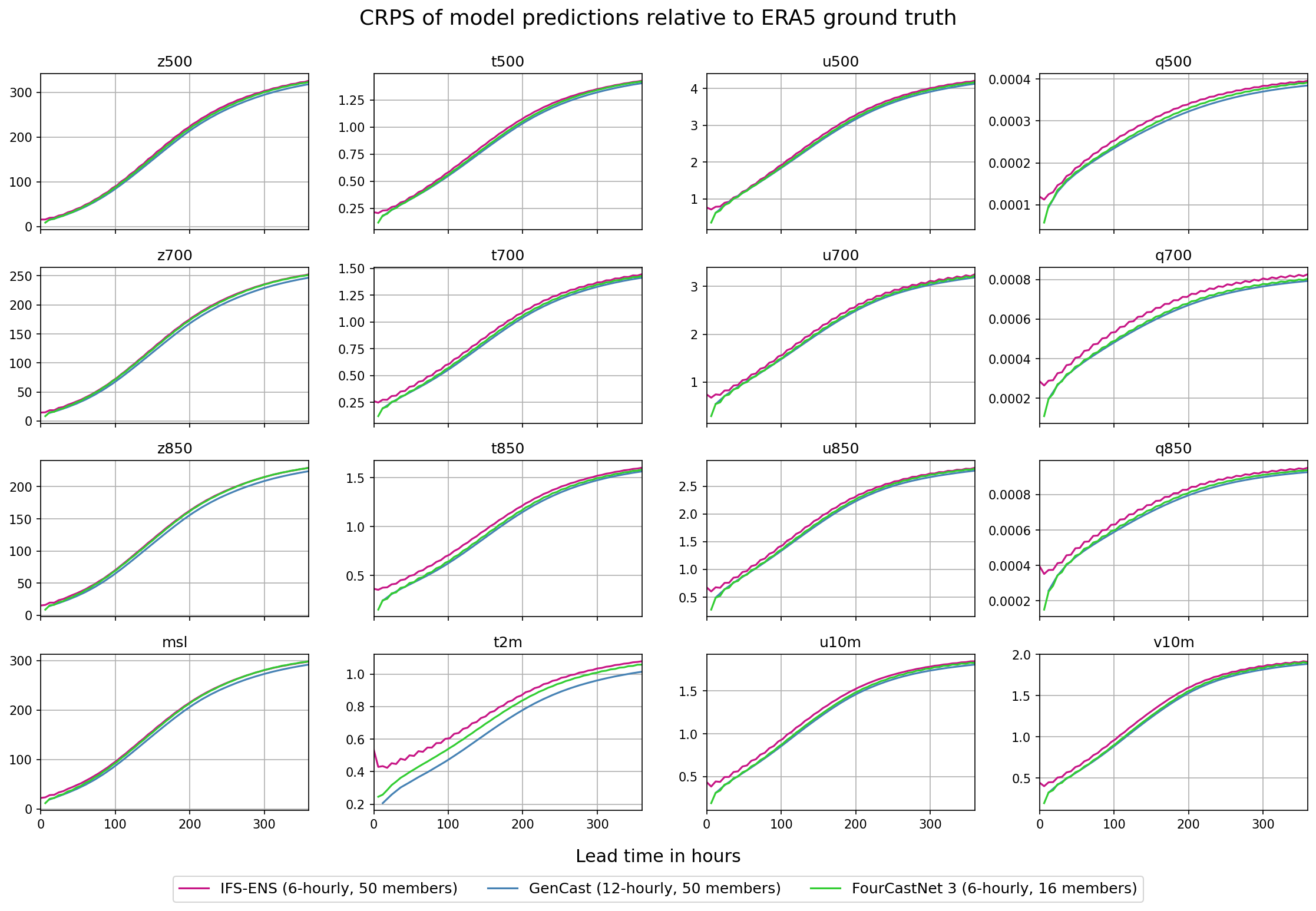}
    \caption{Continuously ranked probability scores (CRPS) averaged over initial conditions at 00:00:00 UTC and 12:00:00 UTC ranging from 2020-01-01 to 2020-12-31. Lower scores indicate better skill.}
    \label{fig:crps_2020_ens50}
\end{figure}

\begin{figure}[tbp]
    \centering
    \includegraphics[width=1.0\linewidth]{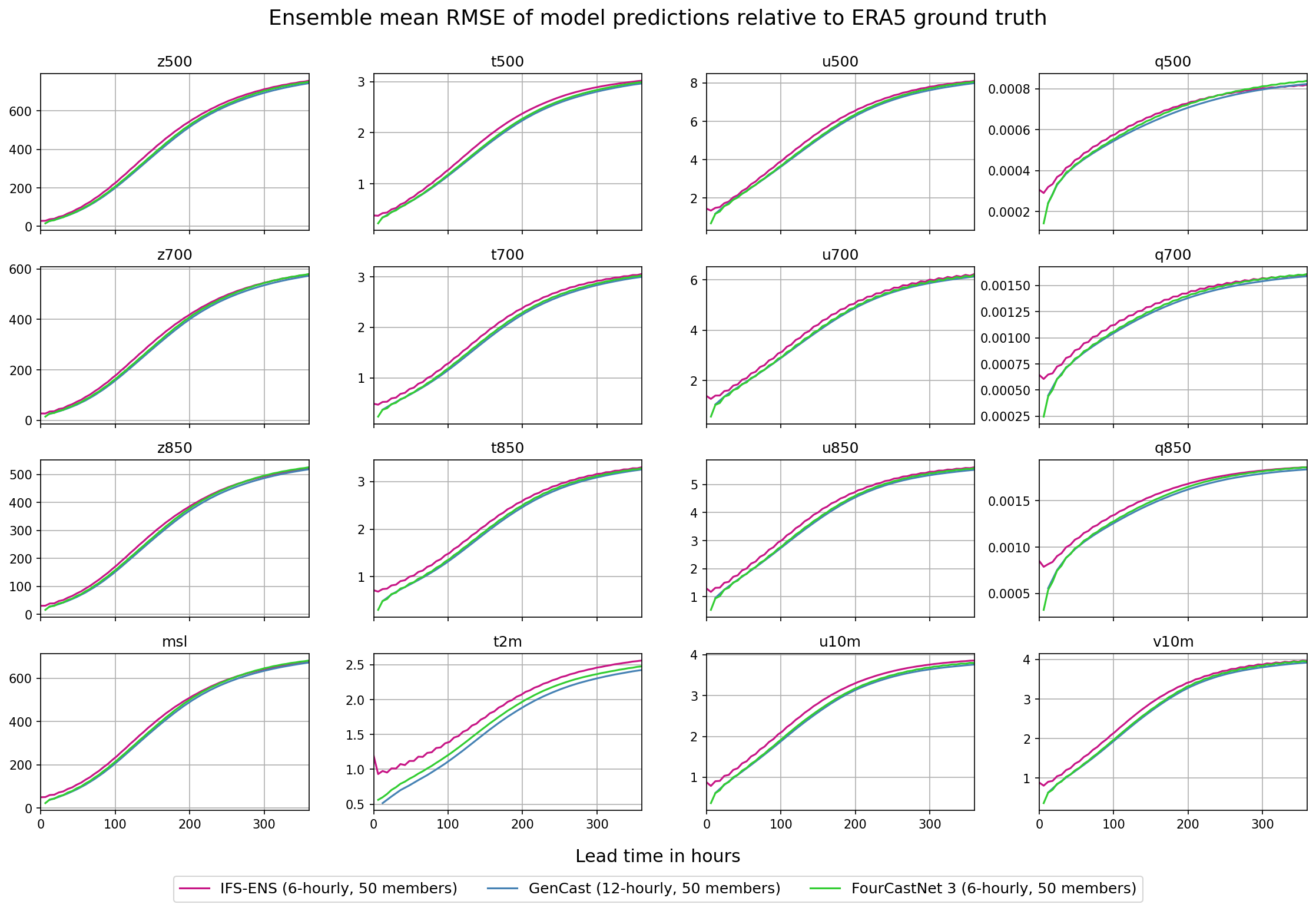}
    \caption{Ensemble-mean root mean squared errors (RMSE), averaged over initial conditions at 00:00:00 UTC and 12:00:00 UTC ranging from 2020-01-01 to 2020-12-31. Lower scores indicate better skill.}
    \label{fig:rmse_2020_ens50}
\end{figure}

We evaluate the probabilistic skill of FourCastNet 3 as outlined in WB2. \Cref{fig:crps_2020_ens50} and \Cref{fig:rmse_2020_ens50} depict the resulting, averaged CRPS and ensemble-mean RMSE scores. FourCastNet 3 achieves competitive skill on the 16 variables contained in the scoring protocol, outperforming ECMWF's gold-standard NWP ensemble forecasting system IFS-ENS, with a computational cost 60x lower than that of IFS-ENS.
Even though FCN3 is a 6-hourly model trained with data up to 2016, its performance is competitive with GenCast \cite{Price2024}, the state-of-the-art diffusion model, which is a 12-hourly model trained with data up to 2019. On 15 out of the 16 channels in the scoring protocol, FCN3 matches and even slightly outperforms Gencast's forecasts at short lead times. A notable exception is 2 meter temperature \texttt{t2m}, where we observe better performance with GenCast. The slightly better rollout skill of GenCast may be attributable to its larger 12-hour timestep. This reduces the number of model evaluations and thus the accumulation of autoregressive. More importantly, it also limits training to inputs and targets from the 00:00 and 12:00 UTC data assimilation windows, when ERA5 fields are typically most accurate \cite{Lam2022}.

The performance of FCN3 is achieved at a fraction of the computational cost of GenCast. At double the temporal resolution, a single 15-day FCN3 rollout is computed in 60 seconds on a single NVIDIA H100, an 8x speed-up over the 8 minutes required to generate a 15-day GenCast forecast on a Google TPU v5 instance \cite{Price2024}. These speed-ups over competing methods not only enable faster forecasts, but also larger ensemble forecasts with more accurate estimation of tail events \cite{Mahesh2024b}. Similarly, better ensemble-mean RMSE scores can be achieved with larger ensemble forecasts.

\begin{figure}[tbp]
    \centering
    \includegraphics[width=1.0\linewidth]{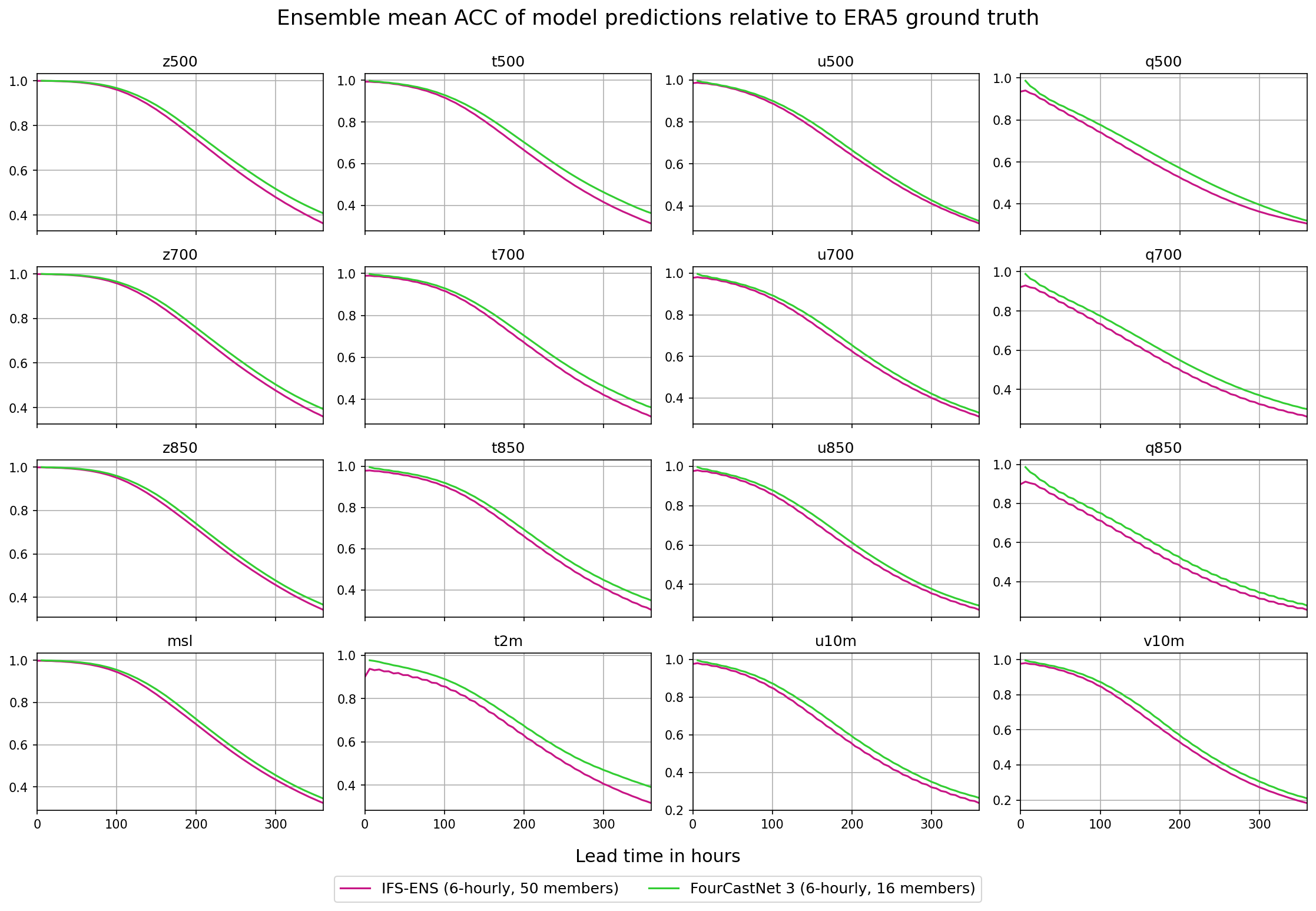}
    \caption{Ensemble-mean anomaly correlation coefficient (ACC), averaged over initial conditions at 00:00:00 UTC and 12:00:00 UTC ranging from 2020-01-01 to 2020-12-31. Higher scores indicate better skill.}
    \label{fig:acc_2020_ens50}
\end{figure}

Finally, \Cref{fig:acc_2020_ens50} depicts the anomaly correlation coefficients (ACC) \eqref{eq:acc} for the ensemble mean predictions of the 50-member FCN3 ensemble.  The ACC is a normalized measure of how well the anomalies within the predictions are spatially correlated with the anomalies in the ground-truth observations (see \Cref{sec:metrics_deterministic}). As such, ACC scores of 1 represent a perfect prediction, whereas scores of 0.0 indicate a prediction of similar quality to the climatology. ACC scores above 0.55 are generally considered as skillful. We observe that the FCN 3 ensemble outperforms IFS-ENS once again

\subsection{Ensemble calibration}
\label{sec:ensemble_calibration}

\begin{figure}[tbp]
    \centering
    \includegraphics[width=1.0\linewidth]{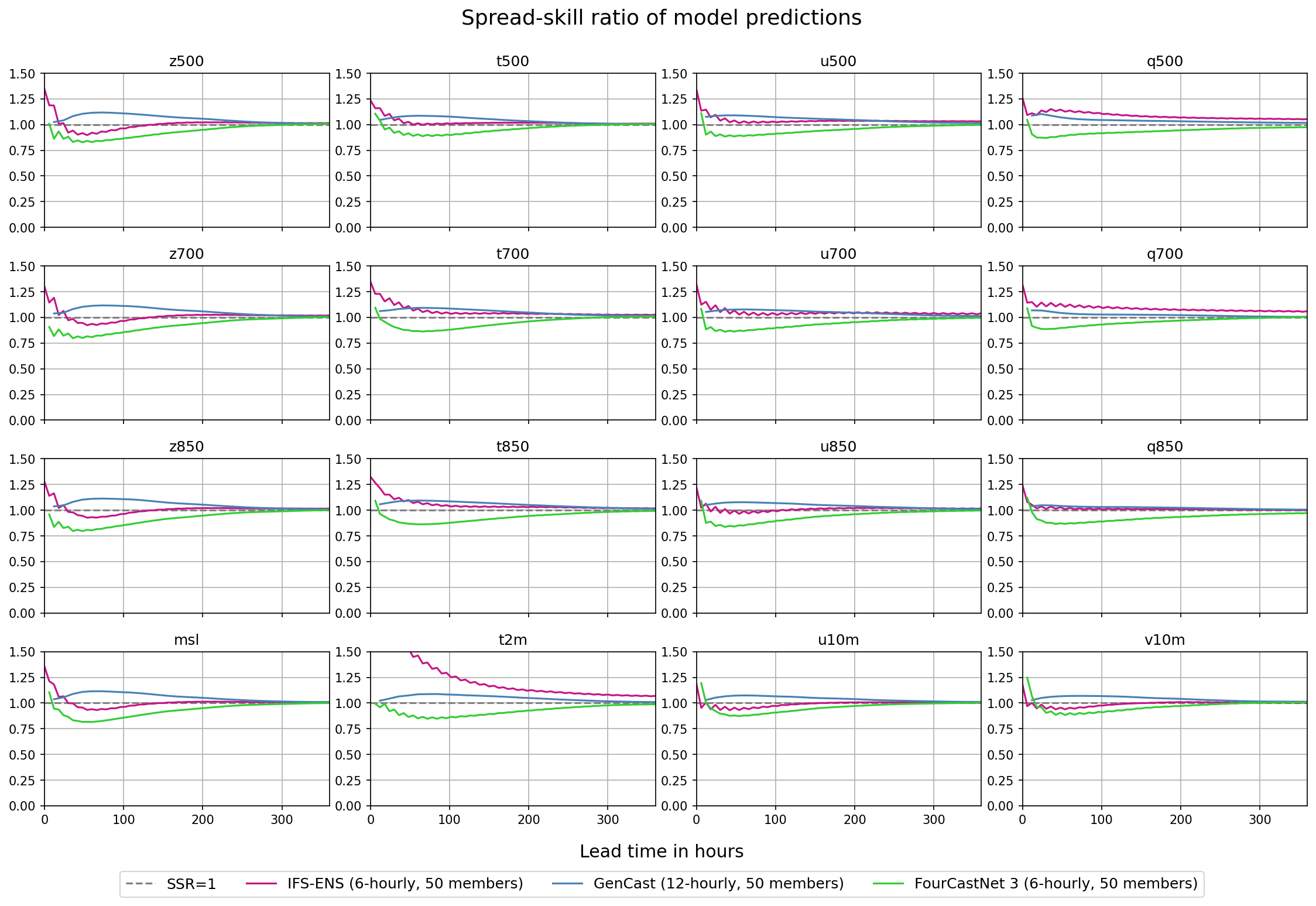}
    \caption{Spread-skill ratios (SSR), averaged over initial conditions at 00:00:00 UTC and 12:00:00 UTC in the range from 2020-01-01 to 2020-12-31. Spread-skill ratios close to 1 are desirable, as they indicate good calibration and therefore interchangeability of observations and predictions.}
    \label{fig:ssr_2020_ens50}
\end{figure}

The aforementioned scores provide a quantitative measure of the predictive accuracy of the FCN3 ensemble. However, they do not evaluate the calibration of the ensemble predictions. The spread-skill ratio \eqref{eq:ssr} quantifies the agreement between the ensemble spread and the error relative to the ground-truth observation. Under the assumption of interchangeability we expect this ratio to be 1 on average. \Cref{fig:ssr_2020_ens50} shows the mean SSR over the inference period in 2020. The FCN3 ensemble attains SSR values near 1, which indicates good calibration. Similar to the classical IFS ensemble, FCN3 is initially overdispersive. For lead times between 24 and 200 hours, FCN3 becomes underdispersive and gradually converges towards a spread-skill ratio of 1. The initial underdispersion relative to the IFS ensemble may be attributed to the absence of initial condition uncertainty in FCN3. Introducing initial condition perturbations, such as bred-vectors \cite{Mahesh2024a}, could potentially improve FCN3's dispersion characteristics. Finally, FCN3 and GenCast display distinct dispersion behaviors, with GenCast being predominantly overdispersive. This difference likely arises from the distinct training objectives of the two models.

\begin{figure}[tbp]
    \centering
    \includegraphics[width=1.0\linewidth]{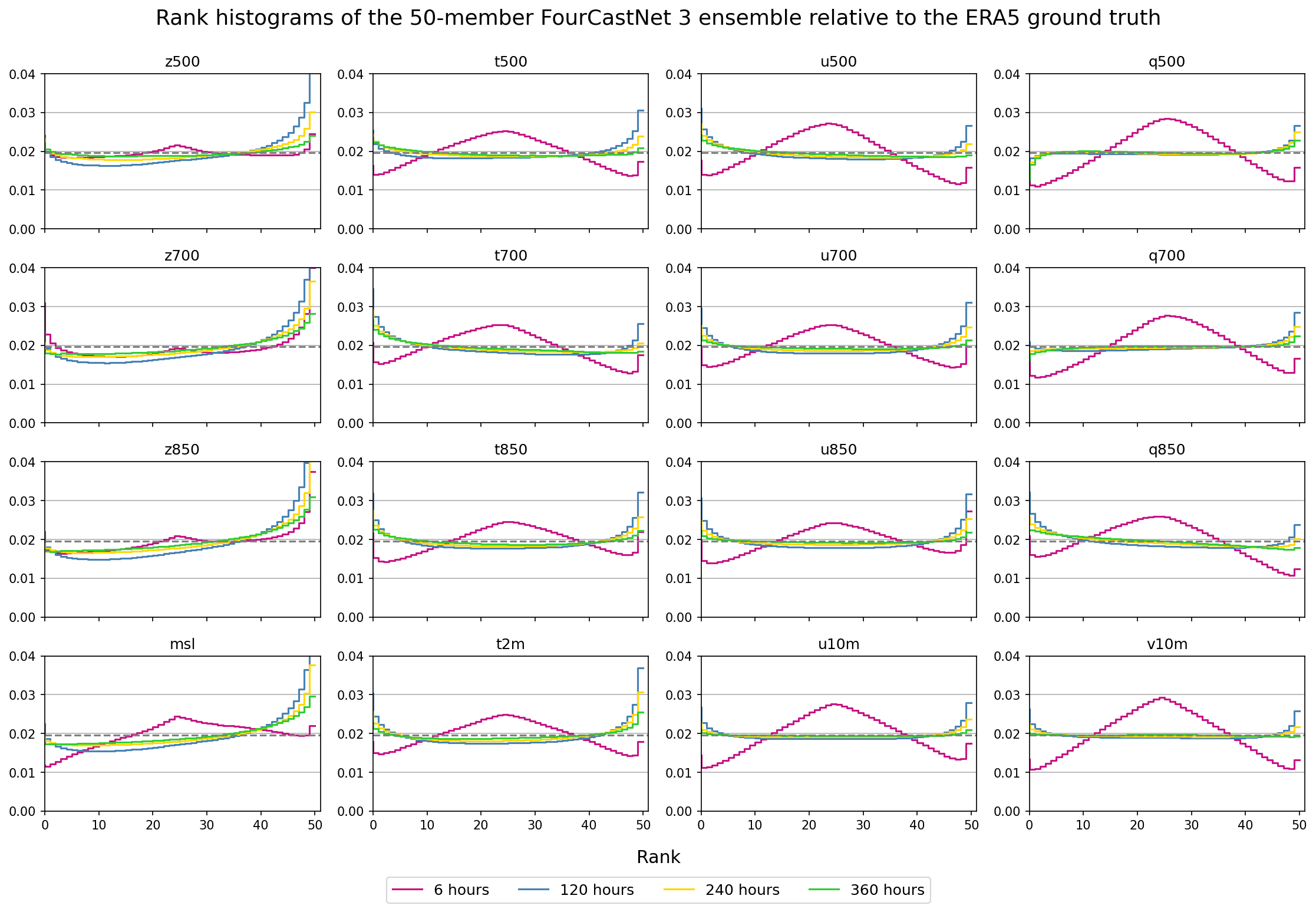}
    \caption{Rank histograms of the ERA5 ground truth within the 50-member FourCastNet 3 ensemble at different lead times. The rank histograms are spatially and temporally averaged for daily initial conditions at 00:00:00 UTC and 12:00:00 UTC ranging from 2020-01-01 to 2020-12-31. The dashed line depicts the optimal uniform distribution at a value of $1/51$.}
    \label{fig:rh_2020_ens50}
\end{figure}

Besides the spread-skill ratio, rank histograms are an essential tool for assessing the calibration of ensemble forecasting systems \cite{Hamill2001}. An ideally calibrated ensemble is expected to result in a uniform distribution of the observation's rank within the ensemble.\footnote{\citet{Hamill2001} provide a useful guide to understanding and interpreting rank histograms.} \Cref{fig:rh_2020_ens50} depicts the rank histogram of the ERA5 ground truth observation within the 50-member FCN3 ensemble. The histograms are spatially and temporally averaged with the correct weighting to account for the spherical geometry. We observe that the initial predictions at 6 hours tend to be over-dispersive, leading to a concave rank distribution. At lead times of 24 hours and more, this trend is reversed, and we observe the rank distribution indicating an underdispersive ensemble with a concave rank histogram. This trend then gradually flattens out, approaching the ideal, uniform distribution. We note, that there appears to be a slight bias in the temperature and geopotential predictions predictions, indicated by a slight asymmetry in the respective rank histograms.

\subsection{Deterministic scores}
\label{sec:deterministic_scores}

\begin{figure}[tbp]
    \centering
    \includegraphics[width=1.0\linewidth]{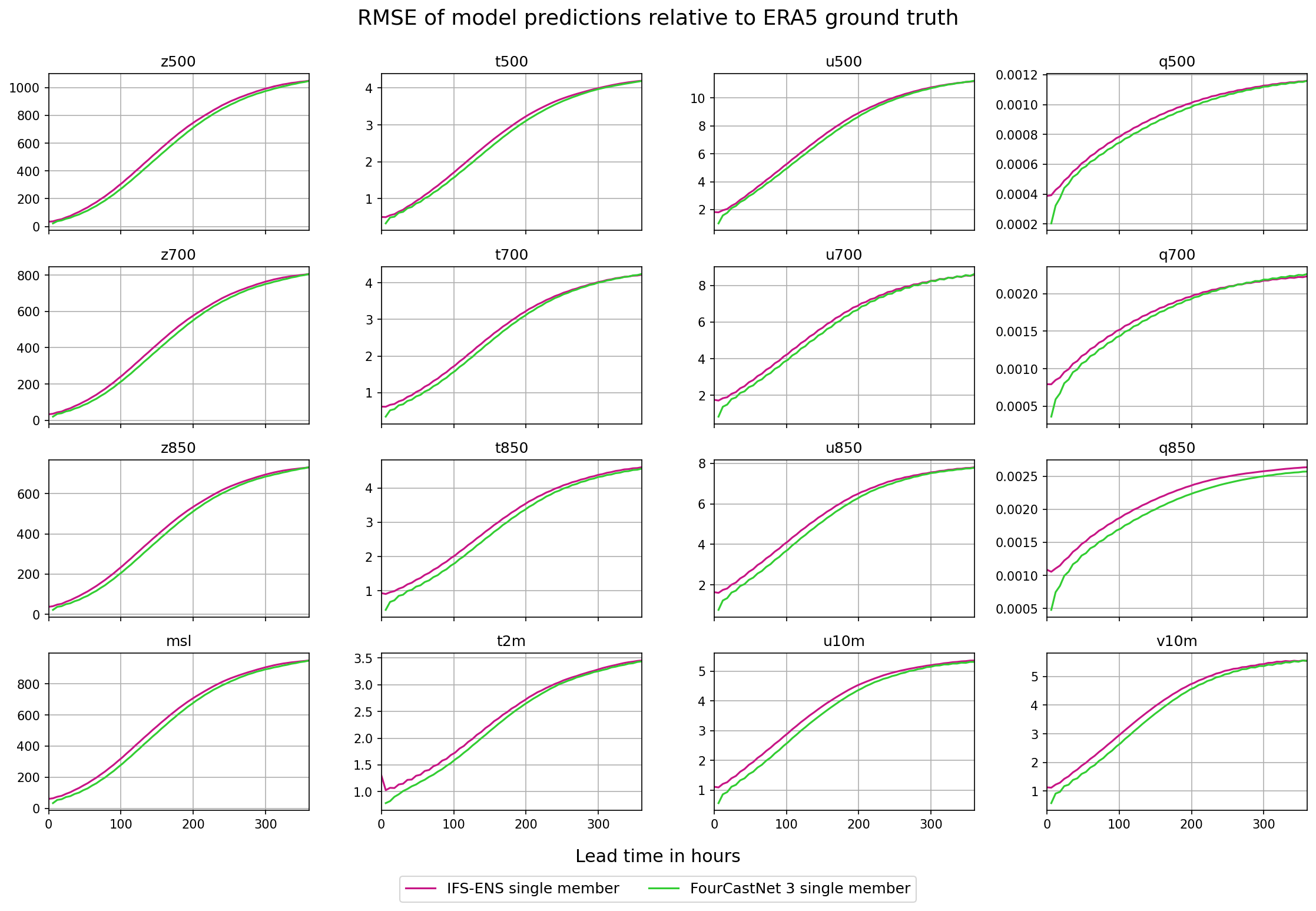}
    \caption{Root mean squared errors (RMSE) of a single ensemble member, averaged over initial conditions at 00:00:00 UTC and 12:00:00 UTC ranging from 2020-01-01 to 2020-12-31. Lower scores indicate better skill.}
    \label{fig:rmse_2020_ens1}
\end{figure}

\begin{figure}[tbp]
    \centering
    \includegraphics[width=1.0\linewidth]{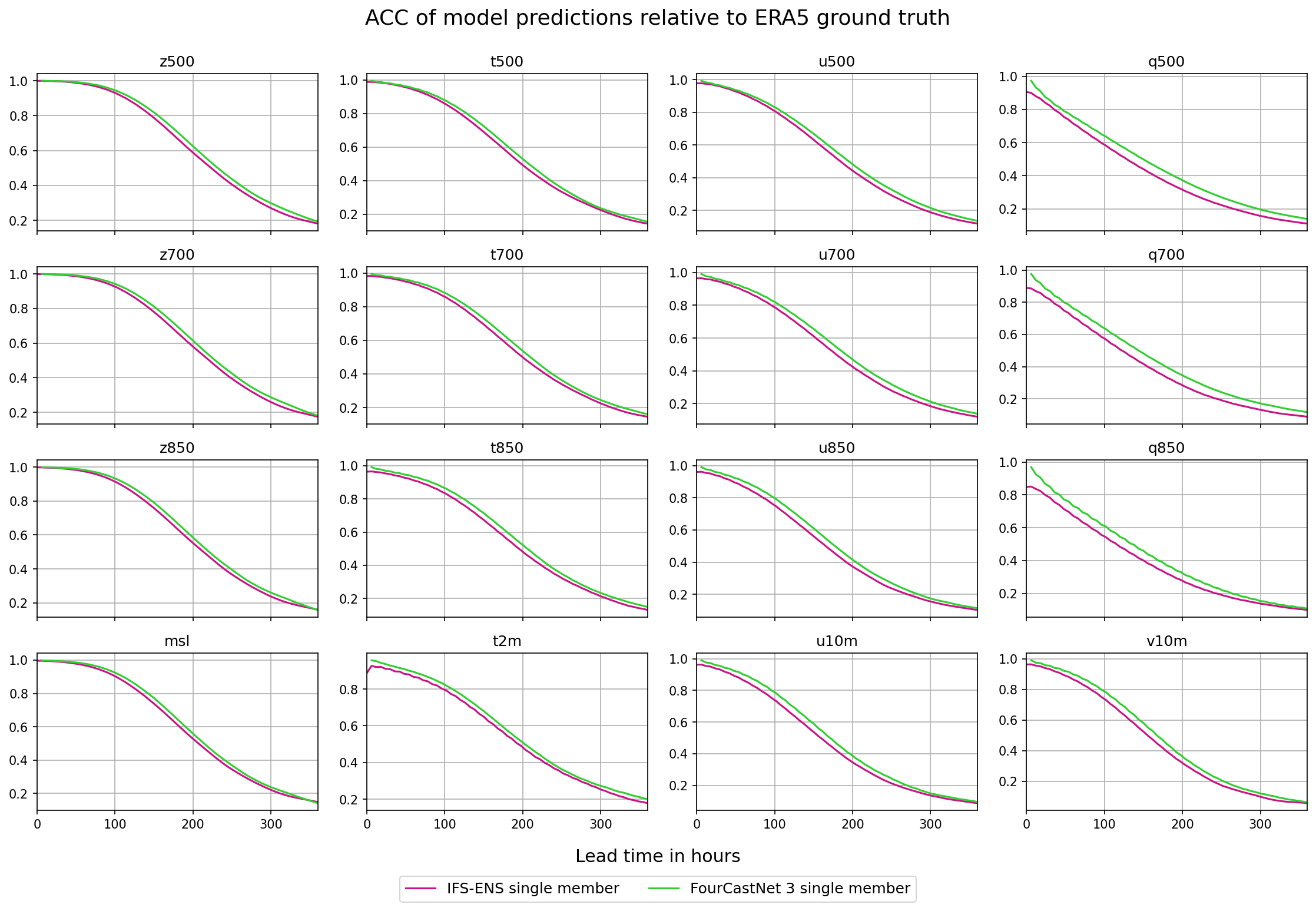}
    \caption{Single member anomaly correlation coefficient (ACC), averaged over initial conditions at 00:00:00 UTC and 12:00:00 UTC ranging from 2020-01-01 to 2020-12-31. Lower scores indicate better skill.}
    \label{fig:acc_2020_ens1}
\end{figure}

While FCN3 operates as an ensemble forecasting system, a single ensemble member can generate deterministic forecasts by fixing the random seeds (and consequently the latent noise variables). This approach yields significantly sharper outputs than traditional ML-based weather forecasts, which can be attributed to the probabilistic loss formulation. Given this enhanced sharpness, FCN3’s deterministic forecasts are more appropriately benchmarked against conventional deterministic numerical weather prediction (NWP) models, as both models share critical attributes: Preservation of physically plausible energy spectra and stability across integration periods. These advantages, however, incur a skill penalty at extended lead times due to inherent predictability limits of atmospheric phenomena, which benefits blurry predictions.

\Cref{fig:rmse_2020_ens1} and \Cref{fig:acc_2020_ens1} depict the RMSE and ACC scores  of a single member FourCastNet 3 forecast relative to the ERA5 ground. Once again, we observe that even a single, deterministic FCN3 forecast outperforms the classical single-member prediction provided by the IFS ensemble.

\subsection{Individual predictions}
\label{sec:individual_predictions}
As the CRPS objective measures a distance between the marginal CDF of the ensemble at a given point for a given channel, it is reasonable to question whether the ML models trained using this objective function result in realistic ensemble members with the correct spatio-temporal behavior.

\begin{figure}[htbp]
    \centering
    \includegraphics[width=1.0\linewidth]{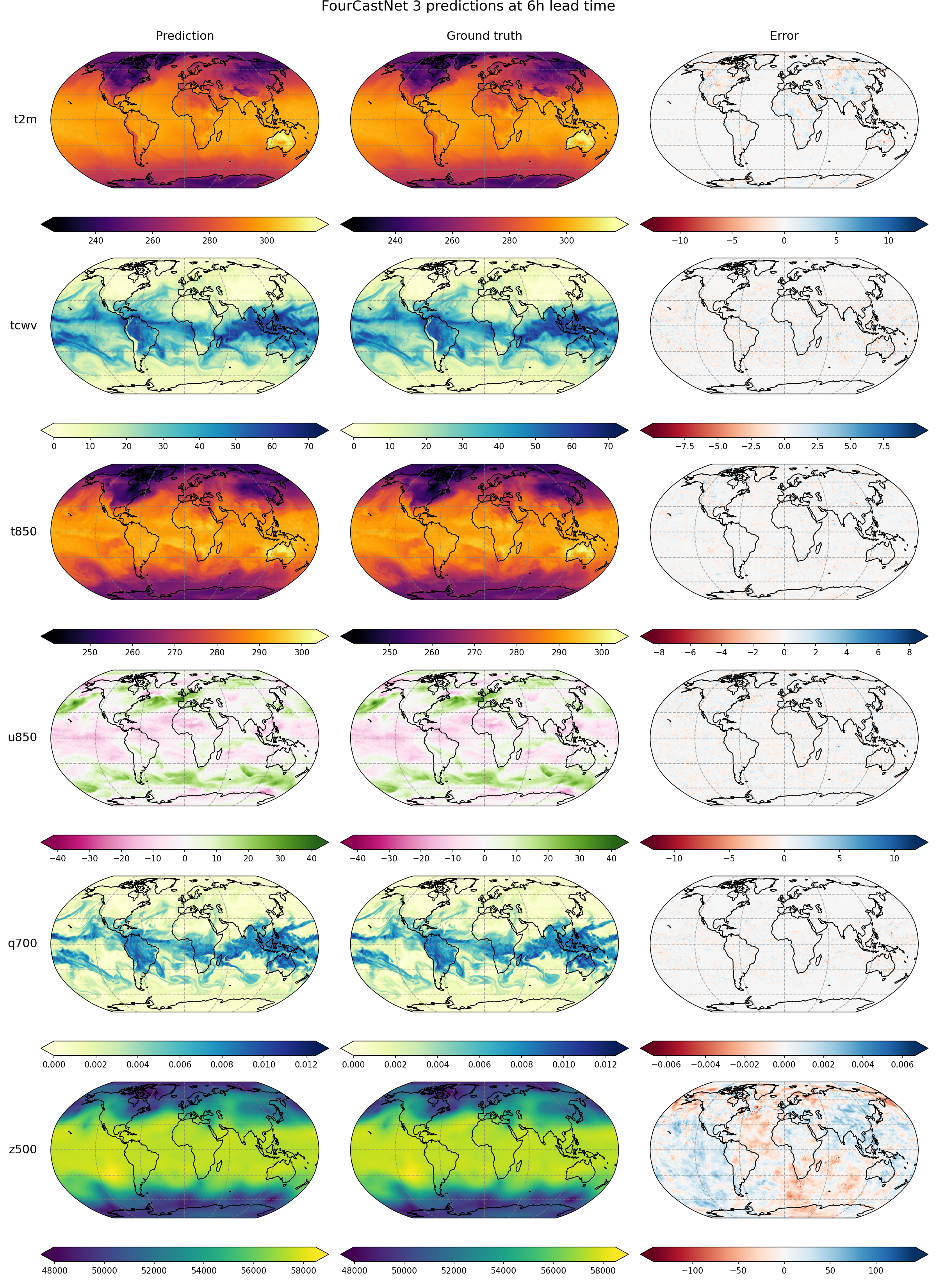}
    \caption{Depiction of a single 6-hour forecast initialized at 2018-01-01 00:00:00 UTC compared to the ground truth data, as well as the difference between the two.}
    \label{fig:6h_prediction_vs_era5}
\end{figure}

\begin{figure}[htbp]
    \centering
    \includegraphics[width=1.0\linewidth]{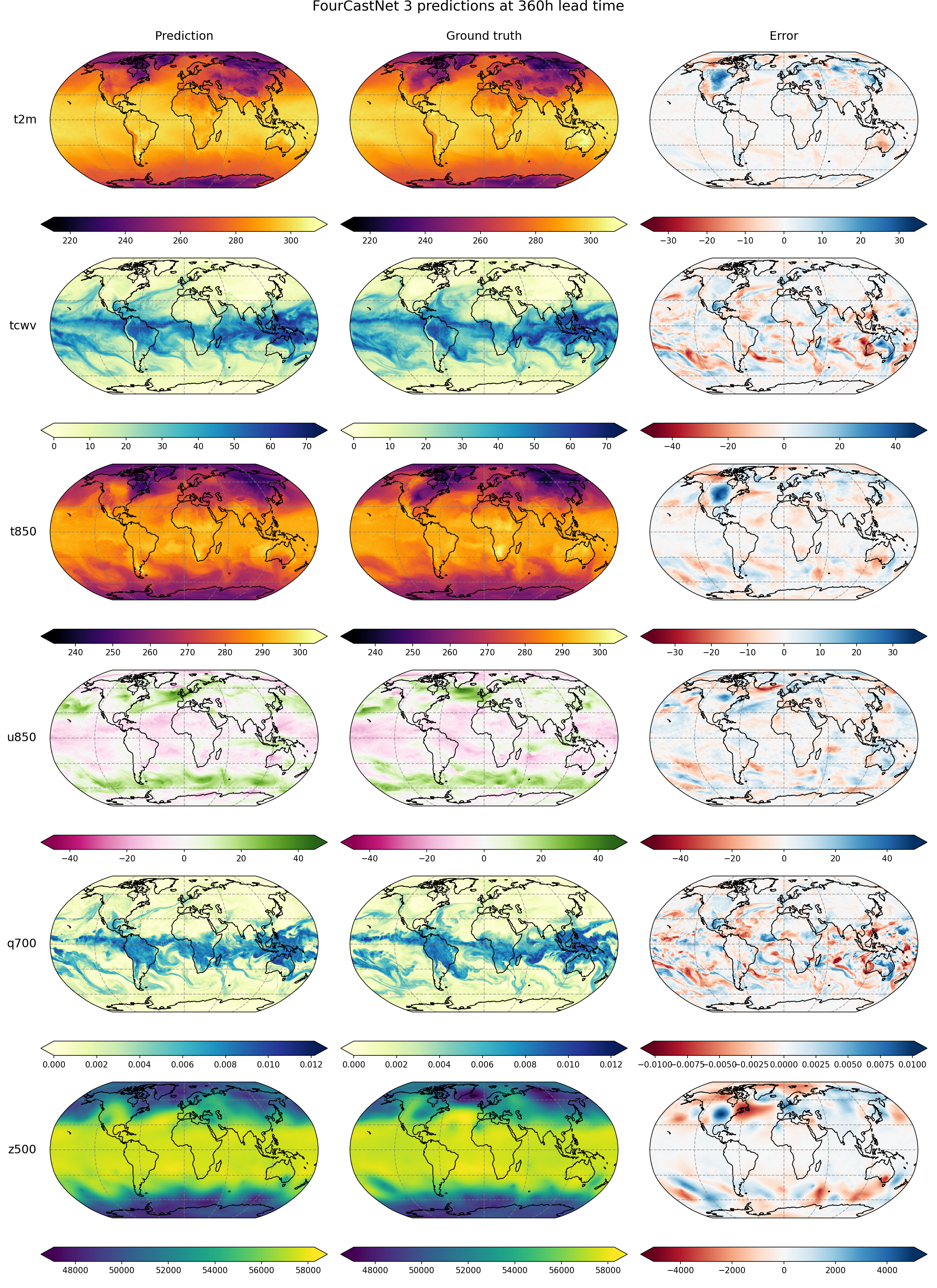}
    \caption{Depiction of a single 6-hour forecast initialized at 2018-01-01 00:00:00 UTC compared to the ground truth data, as well as the difference between the two.}
    \label{fig:360h_prediction_vs_era5}
\end{figure}

\begin{figure}[htbp]
    \centering
    \includegraphics[width=1.0\linewidth]{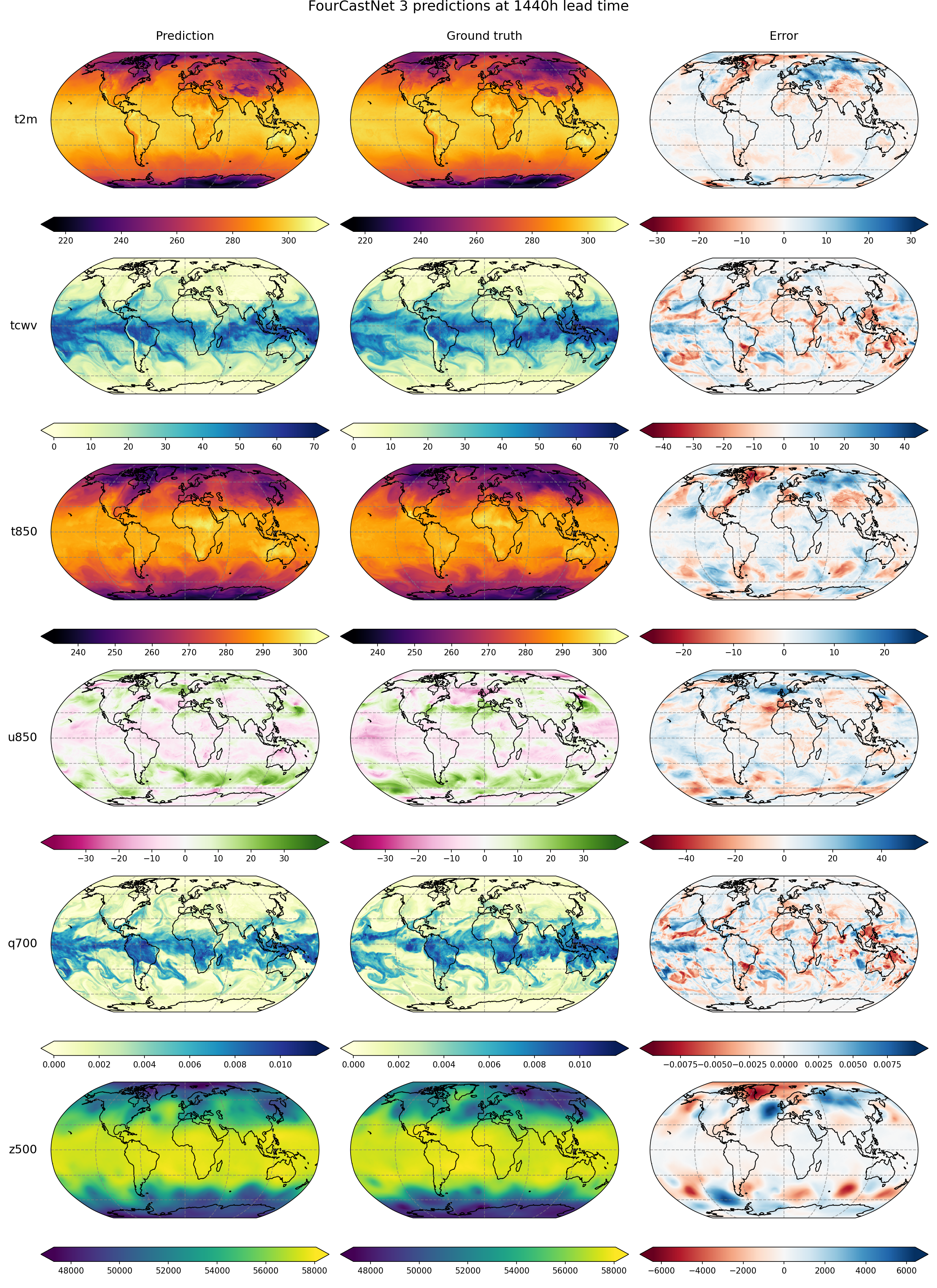}
    \caption{Depiction of a single 60-day forecast initialized at 2018-01-01 00:00:00 UTC compared to the ground truth data, as well as the difference between the two.}
    \label{fig:1440h_prediction_vs_era5}
\end{figure}

We inspect a single ensemble member FCN3 forecast initialized 2018-01-01 at 00:00:00 UTC.
\Cref{fig:6h_prediction_vs_era5}, \Cref{fig:360h_prediction_vs_era5} and \Cref{fig:1440h_prediction_vs_era5} at lead times of 6 hours, 360 hours (15 days) and 1440 hours (60 days), respectively.
Even at these extended lead times, we remark that predictions remain incredibly realistic w.r.t. visual inspection with no observable blurring. This stands in contrast to prior, deterministically trained ML models. Moreover, the stability of the model is remarkable, especially considering that these lead times are well within the subseasonal range. Most autoregressive ML weather models tend to become unstable at these lead times as errors and spurious artifacts build up \cite{Karlbauer2024}.


\subsection{Bias}
\label{sec:bias}

\begin{figure}[htbp]
    \centering
    \includegraphics[width=1.0\linewidth]{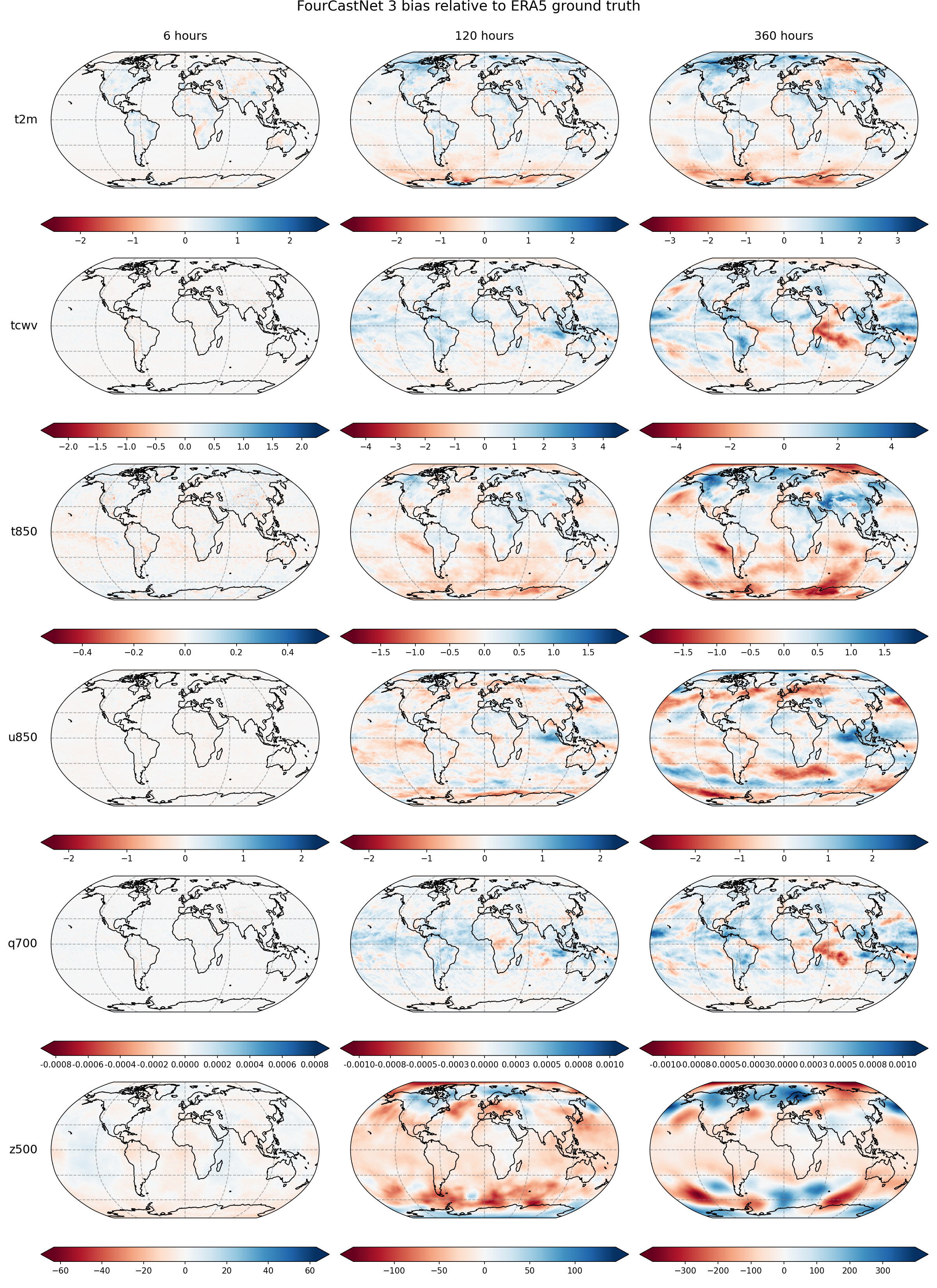}
    \caption{Bias of a single FourCastNet 3 ensemble member relative to the ERA5 ground truth averaged over initial conditions at 00:00:00 UTC and 12:00:00 UTC ranging from 2020-01-01 to 2020-12-31.}
    \label{fig:biases}
\end{figure}

The rank histograms depicted in \Cref{fig:rh_2020_ens50} already hint at a bias in temperature channels predicted by FCN3. To better understand the distribution of this bias $b(x, t)$ over space and lead times, we computed the expected error
\begin{equation}
    b(x, t_n) =  \mathbb{E}_{t_i, e}[u_e(x, t_i + t_n) - u^*(x, t_i + t_n)],
\end{equation}
averaged over initial times $t_i$ and forecast members $e$. \Cref{fig:biases} depicts this bias averaged over the 12-hourly initial conditions in 2020 for a single FCN3 ensemble member at lead times up to 15 days. As expected, we observe a slight cold-bias in t850 and a tendency towards lower 500hPa geopotential in the equatorial region. However, overall biases are distributed fairly uniformly.

\subsection{Spectral properties}
\label{sec:spectral_properties}
While the visual realism and the stability of the method are encouraging, we wish to quantify the degree of physical realism of FCN3 predictions. The spherical signals encountered in atmospheric physics often show a distinct spectral signature, with a characteristic slope resulting from an energy cascade \cite{Tulloch2006}. For individual predictions to be physically consistent, we expect them to obey the same spectral properties. It is therefore useful to analyze the angular power spectral density
\begin{equation}
    \label{eq:psd}
    \mathrm{PSD}[u](\ell) = \sum_{|m| \leq \ell} |\hat{u}^m_\ell|^2,
\end{equation}
given in terms of the squared absolute values of the spherical harmonic coefficients $\hat{u}^m_\ell$ corresponding to $u$. Similarly, we may also compute the zonal power spectral density, which is simply the power-spectrum at a fixed latitude. As such, it is computed by taking the Fourier-transform in the zonal (longitudinal) direction and normalized by the circumference of the latitude.
\begin{equation}
    \label{eq:zonal_psd}
    \mathrm{zonalPSD}[u](\theta, m) = 2 \pi \sin\theta \; \left|\int_0^{2\pi} \, u(\theta, \varphi) e^{-im\varphi}\,\mathrm{d}\varphi \right|^2.
\end{equation}

\begin{figure}[htbp]
    \centering
    \includegraphics[width=1.0\linewidth]{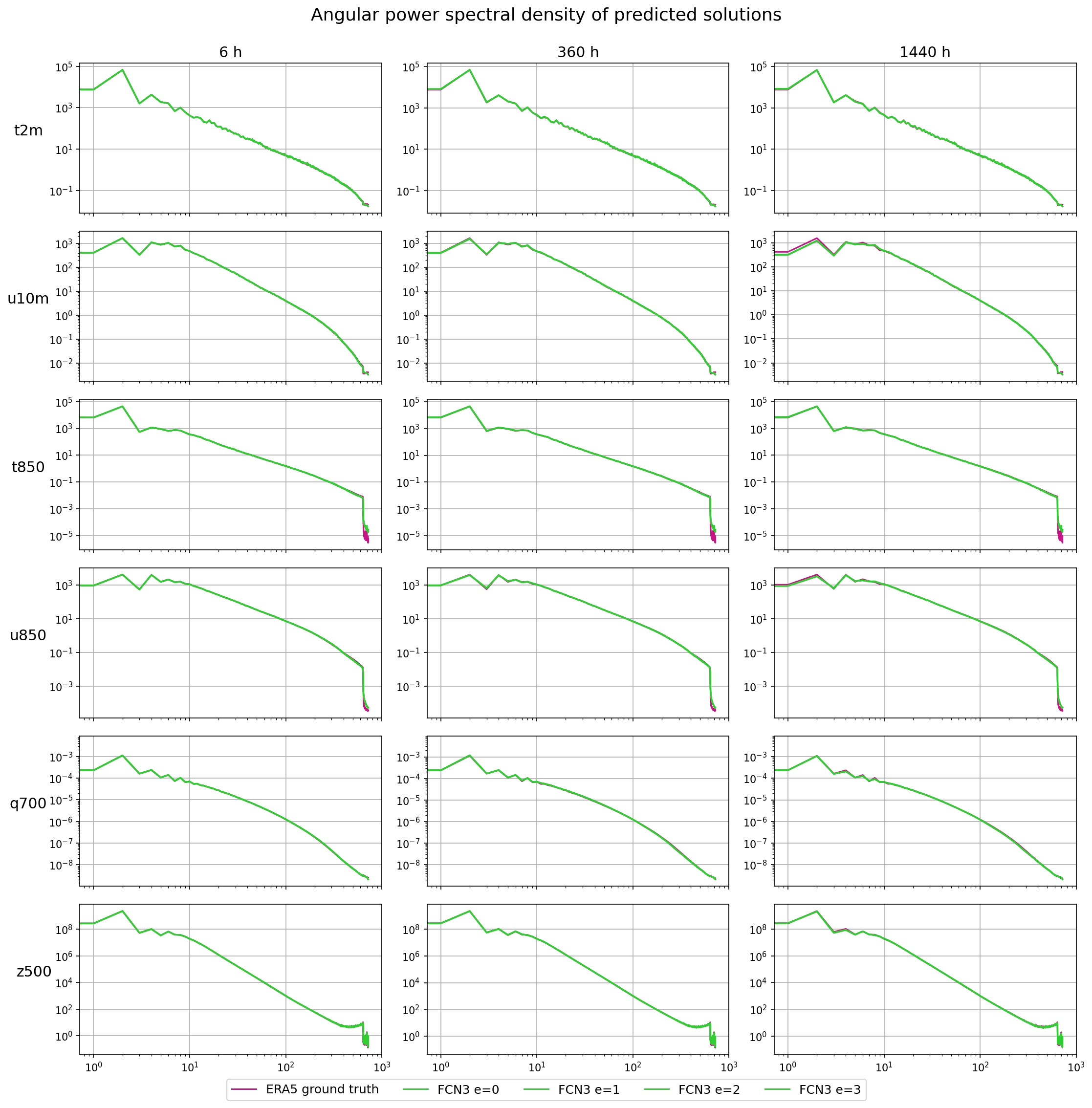}
    \caption{Angular power spectral density of FCN3 forecasts at lead times of 6, 360 and 1440 hours, averaged over daily initial conditions at 00:00:00 UTC and 12:00:00 UTC over the year 2020.}
    \label{fig:angular_psd_2020}
\end{figure}

\Cref{fig:angular_psd_2020} depicts the angular power spectral density of four FCN3 forecast members initialized at 2018-01-01 00:00:00 UTC. We observe that the individual forecast members (in green) faithfully reproduce the power spectra of the ground-truth ERA 5 data. This includes the spectral cutoff which results from the re-analysis being conducted on the T639 grid, which contains spectral coefficients up to degree $l=639$. It is remarkable that spectra are accurate even for extended lead times of 1440 hours, which correspond to 60 days and 240 autoregressive steps. This is well into the subseasonal range, where autoregressive stability and physical faithfulness are imperative to obtain skillful forecasts.

\begin{figure}[htbp]
    \centering
    \includegraphics[width=1.0\linewidth]{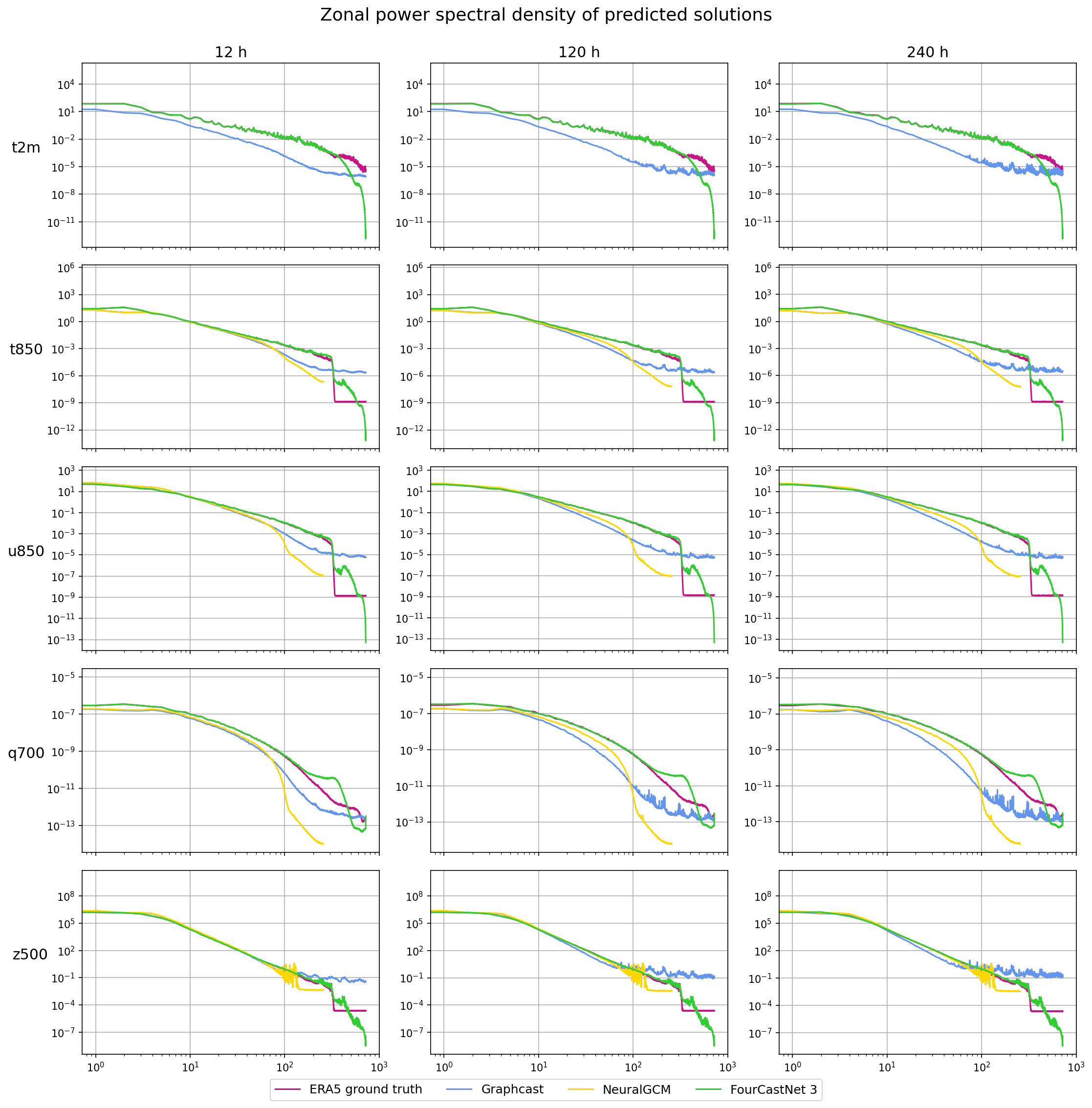}
    \caption{Zonal power spectral density of ML weather forecasts at the 60\textdegree  latitude ring at lead times of 6, 120 and 240 hours, averaged over 6-hourly initial conditions in 2020.}
    \label{fig:zonal_psd_2020}
\end{figure}

\Cref{fig:zonal_psd_2020} further illustrates the advantages of our probabilistic approach. It plots the average zonal power spectral density over 6-hourly initial conditions in 2020 at the 60\textdegree latitude ring on the Northern Hemisphere. Once again, we observe that a single FourCastNet 3 ensemble member correctly approximates the ground truth zonal PSD up to the cutoff frequency at this latitude. The plot also depicts zonal spectra of GraphCast \cite{Lam2022} and NeuralGCM \cite{Kochkov2023}, two popular ML weather models based. While GraphCast is a deterministic ML model, NeuralGCM is a probabilistically trained  hybrid model between a General Circulation model and an ML model. In contrast to FCN3, both of these models exhibit progressive blurring, which results in forecasts resembling an ensemble forecast. This tends to be favorable for good RMSE scores, however it is unrealistic in terms of power spectral density.

\section{Scaling FourCastNet 3}
\label{sec:parallelism}

Training of large ML weather models quickly becomes performance limited by I/O and memory capacity and bandwidth as well as compute performance. Increasing distributed memory parallelism has repeatedly proven to be an effective method for achieving competitive performance in ML models. It is therefore crucial to keep these considerations in mind when designing ML models and training infrastructure. In the case of ML weather models and the broader scientific ML domain, these considerations are different from other areas of machine-learning research. For instance, model inputs, as well as activations are typically high-dimensional signals, which drastically increases the IO load and memory requirements. As such, popular distribution strategies such as FSDP \cite{Zhao2023} and or channel parallelism do not translate well into the scientific ML domain.

To address this problem, we have developed Makani, an open-source library for massively parallel large-scale training of ML weather models, featuring a novel model-parallelism paradigm, which splits both the model and the training data across ranks, using domain decomposition. This is inspired from 

In the following section, we describe this new model-parallelism as well as other aspects of scaling and distributing FourCastNet 3 training.

\subsection{Simultaneous model-, data- and ensemble-parallelism}
\begin{figure}[tp]
    \centering
    \includegraphics[scale=0.3]{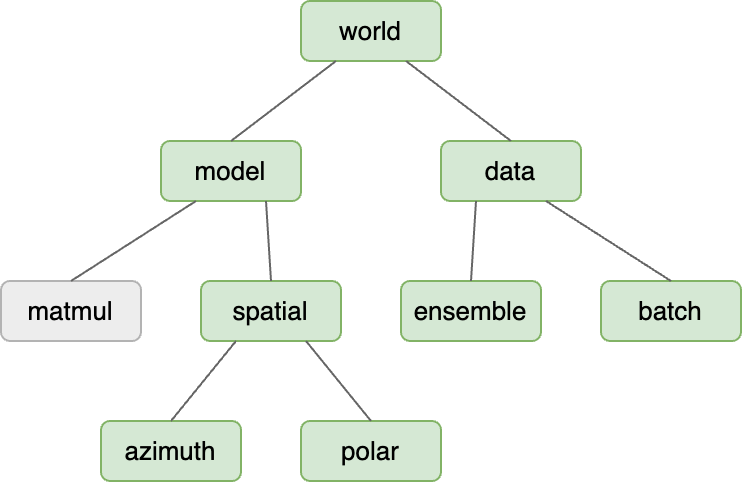}
    \caption{Communicator hierarchy used in training and inference. To enable multiple forms of parallelism, GPUs are grouped into distinct communicator groups, with each GPU assigned a unique rank within each group. Communication for parallel operations occurs only among GPUs within the same group. The matmul communicator group, shown greyed out, is supported by Makani but not utilized in this work.}
    \label{fig:communicator_groups}
\end{figure}

To enable the training of large machine-learning weather models at scale, Makani supports multiple forms of parallelism. To achieve this, Makani creates a hierarchy of orthogonal communicator groups illustrated in \Cref{fig:communicator_groups}. 

Additional structure is added by grouping those into data-parallel and model-parallel groups. The former are comprised of batch and ensemble communicators and the latter of azimuth (longitude) polar (latitude) communicators.\footnote{Note that model-parallelism in this work is recently sometimes referred to ad tensor-parallelism in the community because model-parallelism can also incorporate pipeline-parallelism, which is currently not supported by Makani.}
Makani supports a parallelization mode called matmul, which basically splits matrix multiplications in two-layer MLPs in a fork-join distributed manner by using row-wise and column-wise matrix decompositions. Since most of our feature matrices are tiny, this feature is not used in this work, but transformer architectures with large MLP dimensions can benefit from it.

Orthogonality in the case of communicators means that every rank in a leaf group is a member of exactly one communicator of that group. Additionally, all ranks $k$ for all communicators in a group $A$, are combined into another communicator. The rank ID for ranks in that communicator is derived from the communicator ID of the group $A$. For example, the rank $k$ from communicator $i$ in a group $A$ will be assigned rank $i$ in the new communicator. The union of all those communicators then forms a group $A_\perp$, which is orthogonal to $A$. 

This hierarchy can be displayed as a tree and can be constructed in a top-down fashion, starting at the root node, commonly called the world group. This group consists of all ranks of the whole run.
This group is then subdivided into $2$ or more groups on the same level, where all groups on each level are orthogonal in the sense as described above. 

Nodes on the same level of the abstract communicator tree do not have a canonical order and can in theory be ordered arbitrarily. Order only matters once the abstract construct is mapped to physical network communication hardware underneath. This hardware usually exhibits a hierarchy in terms of latency and bandwidth. Intra-node communication on the machines we are running on usually happens through the NVLink 3, supported by NVSwitch, a high bandwidth, low latency interconnect, while internode communication is routed through the lower bandwidth Infiniband interconnects. Groups of 32 nodes are called leaf groups and can communicate with a single hop over a single switch. Then 32 of these leaf groups are connected to a spine switch each, which in turn is connected to the core group of switches. Due to communication congestion and link sharing, latency and bandwidth drop significantly with every additional hop in the network topology. The mapping of our communicator graph \Cref{fig:communicator_groups} to the underlying network topology needs to reflect that. 

It turns out, that all-to-all communications (i.e. distributed data transpositions) are more highly affected by multi hop communication than, for example, all-reduce operations. The reason is that the latter can be accelerated by in-network computation technologies such as SHARP and further by algorithmic tricks, whereas the former can no. Therefore, it is advised to map communicators which predominantly perform all-to-all communications to lower level components and spread the all-reduce type communications over higher levels. To that end, we sort our leaf nodes such that all model parallel computation is as \emph{close together} in the hardware as possible, spread out across a single or a few nodes at best.
Ensemble parallelism plays a special role, as we will see below: it actually does benefit from all-to-all communications, but we nevertheless map this to a higher level than any of the model parallel communications. This is because most of the ensemble parallel computations are embarrassingly parallel in nature, and just a single all-to-all communication call is performed at the end of each step to compute the CRPS score. Therefore, it is possible to treat that parallel dimension as not performance critical in terms of communication.

\subsection{Model-parallelism through domain decomposition}

The algorithms for model-parallelism employed by Makani are closely related to techniques applied to traditional numerical simulations in high-performance computing over the last 3 to 4 decades. The communication primitives underlying these algorithms are usually highly optimized and tuned for high-performance computer systems with high bandwidth, low latency interconnects. Since these infrastructures are what we are targeting with Makani, it is reasonable to make use of these developments.

\begin{figure}[htbp]
    \centering
    \includegraphics[width=0.6\textwidth, trim={100pt 80pt 100pt 80pt}, clip]{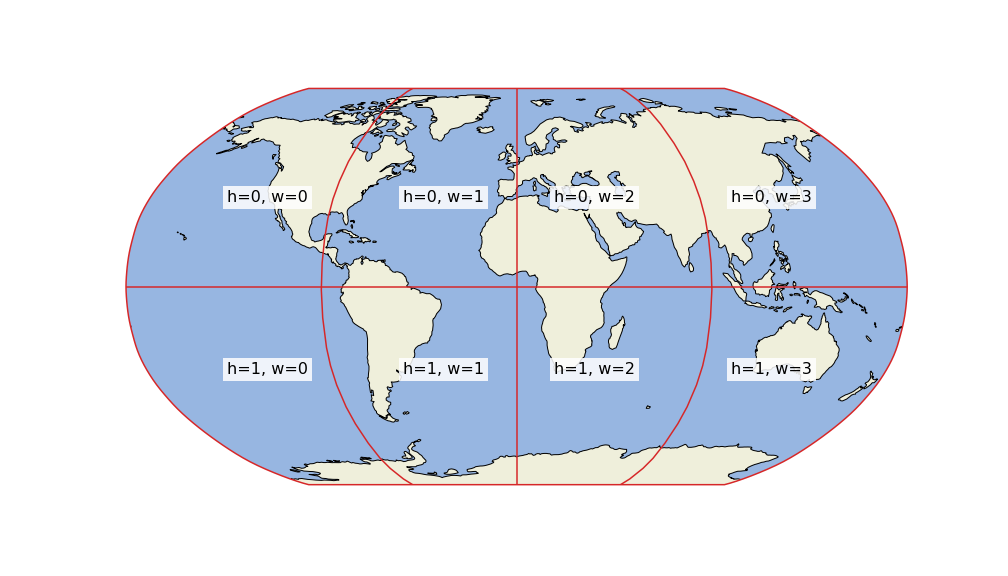}
    \caption{Illustration of domain decomposition within Makani for an azimuth group of 4 and a polar group of 2. The signal is split into 8 subdomains by subdividing the sphere into two parts along the polar (latitudinal) axis and into four parts along the azimuthal (longitudinal) axis.}
    \label{fig:spatial_parallelism}
\end{figure}

The primary mode of model-parallelization is through domain decomposition. Since neural networks perform many element-wise computations (multiplications, additions, activation functions, point-wise MLPs, etc.), these operations are embarrassingly parallel in this setup. Below we explain how we have implemented the more complicated distributed operations which require communication. We decompose all directions as evenly as possible, but since the data was re-gridded from a Gaussian grid, some grid dimensions are not evenly divisible in some cases. Makani supports this situation by keeping track of all the split shapes of all tensors and ensuring that during all-gather and all-to-all operations, the correct shapes are used. We never truncate or interpolate data to enable even splits because such a procedure inherently causes loss of information. During down-sampling operations, we use optimal dimension sizes derived from spherical signal processing, usually leading to down-sampled sizes which are also not evenly divisible. 

Note that for reasons of disambiguation, we will denote the spatial dims of our signal tensors with $H$ and $W$ respectively. This is common in computer vision applications, and it is also easier to abbreviate than latitude and longitude with a single letter. However, despite this nomenclature, keep in mind that we treat all data as spherical, multiplying the corresponding spherical quadrature weights $\omega$ before computing spatial sums. The values of those weights are dependent on the grid the tensor is defined on.

\subsubsection{Notation}

In this section, we denote dimension sizes by capital letters, such as $B,C,E,H,W$ for batch size, number of channels/variables, number of ensemble members, latitudes and longitudes respectively. If we refer to the local dimension sizes, we add a $loc$ suffix to the corresponding dimension. For example, $Hloc = H/nH$, where $nH$ is the corresponding communicator size, i.e. the number of ranks in latitude direction. The only exception is batch size $B$. We never need to distinguish between $B$ and $Bloc$, thus $B$ will always be the rank-local batch size. 

\subsubsection{Distributed spherical Harmonic transforms}
We recall from \Cref{spherical_harmonics_transforms} that the computation of SHTs can be decomposed into a one-dimensional Fourier transform along the azimuthal dimension and a matrix multiplication in the polar dimension to perform the Legendre transformation. To perform this on domain-decomposed data, we use global transpositions (all-to-all) to change the data decomposition temporarily by splitting the feature dimension and collecting the corresponding spatial dimension at the same time. Then, the respective transform (Fourier for longitude or Legendre for latitude) is applied locally but concurrently on all ranks. The data is then transformed back to mimic the previous decomposition.
This method is called pencil decomposition and commonly used for operations such as distributed Fourier transforms in high-performance computing. The pseudocode for this procedure is shown in \Cref{alg:distributed_sht_pseudocode}. 

\begin{algorithm}
\caption{Distributed Spherical Harmonics Transform (Forward)\label{alg:distributed_sht_pseudocode}}
\begin{algorithmic}
\Procedure{Forward}{$x$, $mmax$} 
\Comment{$x$: tensor, shape $\texttt{B} \times \texttt{C} \times \texttt{Hloc} \times \texttt{Wloc}$}
\Comment{$mmax$: integer, number of modes to keep}
\State $xt \gets \texttt{distributed\_transpose}(x, \mathrm{dims}=(1, 3))$
\Comment{$xt$: tensor, shape $\texttt{B} \times \texttt{Cloc} \times \texttt{Hloc} \times \texttt{W}$}
\State $xf \gets 2\pi \cdot \texttt{FFT\_1D\_R2C}(xt, \mathrm{dims}=3)$ \Comment{Apply local FFT in longitudinal direction}
\State $xf \gets xf[…, :\text{mmax}]$ \Comment{Mode truncation to maximum wave-number, if requested}
\State $xt \gets \texttt{distributed\_transpose}(xf, \mathrm{dims}=(3, 1))$ \Comment{$xt$: tensor, shape $\texttt{B} \times \texttt{C} \times \texttt{Hloc} \times \texttt{mmaxloc}$}
\State $x \gets \texttt{distributed\_transpose}(xt, \mathrm{dims}=(1, 2))$
\Comment{$x$: tensor, shape $\texttt{B} \times \texttt{Cloc} \times \texttt{H} \times \texttt{mmaxloc}$}
\State $xc_{bclm} \gets \sum_{k} L_{mlk} x_{bckm} \cdot $ \Comment{Legendre-Gauss quadrature via tensor contraction}
\Comment{$L$: tensor, shape $\texttt{mmaxloc} \times \texttt{lmax} \times \texttt{H}$}
\State $x \gets \texttt{distributed\_transpose}(xc, \mathrm{dims}=(2, 1))$ \Comment{$x$: tensor, shape $\texttt{B} \times \texttt{C} \times \texttt{lmaxloc} \times \texttt{mmaxloc}$}
\State \Return $x$
\EndProcedure
\end{algorithmic}
\end{algorithm}

Here, $\texttt{B}$ denotes the batch size, $\texttt{C}$ the number of channels (or variables/features), $\texttt{H}$ the height (latitude dimension) and $\texttt{W}$ the width (longitude dimension). The spectral mode sizes are denoted by $\texttt{lmax}$ and $\texttt{mmax}$. Since we are performing a real to complex transform, the number of accessible modes in spectral space is usually cut in half and the degrees of freedom are propagated into the phases of the now complex output tensor. For equiangular grids for example, $mmax=W/2$ but when mode truncation is used, it could be smaller. The suffix $\texttt{loc}$ denotes that the corresponding axis was split across ranks, and only the rank-local part of the tensor is stored along this direction.
The tensor $L$ contains pre-computed Legendre coefficients multiplied with the corresponding quadrature weights $\omega_h$ which are mainly dependent on latitude (but not on longitude, for the grids we are considering in this work). Generally, the quadrature weights could be kept separately, but in order to minimize the number of mathematical operations, we fold those weights into the Legendre coefficient tensor $L$.

\subsubsection{Distributed DISCO convolutions}

The DISCO convolution is mathematically formulated as a product of a sparse matrix denoted by $\psi$ (the convolution tensor encoding the basis functions) with a dense matrix $f$ (the input tensor, cf. \cite{Ocampo2022}) for all possible shifts in longitude:

\begin{equation}\label{eq:disco_convolution_formula}
\hat{f}[b,c,k,h,w] = \sum\limits_{{w'}=0}^{\mathrm{nlon}-1} \sum\limits_{{h'}=0}^{\mathrm{nlat}-1}  \psi[k,h,h',w']\,f[b,c,h',w'+w],
\end{equation}
where $b$ is the batch size, $k$ the number of kernel basis functions, $c$ the number of channels and $h,w,h',w'$ are longitude and latitudes for output and input respectively.

Note that $\psi$ does not depend on the output longitude, but instead the dense input tensor is shifted by output longitude. Because of this shift, it is more efficient to actually gather all longitude components on a single rank so that the sum over $w'$ can be performed on each rank. Since there is no channel dependence on $\psi$, we can apply the same trick as we apply in the case of the spherical harmonics transform before: we transpose the data globally so that  afterwards, channels are distributed across ranks and the longitudes are all local. We then compute \Cref{eq:disco_convolution_formula} per rank, followed by a reduce-scatter over the latitude dimension. This last part finalizes the sum over $h'$ and also splits the output tensor in latitude-dimension. Finally, we transpose the data so that channels are rank-local and output longitudes are distributed in longitude-dimension.
The weight contraction is performed per rank in an embarrassingly parallel fashion, since this is only over the rank-local $K$ and $C$ dimensions.

\begin{algorithm}
\caption{Distributed DISCO (Forward)\label{alg:distributed_disco_pseudocode}}
\begin{algorithmic}
\Procedure{Forward}{$x$, $psi$} 
\Comment{$x$: tensor, shape $\texttt{B} \times \texttt{C} \times \texttt{Hloc} \times \texttt{Wloc}$}\\
\Comment{$psi$: sparse tensor, shape $\texttt{K} \times \texttt{Hloc\_out} \times \texttt{Hloc} \times \texttt{W}$}
\State $xt \gets \texttt{distributed\_transpose}(x, \mathrm{dims}=(1, 3))$
\Comment{$xt$: tensor, shape $\texttt{B} \times \texttt{Cloc} \times \texttt{Hloc} \times \texttt{W}$}
\State $xc \gets \texttt{contract}(\psi,xt)$ \Comment{$xc$: tensor, shape $\texttt{B} \times \texttt{Cloc} \times \texttt{K} \times \texttt{H\_out} \times \texttt{W\_out}$}
\State $xr \gets \texttt{reduce\_scatter}(xc, \mathrm{dim}=2)$ \Comment{$xr$: tensor, shape $\texttt{B} \times \texttt{Cloc} \times \texttt{K} \times \texttt{Hloc\_out} \times \texttt{W\_out}$}
\State $x \gets \texttt{distributed\_transpose}(xr, \mathrm{dims}=(3, 1)) $\Comment{$x$: tensor, shape $\texttt{B} \times \texttt{C} \times \texttt{K} \times \texttt{Hloc\_out} \times \texttt{Wloc\_out}$}
\State \Return $x$
\EndProcedure
\end{algorithmic}
\end{algorithm}

Here, \texttt{contract} computes \eqref{eq:disco_convolution_formula} and is performed by an optimized CUDA kernel we have implemented. For the backward pass, we also implemented a CUDA kernel for the contraction part, while the rest of the calculations are handled by PyTorch's autograd functionality \cite{Paszke2019}.

\subsubsection{Distributed evaluation of CRPS loss}\label{sec:distributed_crps}
In this work, we are using the spread-skill \eqref{eq:spread_skill_crps} and CDF definition \eqref{eq:ensemble_crps} of CRPS. The distributed implementation can be unified for these two cases, for which only the rank-local kernels differ.
Both kernels rely on sorting the data across the ensemble dimension, either in order to compute a rank score or a CDF of the data. This score is computed per input sample, channel and grid point. Therefore, we transpose the data globally so that the ensemble dimension becomes local, and the (flattened) spatial dimension of size $\texttt{nlat}\cdot \texttt{nlon}$ is subdivided further. Note that one could also split the channel dimension as we did for the distributed spherical harmonics transform or DISCO. However, since we aim at supporting a large degree of ensemble parallelism where the number of ensemble members is much bigger than the number of channels, subdividing the spatial domain further is a better option to ensure scalability. The algorithm for the distributed computation of the CRPS score is described below:

\begin{algorithm}
\caption{Distributed CRPS (Forward)\label{alg:distributed_crps_pseudocode}}
\begin{algorithmic}
\Procedure{Forward}{$f$, $o$} 
\Comment{$f$: forecast tensor, shape $\texttt{B} \times \texttt{Eloc} \times \texttt{C} \times \texttt{Hloc} \times \texttt{Wloc}$}\\
\Comment{$o$: observation tensor, shape $\texttt{B} \times \texttt{C} \times \texttt{Hloc} \times \texttt{Wloc}$}
\State $ff \gets f.reshape(B, Eloc, C, Sloc)$
\State $ft \gets \texttt{distributed\_transpose}(ff, \mathrm{dims}=(1, 4))$
\Comment{$ft$: tensor, shape $\texttt{B} \times \texttt{E} \times \texttt{C} \times \texttt{SlocE}$}
\State $of \gets f.reshape(B, C, Sloc)$
\State $ot \gets \texttt{scatter}(of, \mathrm{dim}=2)$ \Comment{$ot$: tensor, shape $\texttt{B} \times \texttt{C} \times \texttt{SlocE}$}
\State $sl \gets \mathrm{crps\_kernel}(ft, ot)$ \Comment{$s$: tensor, shape $\texttt{B} \times \texttt{C}$}
\State $s \gets \texttt{all\_reduce}(sl)$ \Comment{Reduction over latitude, longitude and ensemble direction}
\State \Return $s$
\EndProcedure
\end{algorithmic}
\end{algorithm}
where $Sloc = Hloc * Wloc$ and $SlocE = Sloc / nE$ with $nE$ being the size of the ensemble communicator direction. The function $\mathrm{crps\_kernel}$ is the rank-local implementation of the skill-spread or CDF kernel variant. 
The final score is then averaged over the batch and channel dimension, where weights can be applied to the latter average. 

\subsection{Model weight sharding and distributed weight gradients}
The algorithms above describe how to compute distributed forward (and, via PyTorch autograd, also backward) passes, but in order to perform a full weight update step, one also has to compute the weight gradients. Traditional data parallelism implemented via PyTorch \texttt{DistributedDataParallel} (DDP) paradigm, averages all gradients along the batch communication dimension, specified by a process group passed to the corresponding DDP constructor. For most applications, this is sufficient. Makani supports changing tensor parallelism inside the model, such that some layers can be spatial parallel, but others are instead feature parallel. In this case, some weights are shared across model parallel communication dimensions and in some case they are not. The weight gradient reduction has to be performed over all gradients for weights that are shared across ranks, a case which cannot fully be captured by the simplistic DDP wrapper. However, PyTorch allows users to customize the gradient reduction step by registering a communication hook. This hook is executed after the backward pass completes and can be used to implement arbitrary communication patterns for the weight gradients.

On the hook level, the user can only query weight and gradient tensor information, there is no insight into what layer that tensor originated from. To this end, we need to add some additional information to the weight tensors in order to allow proper treatment in the gradient reduction step. The first assumption, which is always correct, is that all weights need to be reduced along the batch communication dimension. This operation is implemented as a non-blocking all-reduce, which returns a future object. To this future object, additional operations can be chained to using the \texttt{then} member. Each subsequent step is required to return another future that can be chained to other operations.

We implement other shared weight reductions over communication dimensions orthogonal to the batch dimension, by registering additional all-reduce operations for respective weight gradients. To identify which gradients need to be averaged over which communication dimensions, we annotate each weight tensor inside a Makani model with a list of communication groups (identified by their lookup keys in the global communicator dictionary), the corresponding gradient tensor needs to be averaged over. This information can be queried during hook construction to determine which additional reductions have to be generated. Those reductions are performed once per communicator dimension for all weight gradients which require reduction in this specific dimension. This ensures that communications are efficiently pooled as opposed to being performed on a per-gradient basis. The latter occurs if distributed weight gradient reductions were implemented using the post-gradient-hook functionality in PyTorch. This allows the user to make individual operations or layers self-contained with respect to tensor parallelism, but at the cost of increased communication time.

For checkpoint saving and restoring, it is further important to know which weights need to be gathered from or split across which communicator groups. For this purpose, we annotate each weight tensor with the corresponding sharding information for each dimension. We do not store the degree of parallelism, just the communication dimension the corresponding tensor dim needs to be split across.
This allows us to also change the degree of tensor parallelism during checkpoint reload, a feature which is important to mitigate potential memory limitations during fine-tuning. For example, we can increase the degree of tensor parallelism in a certain communication dimension to accommodate the memory increase when increasing the autoregressive rollout steps the model is trained on.


\subsection{Inference on distributed memory machines}
\label{sec:distributed_inference}
Since individual ensemble members can be generated in a highly parallel fashion for a minimal computational cost, producing large ensembles is extremely cheap compared to traditional methods such as NWP \cite{Mahesh2024a, Mahesh2024b}. 
This enables but also enforces a paradigm shift, where storing all generated data is neither feasible nor necessary anymore. A close analogy are modern large-language models: those models essentially distill large amounts of text corpus data in a comparable small machine learning model which can be queried by the user. Similarly, machine learning-based weather prediction models could be viewed as distilled versions of the dataset they were trained upon. 
In Makani, we implemented a fully distributed inference pipeline which is capable of computing many relevant skill scores used by the weather forecasting community. In fact, the distributed inference pipeline supports all parallelism options which are also supported by the training pipeline, allowing for a seamless integration into end to end training and evaluation workflows.
Makani still supports storing multi-year rollouts from individual initial conditions on disk for the purpose of archiving, visualization, etc. To achieve this, we employ parallel HDF5 with adjustable double buffering in CPU memory. This allows the user to tune the size of data written vs what is kept in memory. In the future, we are planning to add streaming I/O support directly from the GPU via GPUDirect Storage.

\end{document}